\documentclass{article}

\PassOptionsToPackage{numbers, compress}{natbib}

\usepackage[final]{neurips_2023}
\usepackage{booktabs}       %
\newcommand{\greyrule}{\arrayrulecolor{black!10}\midrule\arrayrulecolor{black}}
\usepackage{graphics}
\usepackage{mathtools}

\usepackage{enumitem}

\usepackage{amsfonts}       %
\usepackage{amsmath}       %
\usepackage{amssymb}
\usepackage{amsthm}
\usepackage{pifont}%
\newcommand{\cmark}{\ding{51}}%
\newcommand{\xmark}{\ding{55}}%
\usepackage{diagbox}
\usepackage{multirow}

\makeatletter
\newcommand\notsotiny{\@setfontsize\notsotiny{6.31415}{7.1828}}
\makeatother

\newcommand{\Fig}[1]{Figure~\ref{#1}}  %
\newcommand{\fig}[1]{Fig.~\ref{#1}}    %

\newcommand{\alg}[1]{Alg.~\ref{#1}}
\newcommand{\tab}[1]{Table~\ref{#1}}

\newcommand{\eqn}[1]{Eq.~\ref{#1}} %
\newcommand{\eqnp}[1]{(Eq.~\ref{#1})} %
\renewcommand{\sec}[1]{Sec.~\ref{#1}} %
\newcommand{\supp}[1]{Suppl.~\ref{#1}}
\newcommand{\app}[1]{Appendix.~\ref{#1}}

\usepackage{xspace}
\makeatletter
\DeclareRobustCommand\onedot{\futurelet\@let@token\@onedot}
\def\@onedot{\ifx\@let@token.\else.\null\fi\xspace}
\def\eg{e.g\onedot}
\def\ie{i.e\onedot}

\makeatother

\DeclareMathOperator*{\argmax}{arg\,max}
\DeclarePairedDelimiter{\nint}\lfloor\rceil

\let\originalleft\left
\let\originalright\right
\renewcommand{\left}{\mathopen{}\mathclose\bgroup\originalleft}
\renewcommand{\right}{\aftergroup\egroup\originalright}

\newcommand{\g}[1]{%
  \ifthenelse{\equal{#1}{(}}
  {\left( }%
    { \ifthenelse{\equal{#1}{)}}
      { \right)}%
    { \ifthenelse{\equal{#1}{[}}
      {\left[}%
        { \ifthenelse{\equal{#1}{]}}
          { \right]}%
        {#1}}
    }
  }
}

\usepackage[table]{xcolor}
\definecolor{ourblue}{rgb}{0.368,0.507,0.71}
\definecolor{ourorange}{rgb}{0.881,0.611,0.142}
\definecolor{ourgreen}{rgb}{0.56,0.692,0.195}
\definecolor{ourred}{rgb}{0.923,0.386,0.209}
\definecolor{ourviolet}{rgb}{0.528,0.471,0.701}
\definecolor{ourbrown}{rgb}{0.772,0.432,0.102}
\definecolor{ourlightblue}{rgb}{0.364,0.619,0.782}
\definecolor{ourdarkgreen}{rgb}{0.572,0.586,0.}
\definecolor{ourdarkblue}{RGB}{28,99,148}
\definecolor{ourdarkred}{RGB}{169,53,17}

\usepackage{etoolbox}
\makeatletter
\patchcmd{\NAT@test}{\else \NAT@nm}{\else \NAT@nmfmt{\NAT@nm}}{}{}
\DeclareRobustCommand\citeposs
  {\begingroup
   \let\NAT@nmfmt\NAT@posfmt%
   \NAT@swafalse\let\NAT@ctype\z@\NAT@partrue
   \@ifstar{\NAT@fulltrue\NAT@citetp}{\NAT@fullfalse\NAT@citetp}}

\let\NAT@orig@nmfmt\NAT@nmfmt
\def\NAT@posfmt#1{\NAT@orig@nmfmt{#1's}}
\makeatother

\usepackage{xr-hyper}

\makeatletter
\newcommand*{\addFileDependency}[1]{%
  \typeout{(#1)}
  \@addtofilelist{#1}
  \IfFileExists{#1}{}{\typeout{No file #1.}}
}
\makeatother

\usepackage[utf8]{inputenc} %
\usepackage[T1]{fontenc}    %
\usepackage{hyperref}       %
\usepackage{url}            %
\usepackage{nicefrac}       %
\usepackage{microtype}      %
\usepackage{xcolor}         %
\usepackage{DejaVuSans}
\usepackage{subcaption}
\usepackage{svg}
\usepackage{wrapfig}
\usepackage{bbold}
\usepackage{algorithm}
\usepackage{algorithmicx}
\usepackage[noend]{algpseudocode}
\usepackage{stmaryrd}
\usepackage[bottom]{footmisc}
\usepackage{cleveref}

\definecolor{ours}{HTML}{8778b3}
\definecolor{ceeus}{HTML}{8fb030}
\definecolor{baseline1}{HTML}{5c7fb3}
\definecolor{baseline2}{HTML}{e19c23}
\definecolor{baseline3}{HTML}{eb6133}
\definecolor{baseline1darker}{HTML}{4d71a6}

\definecolor{ours_test_pure}{HTML}{86a0c7}
\definecolor{ours_test_pure_xyz}{HTML}{384e6d}
\definecolor{cee_us_test}{HTML}{eb6235}
\definecolor{ours_test_dis}{HTML}{ba6d73}

\hypersetup{
	colorlinks=true,       %
	linkcolor=baseline1darker,        %
	citecolor=baseline1darker,        %
	filecolor=baseline1darker,     %
	urlcolor=baseline1darker
}

\renewcommand{\figurename}{\bf Figure}
\renewcommand{\tablename}{\bf Table}

\newcommand{\ourMethod}{RaIR\xspace}
\newcommand{\ourMethodA}{\mbox{RaIR\,+\,CEE-US}\xspace}
\newcommand{\oldMethod}{CEE-US\xspace}
\newcommand{\obj}{\textrm{obj}}

\newcommand{\FPP}{\textsc{Construction}\xspace}
\newcommand{\Grid}{\textsc{ShapeGridWorld}\xspace}

\title{Regularity as Intrinsic Reward for Free Play}

\author{
Cansu Sancaktar$^{1}$ \quad Justus Piater$^{2}$ \quad Georg Martius$^{1,3}$ \\
\shortstack{$^1$Max Planck Institute for \\ Intelligent Systems, Germany}
\quad \shortstack{$^2$University of Innsbruck \\ Austria}
 \quad  \shortstack{$^3$University of T\"ubingen \\  Germany}
\\
\texttt{\ cansu.sancaktar@tuebingen.mpg.de}
}

\begin{document}

\maketitle

\setcounter{footnote}{0} 

\begin{abstract}
    We propose regularity as a novel reward signal for intrinsically-motivated reinforcement learning. Taking inspiration from child development, we postulate that striving for structure and order helps guide exploration towards a subspace of tasks that are not favored by naive uncertainty-based intrinsic rewards. Our generalized formulation of Regularity as Intrinsic Reward (RaIR) allows us to operationalize it within model-based reinforcement learning. In a synthetic environment, we showcase the plethora of structured patterns that can emerge from pursuing this regularity objective. We also demonstrate the strength of our method in a multi-object robotic manipulation environment. We incorporate RaIR into free play and use it to complement the model’s epistemic uncertainty as an intrinsic reward. Doing so, we witness the autonomous construction of towers and other regular structures during free play, which leads to a substantial improvement in zero-shot downstream task performance on assembly tasks. Code and videos are available at \url{https://sites.google.com/view/rair-project}.%
\end{abstract}

\begin{figure}[ht]
  \centering
    \begin{subfigure}[b]{0.137\textwidth}
         \centering
         \includegraphics[width=\textwidth]{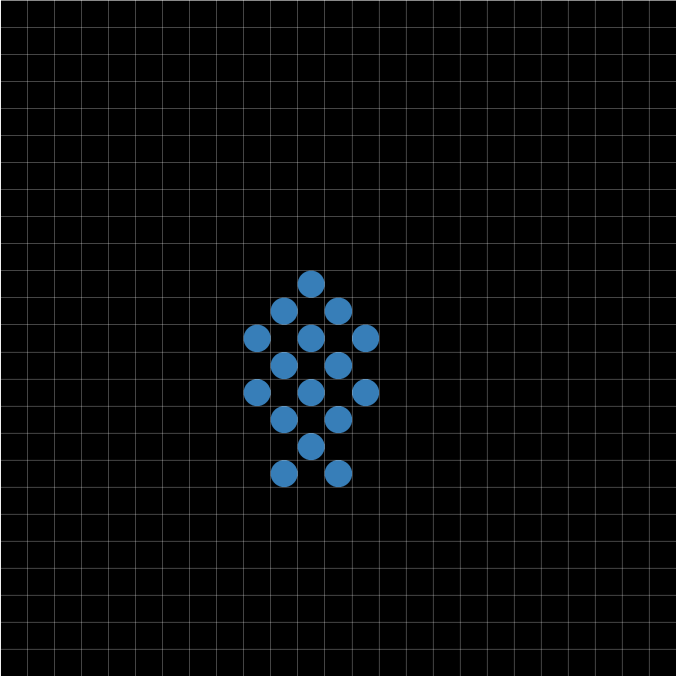}
     \end{subfigure}
      \begin{subfigure}[b]{0.137\textwidth}
         \centering
         \includegraphics[width=\textwidth]{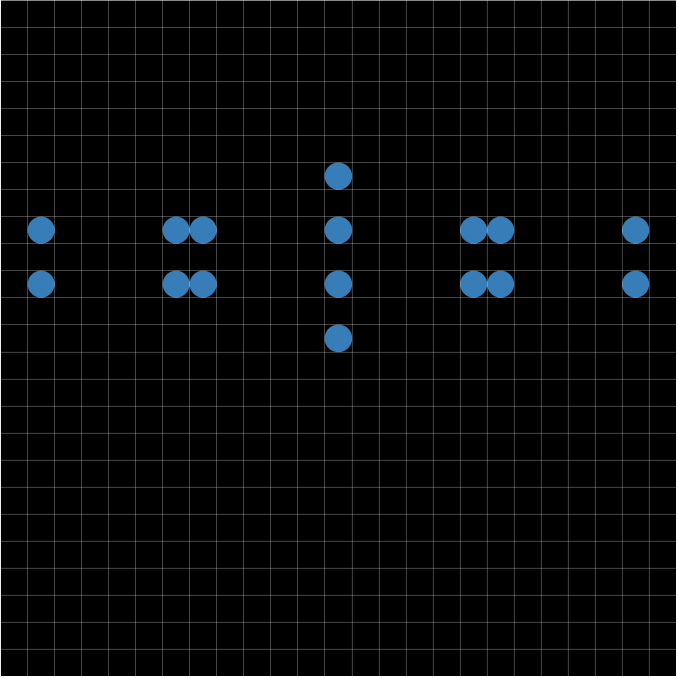}
     \end{subfigure}
 \begin{subfigure}[b]{0.137\textwidth}
         \centering
         \includegraphics[width=\textwidth]{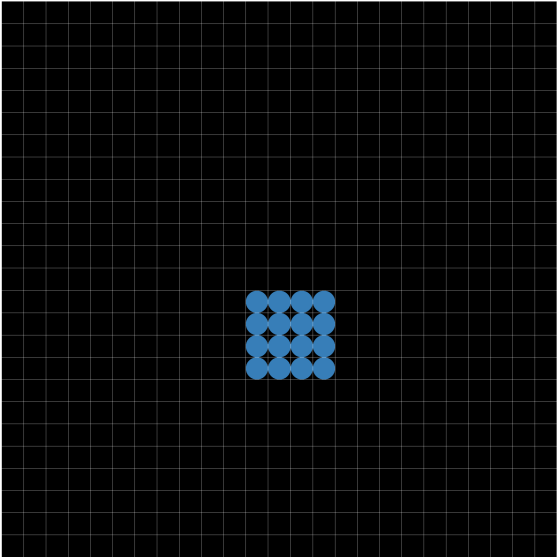}
     \end{subfigure}
    \begin{subfigure}[b]{0.137\textwidth}
         \centering
         \includegraphics[width=\textwidth]{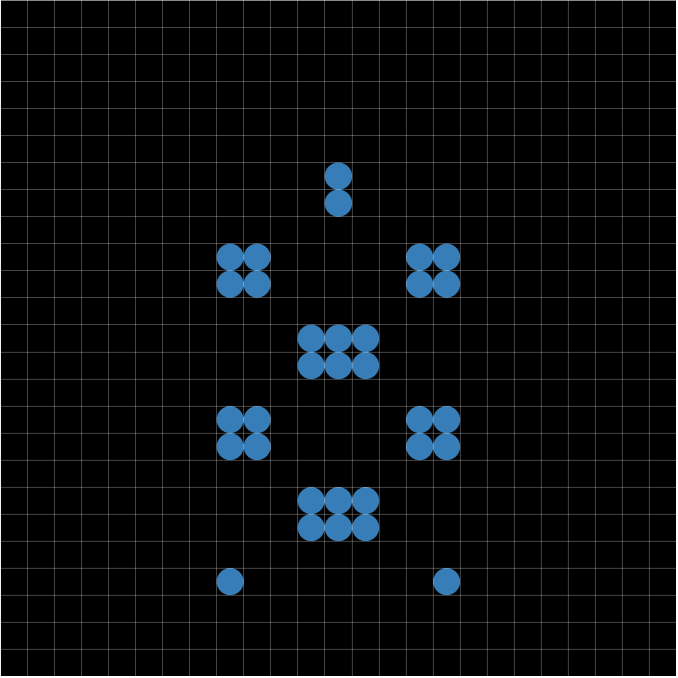}
     \end{subfigure}
     \begin{subfigure}[b]{0.137\textwidth}
        \centering
         \includegraphics[width=\textwidth]{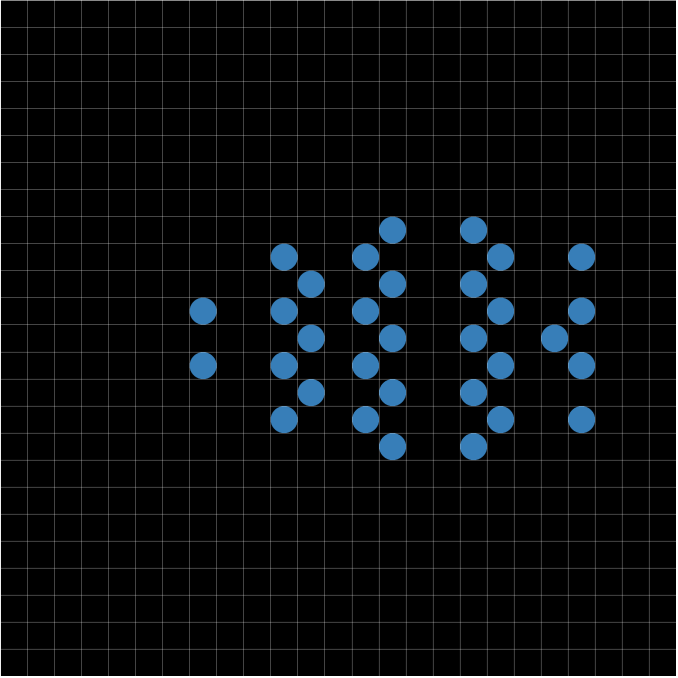}
     \end{subfigure}
     \begin{subfigure}[b]{0.137\textwidth}
         \centering
              \includegraphics[width=\textwidth]{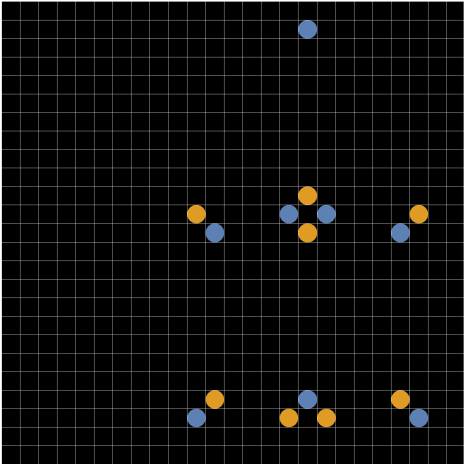}
     \end{subfigure}
     \begin{subfigure}[b]{0.137\textwidth}
         \centering
         \includegraphics[width=\textwidth]{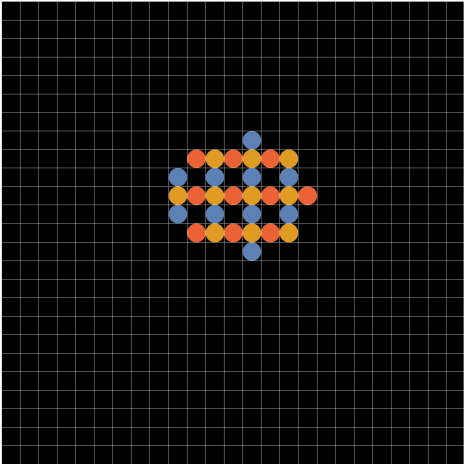}
     \end{subfigure}
     \\
     \vspace{.2em}
         \begin{subfigure}[b]{0.1375\textwidth}
         \centering
        \includegraphics[width=\textwidth]{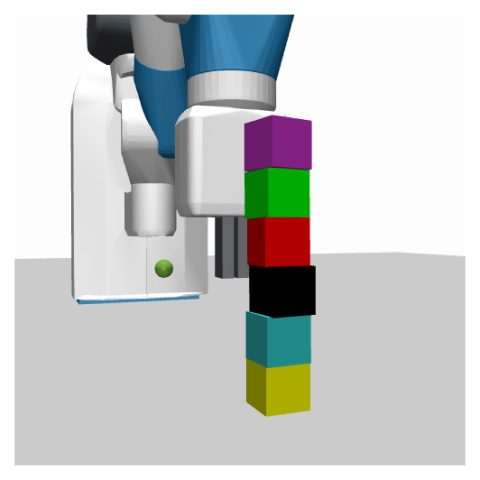}
     \end{subfigure}
      \begin{subfigure}[b]{0.1375\textwidth}
         \centering
        \includegraphics[width=\textwidth]{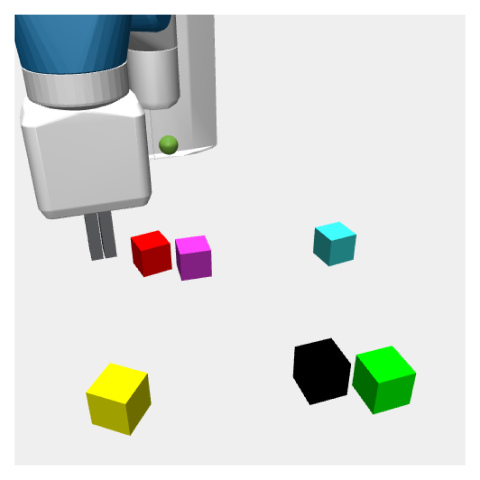}
     \end{subfigure}
     \begin{subfigure}[b]{0.1375\textwidth}
         \centering
        \includegraphics[width=\textwidth]{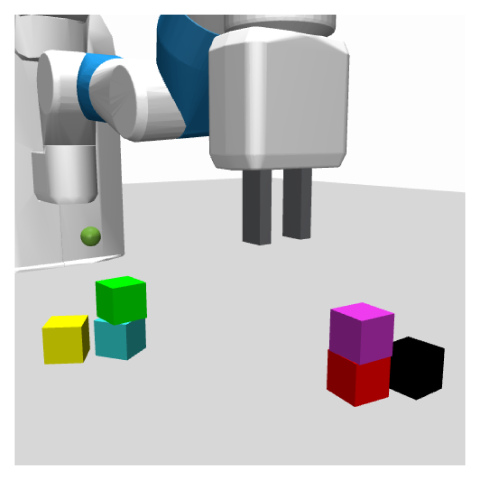}
     \end{subfigure}
 \begin{subfigure}[b]{0.1375\textwidth}
         \centering
        \includegraphics[width=\textwidth]{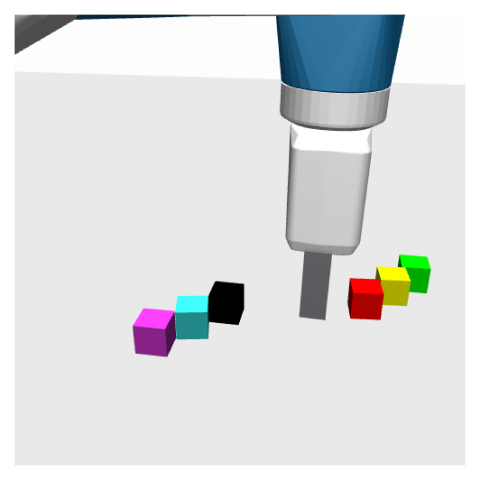}
     \end{subfigure}
    \begin{subfigure}[b]{0.1375\textwidth}
         \centering
        \includegraphics[width=\textwidth]{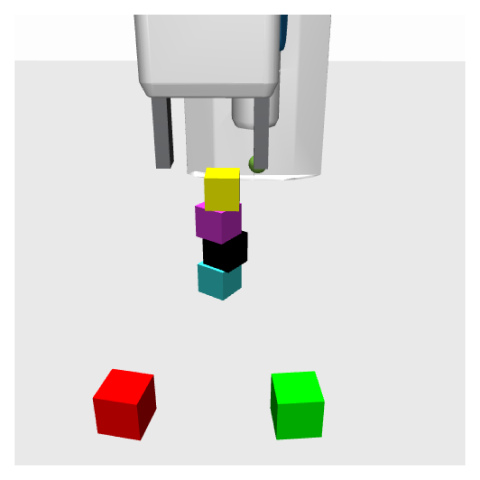}
     \end{subfigure}
    \begin{subfigure}[b]{0.1375\textwidth}
        \centering
        \includegraphics[width=\textwidth]{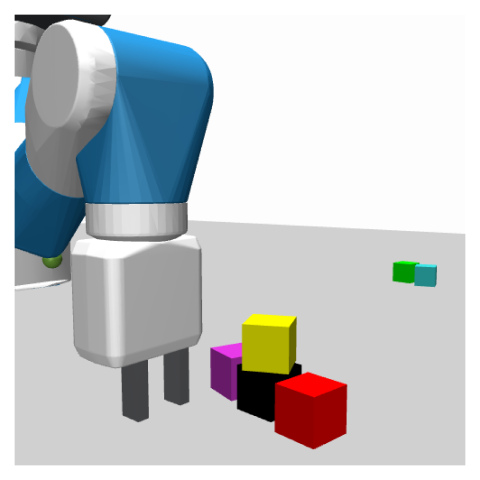}
     \end{subfigure}
     \begin{subfigure}[b]{0.1375\textwidth}
         \centering
        \includegraphics[width=\textwidth]{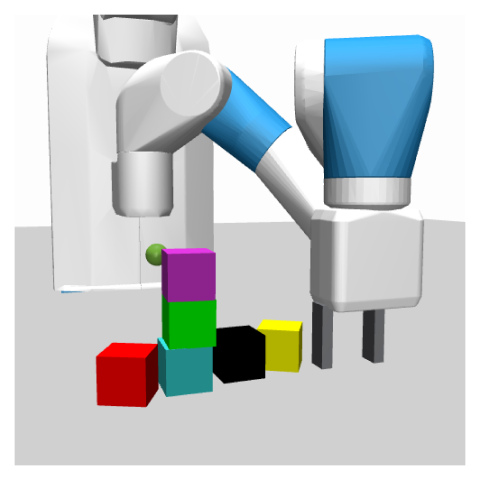}
     \end{subfigure}\vspace{-.3em}
  \caption{{\bf{Regularity as intrinsic reward yields ordered and symmetric patterns.}} In \Grid (top row) and in \FPP (bottom row), we showcase the generated constellations when maximizing our proposed regularity reward \ourMethod with ground truth (GT) models.}
       \label{fig:envs}
\end{figure}

\section{Introduction} \label{sec:introduction}

Regularity, and symmetry as a specific form of regularity, are ubiquitous in nature as well as in our manufactured world.
The ability to detect regularity helps to identify essential structures, minimizing redundancy and allowing for efficient interaction with the  world \cite{liu2010computational}.
Not only do we encounter symmetries in arts, design, and architecture, but our preference also showcases itself in play behavior. Adults and children have both been observed to prefer symmetry in visual perception, where symmetric patterns are more easily detected, memorized, and copied \cite{attneave1955symmetry, bornstein1981perception}.
Several works in developmental psychology show that regular patterns and symmetries are actively sought out during free play in children as well as adults \cite{bailey1933scale, zingrone2014construction, golomb1987development}.

Considering this in the context of a child's developmental cycle is intriguing.
Studies show that children at the age of 2 exhibit a shift in their exploratory behavior.
They progress from engaging in random actions on objects and unstable arrangements to purposefully engaging in functional activities and intentionally constructing stable configurations \cite{gibson1988exploratory, langer1986origins, zingrone2014construction}.
\citet{bailey1933scale} reports that by 5 years of age, children build more structured arrangements out of blocks that exhibit alignment, balance, and examples of symmetries \cite{zingrone2014construction}.

Despite the dominance of regularity in our perceptual systems and our preference for balance and stability during play, these principles are not yet well investigated within intrinsically-motivated reinforcement learning (RL).
One prominent intrinsic reward definition is novelty, \ie the agent is incentivized to visit areas of the state space with high expected information gain \cite{pathak2017curiosity, pathak2019self, sekar2020planning, Sancaktaretal22}.
However, one fundamental problem with plain novelty-seeking objectives is that the search space is often unconstrained and too large. As an agent only has limited resources to allocate during play time, injecting appropriate inductive biases is crucial for sample efficiency, good coverage during exploration, and emergence of diverse behaviors. As proposed by \citet{Sancaktaretal22}, using structured world models to inject a relational bias into exploration, %
yields more object and interaction-related novelty signals.
However, which types of information to prioritize are not explicitly encoded in any of these methods. The direction of exploration is often determined by the inherent biases in the practical methods deployed.
With imperfect world models that have a limited learning capacity and finite-horizon planning, novelty-seeking methods are observed to prefer ``chaotic'' dynamics, where small perturbations lead to diverging trajectories, such as throwing, flipping, and poking objects.
This in turn means that behaviors focusing on alignment, balance, and stability are overlooked. Not only are these behaviors relevant, as shown in developmental psychology,  %
they also enable expanding and diversifying the types of behavior uncovered during exploration.
As the behaviors observed during exploration are highly relevant for being able to solve  downstream tasks,
a chaos-favoring exploration will make it hard to solve assembly tasks, such as stacking.
Indeed, successfully solving assembly tasks with more than 2 objects has been a challenge for intrinsically-motivated reinforcement learning.

We pose the question:  how can we define an intrinsic reward signal such that RL agents prefer structured and regular patterns?
We propose \ourMethod{}: \textbf{R}egularity \textbf{a}s \textbf{I}ntrinsic \textbf{R}eward, which aims to achieve highly ordered states.
Mathematically, we operationalize this idea using entropy minimization of a suitable state description.
Entropy and symmetries have been linked before \cite{lin1996correlation,lazarev2022information},
however, we follow a general notion of regularity, \ie where patterns reoccur and thus their description exhibits high redundancy / low entropy. In this sense, symmetries are a consequence of being ordered \citep{bormashenko2020entropy}. Regularity also means that the description is compressible, which is an alternative formulation.
As argued by \citet{schmidhuber2009:compression}, aiming for compression-progress is a formidable curiosity signal, however, it is currently unclear how to efficiently predict and optimize for it. %

After studying the design choices in the mathematical formulation of regularity and the relation to symmetry operations,
we set out to evaluate our regularity measure in the context of model-based reinforcement learning/planning,
as it allows for highly sample-efficient exploration and solving complex tasks zero-shot, as shown in \citet{Sancaktaretal22}.
To get a clear understanding of \ourMethod, we first investigate the generated structures when directly planning to optimize it using the ground truth system dynamics. A plethora of patterns emerge that are highly \emph{regular}, as illustrated in \fig{fig:envs}. %

Our ultimate goal is, however, to inject the proposed regularity objective into a free-play phase, where a robot can explore its capabilities in a task-free setting. During this free play, the dynamics model is learned on-the-go.
We build on \oldMethod \cite{Sancaktaretal22}, a free-play method that uses an ensemble of graph neural networks as a structured world model and
the model's epistemic uncertainty as the only intrinsic reward.
The epistemic uncertainty is estimated by the ensemble disagreement and acts as an effective novelty-seeking signal.
We obtain \emph{structure-seeking free play} by combining the conventional novelty-seeking objective with \ourMethod.

Our goal is to operationalize regularity, which is a well-established concept in developmental psychology, within intrinsically motivated RL.
Furthermore, we showcase that biasing information-search towards regularity with \ourMethod indeed leads to the construction of diverse regular structures during play and significantly improves zero-shot performance in downstream tasks that also favor regularity, most notably assembly tasks.
Besides conceptual work on compression \cite{schmidhuber2009:compression,schmidhuber2013powerplay}, to our knowledge, we are the first to investigate regularity as an intrinsic reward signal, bridging the gap between the diversity of behaviors observed in children's free play and what we can achieve with artificial agents.

\begin{figure}
  \centering
      \begin{subfigure}[b]{0.73\textwidth}
         \centering
  	\includegraphics[width=\textwidth]{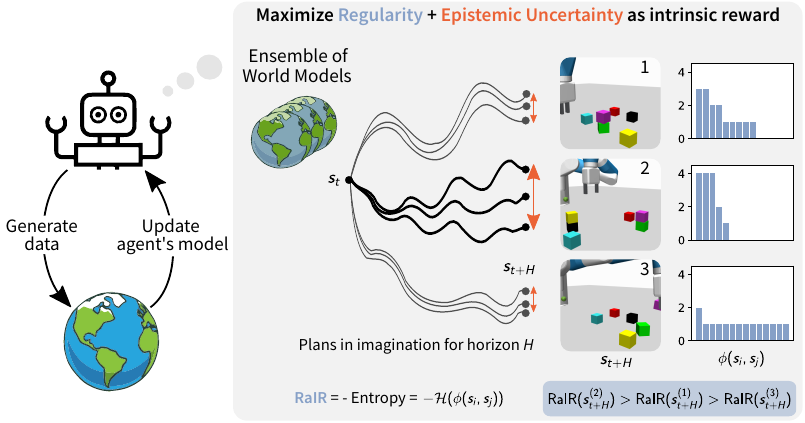}
        \subcaption{\scriptsize \ourMethodA: Regularity with Epistemic Uncertainty of Structured World Models}\label{fig:overview:setup}
     \end{subfigure}%
    \begin{subfigure}[b]{0.25\textwidth}
         \centering
  	\includegraphics[width=.7\textwidth]{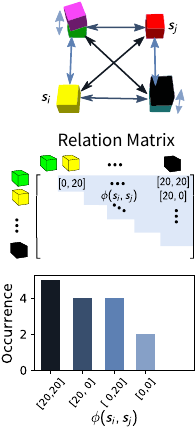}
        \subcaption{\scriptsize \ourMethod computation}\label{fig:overview:rair}
     \end{subfigure}
       \caption{{\bf Regularity as intrinsic reward during free play.}
        (a) \ourMethodA uses model-based planning to optimize $H$ timesteps into the future for the combination of \ourMethod \eqnp{eqn:rair} and epistemic uncertainty (ensemble disagreement of world models).
        (b) Here, for \ourMethod we use the absolute difference vector between objects:
        $\phi(s_i,s_j) = \{(|\nint{s_{i,x} - s_{j,x}}|, |\nint{s_{i,y} - s_{j,y}}|)\}$.}
\label{fig:overview}
\end{figure}

\section{Method}
First, we introduce our intrinsic reward definition for regularity. Then, we present its practical implementation and explain how we combine this regularity objective into model learning within free play.\looseness-1

\subsection{Preliminaries} \label{sec:preliminaries}
In this work, we consider environments that can be described by a fully observable Markov Decision Process (MDP),
 given by the tuple \( (\mathcal{S}, \mathcal{A}, f_{ss^{\prime}}^{a}, r_{ss^{\prime}}^{a}) \), with the state-space \( \mathcal{S} \in \mathbb{R}^{n_{s}} \), the action-space \( \mathcal{A} \in \mathbb{R}^{n_{a}} \), the transition kernel \( f: \mathcal{S} \times \mathcal{A} \xrightarrow{} \mathcal{S} \), and the reward function \( r \). %
Importantly, we consider the state-space to be factorized into the different entities, \eg $\mathcal{S} = (\mathcal{S}_{\obj})^N \times \mathcal{S}_{\textrm{robot}}$ for the state space of a robotic agent and $N$ objects. %
We use model-based reinforcement learning, where data from interactions with the environment is used to learn a model $\tilde{f}$ of the MDP dynamics \cite{chua2018}.
Using this model, we consider finite-horizon ($H$) optimization/planning for undiscounted cumulative reward:\looseness-1
\begin{equation}
    \mathbf{a}_t^{\star} = \argmax_{\mathbf{a}_t} \sum_{h=0}^{H-1} r(s_{t+h}, a_{t+h}, s_{t+h+1}), \label{eqn:action_planner}
\end{equation}
where \(s_{t+h}\) are imagined states visited by rolling out the actions using \(\tilde f\), which is assumed to be deterministic.
The optimization of \eqn{eqn:action_planner} is done with the improved Cross-Entropy Method (iCEM)~\cite{pinneri2020:iCEM}
 in a model predictive control (MPC) loop, \ie re-planning after every step in the environment.
Although this is not solving the full reinforcement learning problem (infinite horizon and stochastic environments),
it is very powerful in optimizing for tasks on-the-fly and is thus suitable for optimizing changing exploration targets and solving downstream tasks zero-shot.

\subsection{Regularity as Intrinsic Reward}\label{sec:rair_intro}
Quite generally, regularity refers to the situation in which certain patterns reoccur.
Thus, we formalize regularity as the {\bf redundancy} in the description of the situation, to measure the degree of sub-structure recurrence.
A decisive question is: which description should we use?
Naturally, there is certain freedom in this choice, as there are many different coordinate frames.
For instance, we could consider the list of absolute object positions or rather a relative representation of the scene.

To formalize, we define a mapping $\Phi: \mathcal S \to \{\mathcal X\}^+$ from state to a multiset $\{\mathcal X\}^+$ of symbols (\eg coordinates).
A multiset is a set where elements can occur multiple times, \eg $\{a, a, b\}^+$. This multiset can equivalently be described by a tuple $(X, m)$, where $X$ is the set of the unique elements, and $m: X \to \mathbb Z^+$ is a function assigning the multiplicity, \ie the number of occurrences $m(x)$ for the elements $x \in X$. For the previous example, we get $(\{a, b\},\{a:2, b:1\})$.
Given the multiset $(X,m) \in \{\mathcal X\}^+$, we define the discrete empirical distribution, also referred to as a histogram, 
by the relative frequency of occurrence $p(x) = m(x)/\sum_{x'\in X} m(x')$ for $x\in X$. 

We define the regularity reward metric using (negative) Shannon entropy \cite{shannon1948mathematical} of this distribution as:
\begin{align}
    r_{\mathrm{\ourMethod}}(s) \coloneqq -\mathcal{H}(\Phi(s)) &= \!\!\sum_{x \in X} p(x) \log p(x) & \text{with } (X,m) = \Phi(s),\quad  p(x) = \frac{m(x)}{\sum\limits_{x'\in X} m(x')}. \label{eqn:rair}
\end{align}
We will now discuss concrete cases for the mapping $\Phi$, \ie how to describe a particular state.

\paragraph{Direct \ourMethod.}

In the simplest case, we describe the state $s$ directly by the properties of each of the entities.
For that, we define the function $\phi: \mathcal S_\obj \to \{\mathcal X\}^+$, that maps each entity to a set of symbols and obtain $\Phi(s) = \Cup_{i=1}^N \phi(s_{\obj,i})$ as a union of all symbols.
The symbols can be, for instance, discretized coordinates, colors, or other properties of the entities.

\begin{wrapfigure}[9]{r}{.3\linewidth}
    \vspace*{-1em}
    \includegraphics[width=\linewidth]{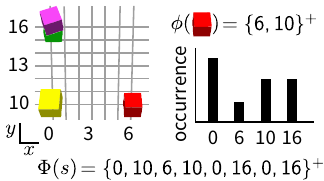}
    \caption{Illustration of direct
 RaIR for
    $\phi = \{\nint{x}, \nint{y}\}$.}\label{fig:rair:unilateral}
\end{wrapfigure}
Let us consider the example where $\phi$ is extracting the object's Cartesian $x$ and $y$ coordinates in a rounded manner as
$\phi(s) = \{\nint{s_x}, \nint{s_y}\}^+$, as shown in \fig{fig:rair:unilateral}.
The most irregular configuration would be when no two objects share the same rounded value in $x$ and $y$.
The object configuration becomes more and more regular the more objects share the same $\nint{x}$ and $\nint{y}$ coordinates. The most regular configuration is if all objects are in the same place.
Note that this choice favors an axis-aligned configuration, and it is not invariant under global rotations.

\paragraph{Relational \ourMethod of order $k$.}
Our framework for regularity quantification can easily be extended to a relational perspective, where we don't compute the entropy over aspects of individual entity properties, but instead on their pairwise or higher-order \textbf{relations}. This means that for a $k$-order regularity measure, we are interested in tuples of $k$ entities. Thus, the mapping function $\phi$ no longer takes single entities as input, but instead operates on $k$-tuples:
\begin{equation}
\phi: (\mathcal{S}_{\obj})^k \xrightarrow{} \{\mathcal X\}^+.
\end{equation}
$\phi$ is a function that describes some relations between the $k$ input entities by a set of symbols.

For $k$-order regularity, the multiset $\Phi$, over which we compute the entropy, is now given by
\begin{align}
    \Phi^{(k)} = \!\!\bigcup_{\{i_1, \dots, i_k\} \in \mathcal{P}}\!\! \phi(s_{\obj,i_1}, \dots, s_{\obj,i_k})  & \quad \text{with } \mathcal{P}=\operatorname{P}(\{1, \dots, N\}, k) \label{eqn:rair:general}
\end{align}
merged from all $k$-permutations of the $N$ entities, denoted as $\operatorname{P}(\{1, \dots, N\}, k)$. 
In the case of a permutation invariant $\phi$, \eqn{eqn:rair:general} regards only the combinations $\operatorname{C}(\{1, \dots, N\},k)$.
Note that direct \ourMethod is equivalent to the relational \ourMethod of order $1$.
Given the mapping $\Phi$, the \ourMethod measure is computed as before with \eqn{eqn:rair}.

Depending on the order $k$ and the function $\phi$, we can select which regularities are going to be favored.
Let us consider the example of a pairwise relational \ourMethod ($k=2$), where $\phi$ computes the relative positions: $\phi(s_i,s_j) = \{\nint{s_{i} - s_{j}}\}$, and rounding is performed elementwise.
Whenever two entities have the same relative position to each other, the redundancy is detected.
For $k=3$ our regularity measure would be able to pick up sub-patterns composed of three objects, such as triangles and so forth.\looseness-1

As we are interested in physical interactions of the robot with objects and objects with objects, we choose \ourMethod of order $k=2$ and explore various $\phi$ functions.

\newlength{\picw}
\setlength{\picw}{0.11\linewidth}

\begin{table}[htb]
    \caption{{\bf Properties of \ourMethod with different $\phi$ regarding symmetry operations}.
    The first block indicates to which operations \ourMethod is invariant, ignoring rounding (a.a.: axes aligned). The second block assesses whether a pattern, where the given symmetry operation maps several entities to another, has increased regularity. Rounding and absolute value are elementwise. Distance $d$ is also rounded.}
    \label{tab:symmetry}
    \centering
    \vspace{.3em}
    \def\arraystretch{1.25}
    \begin{tabular}{@{}l@{\ }c@{\ \ }c@{\ \ }c@{\ \ }c|c@{\ \ }c@{\ \ }c@{\ \ }l@{\hspace{-1em}}l@{}}
        \toprule
                  & \multicolumn{4}{c|}{invariant?}  & \multicolumn{4}{c}{favored / increases \ourMethod ?}\\[-.25em]

       symmetry     & direct   & rel. pos & $|$rel. pos$|$ & distance&  direct &rel. pos & $|$rel. pos$|$ & distance\\[-0.25em]
        operation  & \hspace*{-1.5em}\scriptsize$\phi=\nint{s_i}$ & \scriptsize$\nint{s_i-s_j}$ & \scriptsize$|\nint{s_i\!-\!s_j}|$ & \scriptsize$\operatorname{d}(s_i,s_j)$ & \scriptsize$\nint{s_i}$ & \scriptsize$\nint{s_i-s_j}$ & \scriptsize$|\nint{s_i\!-\!s_j}|$ & \scriptsize$\operatorname{d}(s_i,s_j)$&\multicolumn{1}{r@{\ }}{\includegraphics[width=3em]{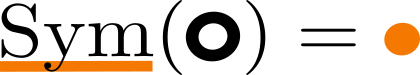}}\\[-0.15em]
        \midrule
        translation           & \xmark   & \cmark   & \cmark         & \cmark  &  \xmark   &\cmark   & \cmark         & \cmark & \multirow{2}{*}{\includegraphics[width=\picw]{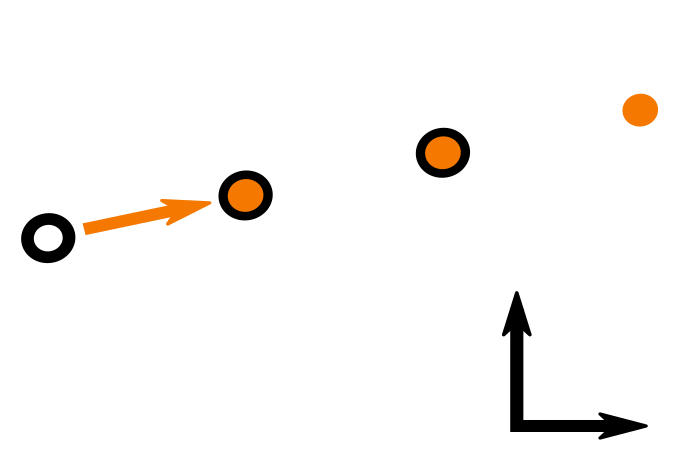}}\\
        translation -- a.a.    & \cmark   & \cmark   & \cmark         & \cmark  &  \cmark   &\cmark   & \cmark         & \cmark\\\greyrule
        rotation              & \xmark   & \cmark   & \xmark         & \cmark  &  \xmark   &\xmark   & \xmark         & \cmark & \multirow{2}{*}{\includegraphics[width=\picw]{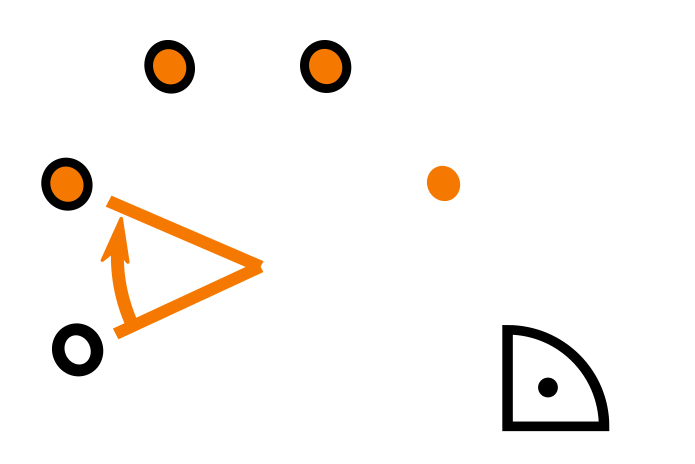}}\\
        rotation -- $90^\circ$ & \cmark   & \cmark   & \cmark         & \cmark  &  \xmark   &\xmark   & \cmark         & \cmark\\\greyrule
        reflection            & \xmark   & \cmark   & \xmark         & \cmark  &  \xmark   &\xmark   & \xmark         & \cmark & \multirow{2}{*}{\includegraphics[width=\picw]{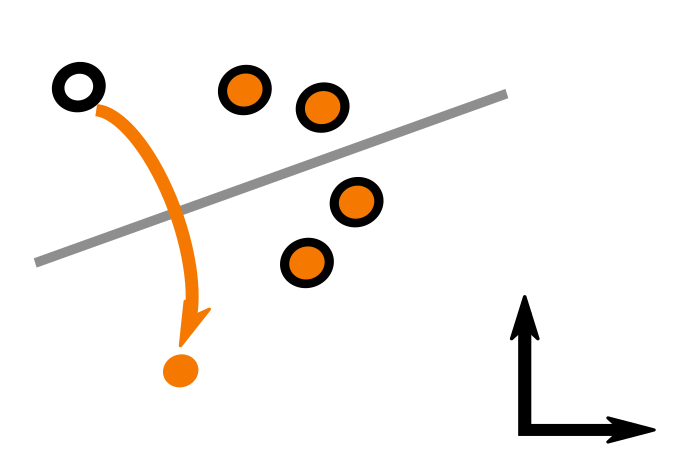}}\\
        reflection -- a.a.     & \cmark   & \cmark   & \cmark         & \cmark  &  \cmark   &\xmark   & \cmark         & \cmark\\\greyrule
        glide refl.           & \xmark   & \cmark   & \xmark         & \cmark  &  \xmark   &\phantom{$^{(1)}$}\xmark$^{(1)}$   & \phantom{$^{(1)}$}\xmark$^{(1)}$         & \cmark & \multirow{2}{*}{\includegraphics[width=\picw]{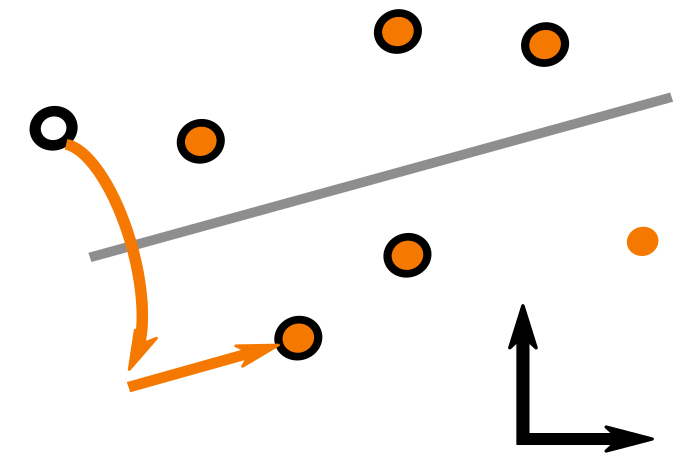}}\\
        glide refl. -- a.a.   & \cmark   & \cmark   & \cmark         & \cmark  &  \cmark   &\phantom{$^{(1)}$}\xmark$^{(1)}$   & \cmark         & \cmark\\
        \bottomrule
    \end{tabular}
\footnotesize{$^{(1)}$This is for one glide refl. operation. \ourMethod is increased for 2 glide refl. composition as it collapses onto transl.}
\end{table}
\subsubsection{Properties of \ourMethod with Pairwise Relations and Practical Implementation} \label{sec:practical_rair}
For simplicity, we are considering in the following that $\phi$ maps to a single symbol. Then for pairwise relationships ($k=2$), \ourMethod can be implemented
using a relation matrix $F \in \mathcal{X}^{N \times N} $.
The entries $F_{ij}$ are given by $\phi(s_i,s_j)$ with $s_i, s_j \in S_{\obj}$.
After constructing the relation matrix, we need the histogram of occurrences of unique values in this matrix to compute the entropy \eqnp{eqn:rair}.
For continuous state spaces, the mapping function needs to implement a discretization step, which we implement by a binning of size $b$. For simplicity of notation, we reuse the rounding notation $\nint{\cdot}$ for this discretization step.
This bin size $b$ determines the precision of the measured regularity.
In practice, we do not apply $\phi$ on the full entity state space, but on a subspace that contains \eg the $x$-$y$(-$z$) positions.

To understand the properties of our regularity measure
for different $\phi$, we present in \tab{tab:symmetry}
a categorization using the known symmetry operations in 2D
 and the following $\phi$ (applied to $x$-$y$ positions):
 direct $\phi(s_i) = \nint{s_i}$(see previous section), relative position (difference vector) $\phi(s_i,s_j) =\nint{s_i - s_j}$, absolute value of the relative position\footnote{The rounding and the absolute value functions are applied coordinate-wise.} $\phi(s_i,s_j) =|\nint{s_i - s_j}|$, and Euclidean distance $\phi(s_i,s_j) =\nint{\|s_i - s_j\|}$.
\Fig{fig:overview:rair} illustrates the \ourMethod computation using the absolute value of the relative position.\looseness-1

In \tab{tab:symmetry}, we first consider whether the measure is invariant under symmetry operations. That means if the value of \ourMethod stays unchanged when the entire configuration is transformed.
We find that both Euclidean distance and relative position are invariant to all symmetry operations.
The second and possibly more important question is whether a configuration with sub-structures of that symmetry has a higher regularity value than without, \ie will patterns with these symmetries be favored.
We find that Euclidean distance favors all symmetries, followed by absolute value of the relative position.
A checkmark in this part of the table means that the more entities can be mapped to each other with the same transformation, the higher \ourMethod.
Although the Euclidean distance seems favorable, we find that it
mostly clumps entities and creates fewer alignments.
To get a sense of the patterns scoring high in the regularity measure, \fig{fig:envs} showcases situations that emerge when \ourMethod with absolute value of relative position is optimized (details below).\looseness-1

\subsection{Regularity in Free Play}

Our goal is to explicitly put the bias of regularity into free play via \ourMethod, as in \fig{fig:overview:setup}.
What we want to achieve is not just that the agent creates regularity, but that it gathers valuable experience in creating regularity. This ideally leads to directing exploration towards patterns/arrangements that are novel. %

We propose to use \ourMethod to augment plain novelty-seeking intrinsic rewards, in this work specifically ensemble disagreement~\cite{pathak2019self}. We choose ensemble disagreement because 1) we need a reward definition that allows us to predict future novelty, such that we can use it inside model-based planning (this constraint makes methods relying on retrospective novelty such as Intrinsic Curiosity Module (ICM) \cite{pathak2017curiosity} ineligible), and 2) we want to use the models learned during free play for zero-shot downstream task generalization via planning in a follow-up extrinsic phase.
It has been shown in previous works that guiding exploration by the model's own epistemic uncertainty, approximated via ensemble disagreement, leads to learning more robust world models compared to \eg Random Network Distillation (RND) \cite{burda2018exploration} (\supp{app:rnd}), resulting in improved zero-shot downstream task performance \cite{Sancaktaretal22}. That is why we choose ensemble disagreement to compute expected future novelty.\looseness-1

We train an ensemble of world models $\{(\tilde{f}{_{\theta}}_m)_{m=1}^M\}$, where $M$ denotes the ensemble size. The model's epistemic uncertainty is approximated by the disagreement of the ensemble members' predictions. The disagreement reward is given by the trace of the covariance matrix \cite{Sancaktaretal22}:
\begin{equation}
r_{\mathrm{Dis}} = \mathrm{tr}\big(\mathrm{Cov}(\{\hat{s}^m_{t+1}={\tilde{f}{_{\theta}}_m}(s_t, a_t) \mid m=1,\ldots,M\} )\big). \label{eqn:epistemic}
\end{equation}

We incorporate our regularity objective into free play by using a linear combination of \ourMethod and ensemble disagreement. Overall, we have the intrinsic reward:
\begin{equation}
    r_{\mathrm{intrinsic}} = r_{\mathrm{\ourMethod}} + \lambda \cdot r_{\mathrm{Dis}},
\label{eqn:combination}
\end{equation}
where $\lambda$ controls the trade-off between regularity and pure epistemic uncertainty.

\paragraph{Model-based Planning with Structured World Models}
To optimize the reward function on-the-fly, we use model-based planning using zero-order trajectory optimization, as introduced in \sec{sec:preliminaries}.
Concretely, we use \oldMethod \citep{Sancaktaretal22}, which combines structured world models and epistemic uncertainty \eqnp{eqn:epistemic} as intrinsic reward.
The structured world models are ensembles of message-passing Graph Neural Networks (GNNs) \cite{battaglia2018relational}, where each object corresponds to a node in the graph. The node attributes $\{s_{t,i} \in \mathcal{S}_{\obj} \mid i=1,\ldots,N\}$ are the object features such as position, orientation, and velocity at time step $t$. The state representation of the actuated agent $s_{\textrm{robot}} \in \mathcal{S}_{\textrm{robot}}$ similarly contains position and velocity information about the robot. We treat the robot as a global node in the graph \citep{Sancaktaretal22}. %
We refer to the combination of \ourMethod with ensemble disagreement, medium-horizon planning (20-30 time steps), and structured world models as \ourMethodA.

\section{Experiments}
We evaluate \ourMethod in the two environments shown in \fig{fig:envs}.

\textbf{ShapeGridWorld} is a grid environment, where each circle represents an entity/agent that is controlled separately in $x$-$y$ directions. Entities are controlled one at a time. Starting from timestep $t=0$, the entity with $i=1$ is actuated for $T$ timesteps, where $T$ is the entity persistency. Then, at $t=T$, actuation switches over to entity $i=2$ and we keep iterating over the entities in this fashion. Each circle is treated as an entity for the regularity computation with a 2D-entity state space $S_{\obj}$ with $x$-$y$ positions.\looseness-1

\textbf{Fetch Pick \& Place Construction} is an extension of the Fetch Pick \& Place environment~\cite{Plappert2018MultiGoalRL} to more cubes~\cite{li2020towards} and a large table \cite{Sancaktaretal22}. An end-effector-controlled robot arm is used to manipulate blocks.
The robot state $S_{\textrm{robot}} \in \mathbb{R}^{10}$ contains the end-effector position and velocity, and the gripper's state (open/close) and velocity.
Each object's state $S_{\obj} \in \mathbb{R}^{12}$ is given by its pose and velocities.
For free play, we use 6 objects and consider several downstream tasks with varying object numbers. %

\subsection{Emerging Patterns in \Grid and \FPP with \ourMethod{}}
\label{sec:emerging_gt}
To get a sense of what kinds of patterns emerge following our regularity objective with \ourMethod, we do planning using ground truth (GT) models, \ie with access to the true simulator itself for planning. We perform these experiments to showcase that we can indeed get \emph{regular} constellations with our proposed formulation. Since we can perform multi-horizon planning without any accumulating model errors using ground truth models, we can better investigate the global/local optima of our regularity reward. Note that as we are using a zero-order trajectory optimizer with a limited sample budget and finite-horizon planning, we don't necessarily converge to the global optima. We use $\phi(s_i,s_j) = \{(|\nint{s_{i,x} - s_{j,x}}|, |\nint{s_{i,y} - s_{j,y}}|)\}$ for \ourMethod in both environments. The emerging patterns are shown in \fig{fig:envs}.

In the 2D \Grid environment, we indeed observe that regular patterns with translational, reflectional (axis-aligned), glide-reflectional (axis-aligned), and rotational symmetries emerge. Regularity is not restricted to spatial relations and can be applied to any set of symbols. To showcase this, we apply \ourMethod to colored \Grid, where color is part of $S_{\text{obj}}$. Generated patterns are not just regular in $x$-$y$ but also in color, as shown in \fig{fig:colored_grid_snapshots} and the 2 top right examples in \fig{fig:envs}.\looseness-1

\begin{wrapfigure}[10]{r}{0.45\textwidth}
    \centering
    \includegraphics[width=\linewidth]{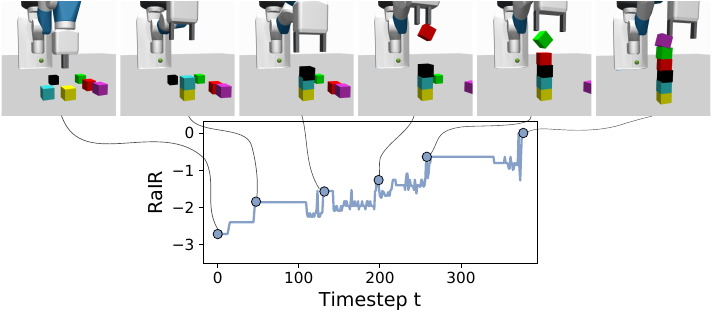}\vspace{-.65em}
    \caption{{\bf \ourMethod throughout a rollout} starting from a random initial configuration when optimizing only for regularity with the GT model.}
    \label{fig:rair_evolution}
\end{wrapfigure}
For \FPP, we also observe complex constellations with regularities, even stacks of all 6 objects.
Since we are computing \ourMethod on the $x$-$y$ positions, a stack of 6 is the global optimum. The optimization of \ourMethod for this case is shown in \fig{fig:rair_evolution}. Note that stacking itself is a very challenging task, and was so far only reliably achievable with reward shaping or tailored learning curricula \cite{li2020towards}. %
The fact that these constellations appear naturally from our regularity objective, achievable with a planning horizon of 30 timesteps, is by itself remarkable.\looseness-1

Additional example patterns generated in \FPP with \ourMethod on the $x$-$y$-$z$ positions can be found in the \supp{app:3dim_compression}. %
In that case, a horizontal line on the ground and a vertical line into air, \ie a stack, are numerically equivalent with respect to \ourMethod. Choosing to operate on the $x$-$y$-subspace is injecting the direction of gravity and provides a bias towards vertical alignments. 
We also apply \ourMethod to a custom \FPP environment with different shapes and masses (cubes, columns, balls and flat blocks) and once again observe regular arrangements, as in \fig{fig:custom_shapes}. More details in \supp{app:custom_fpp}.\looseness-1

\subsection{Free Play with \ourMethod in \FPP}\label{sec:free_play_w_rair}

We perform free play in \FPP, \ie only optimize for intrinsic rewards, where we learn models on-the-go. %
During free play, we start with randomly initialized models and an empty replay buffer. Each iteration of free play consists of data collection with environment interactions (via online planning), and then model training on the collected data so far (offline).
 
In each iteration of free play, we collect 2000 samples (20 rollouts with 100 timesteps each) and add them to the replay buffer. During the online planning part for data collection, we only perform inference with the models and no training is performed. Afterwards, we train the model for a fixed number of epochs on the replay buffer. We then continue with data collection in the next free play iteration. More details can be found in \supp{app:cee_us}.
For this intrinsic phase, we combine our regularity objective with ensemble disagreement as per \eqn{eqn:combination}. The goal is to bias exploration and the search for information gain towards regular structures, corresponding to the optima that emerge with ground truth models, as shown in \fig{fig:envs}. We also show results for the baselines RND~\cite{burda2018exploration} and Disagreement (Dis)~\cite{pathak2019self}, using the same model-based planning backbone as CEE-US (see \supp{app:baselines_detail}). The Dis baseline also uses ensemble disagreement as intrinsic reward, however unlike \oldMethod, only plans for one step into the future.\looseness-1

\begin{figure}[b]
    \centering
    {\scriptsize
      \textcolor{ours_test_pure}{\rule[2pt]{20pt}{1.2pt}} \ourMethod{}
      \quad \textcolor{ours}{\rule[2pt]{20pt}{1.2pt}} \ourMethodA{} %
       \quad \textcolor{cee_us_test}{\rule[2pt]{20pt}{1.2pt}} \oldMethod
      \quad \textcolor{ourorange}{\rule[2pt]{20pt}{1.2pt}} RND
      \quad \textcolor{ourgreen}{\rule[2pt]{20pt}{1.2pt}} Dis
    }\vspace{.1em}\\
    \begin{subfigure}[b]{0.01\textwidth}
        \notsotiny{\rotatebox{90}{\hspace{.5cm}\textsf{relative time}}}
    \end{subfigure}
    \begin{subfigure}[t]{.24\linewidth}
        \includegraphics[width=\linewidth]{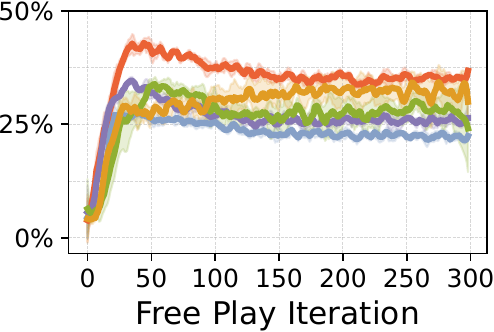}\vspace{-.3em}
    \caption{1 object moves}
    \end{subfigure}
    \hfill
    \begin{subfigure}[t]{.24\linewidth}
    \includegraphics[width=\linewidth]{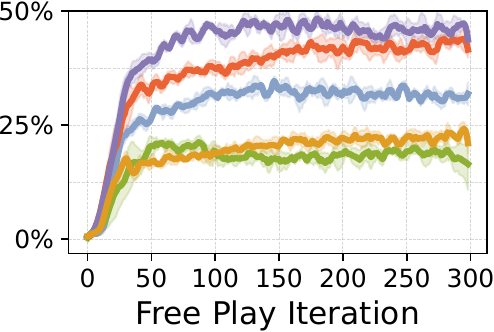}\vspace{-.3em}
    \caption{2 or more objects move}
    \end{subfigure}\hfill
    \begin{subfigure}[t]{.24\linewidth}
        \includegraphics[width=\linewidth]{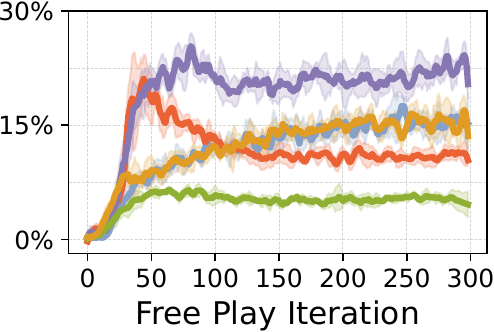}\vspace{-.3em}
    \caption{object(s) in air}
    \end{subfigure}
    \begin{subfigure}[t]{.24\linewidth}
        \includegraphics[width=\linewidth]{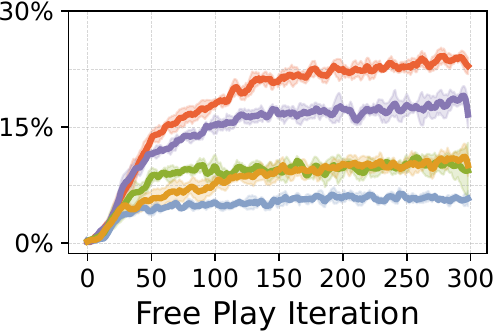}\vspace{-.3em}
    \caption{object(s) flipped}
    \end{subfigure}
    \vspace{-.5em}
    \caption{{\bf Comparison of interactions during free play in \FPP when combining ensemble disagreement with \ourMethod (with $\lambda$ = 0.1) compared to \oldMethod,  pure \ourMethod, RND and Dis.}
     These metrics count the relative amount of timesteps the agent performs certain types of interactions in the 2K transitions collected at each free play iteration. 
     (a) \textit{1 object moves} checks the amount of time the agent spends moving only one object. 
     (b) \textit{2 or more objects move} checks if at least 2 objects are moving at the same time. (c) \textit{Object(s) in air} means one or more objects are in air (including being held in air by the agent or being on top of another block). (d) \textit{Object(s) flipped} checks for angular velocities above a threshold for one or more objects, i.e. if they are rolled/flipped.
     We used 5 independent seeds. \looseness-1
    }
    \label{fig:interactions:FPP}
\end{figure}

\begin{figure}[t]
  \centering
    \begin{subfigure}[b]{0.136\textwidth}
         \centering
         \includegraphics[width=\textwidth]{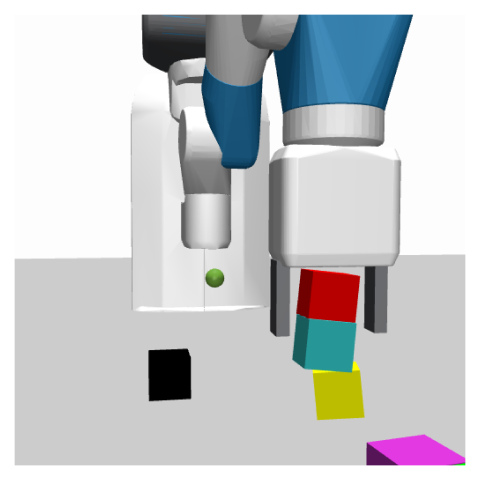}
        \centering{\scriptsize Iteration 251}
     \end{subfigure}
      \begin{subfigure}[b]{0.136\textwidth}
         \centering
         \includegraphics[width=\textwidth]{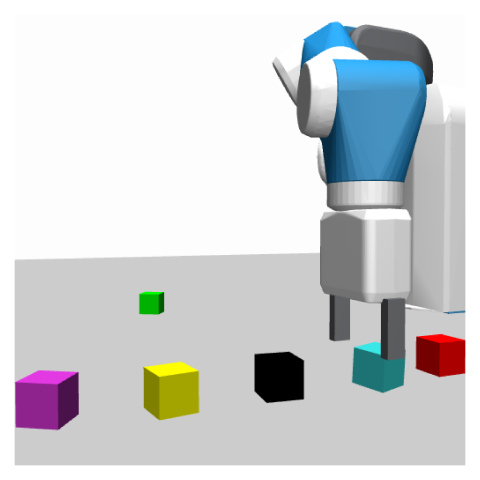}
        \centering{\scriptsize Iteration 255}
     \end{subfigure}
     \begin{subfigure}[b]{0.136\textwidth}
         \centering
         \includegraphics[width=\textwidth]{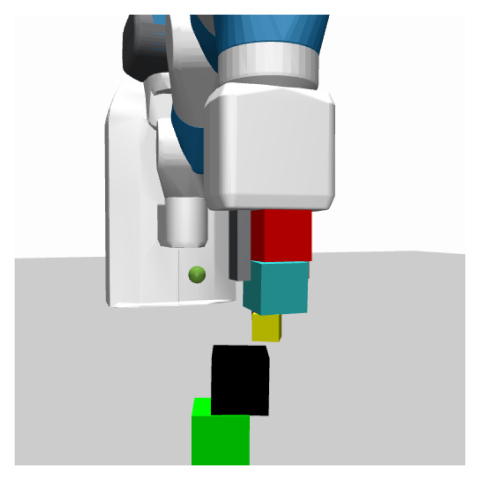}
        \centering{\scriptsize Iteration 259}
     \end{subfigure}
 \begin{subfigure}[b]{0.136\textwidth}
         \centering
         \includegraphics[width=\textwidth]{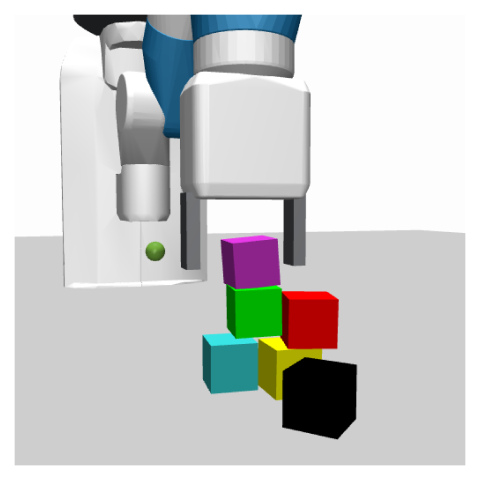}
        \centering{\scriptsize Iteration 275}
     \end{subfigure}
    \begin{subfigure}[b]{0.136\textwidth}
         \centering
         \includegraphics[width=\textwidth]{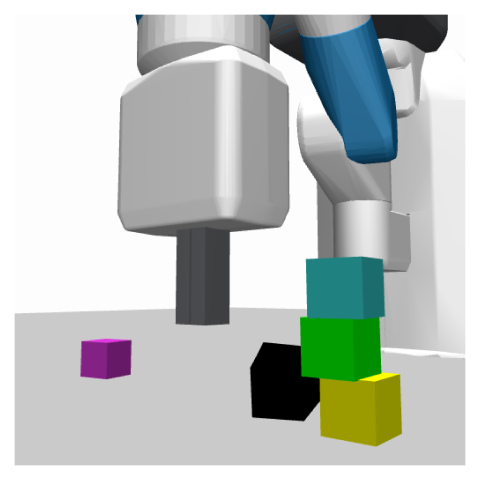}
        \centering{\scriptsize Iteration 284}
     \end{subfigure}
     \begin{subfigure}[b]{0.136\textwidth}
        \centering
         \includegraphics[width=\textwidth]{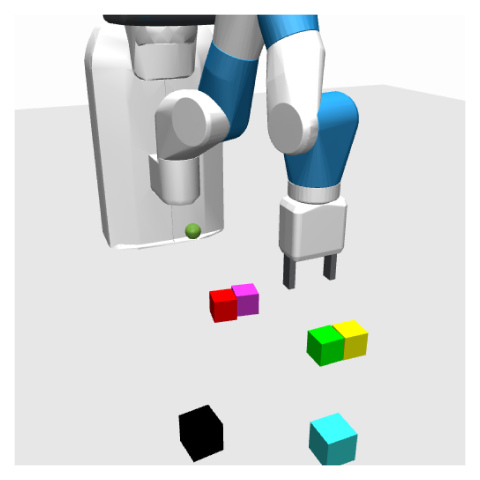}
        \centering{\scriptsize Iteration 286}
     \end{subfigure}
     \begin{subfigure}[b]{0.136\textwidth}
         \centering
         \includegraphics[width=\textwidth]{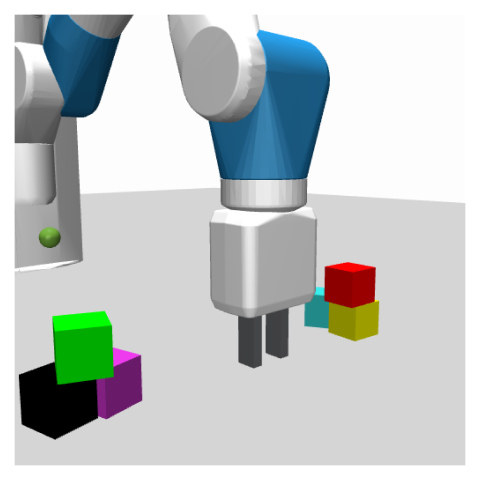}
        \centering{\scriptsize Iteration 296}
      \end{subfigure}
      \addtocounter{figure}{-1}
  \captionof{figure}{{\bf{Snapshots from free play with \ourMethodA{}.}} We showcase snapshots of highest \ourMethod values, equivalent to lowest entropy, from exemplary rollouts at different iterations of free play. Following the regularity objective, stacks and alignments are generated. %
  }
  \vspace{-10pt}
\label{fig:snapshots_freeplay}
\end{figure}

In \Fig{fig:interactions:FPP}, we analyze the quality of data generated during free play, in terms of observed interactions, for \ourMethodA with the augmentation weight $\lambda=0.1$, a pure \ourMethod run with no information-gain component in the intrinsic reward ($\lambda=0$),  \oldMethod, as well as RND and Disagreement.
For pure \ourMethod, we observe a decrease in the generated interactions. This has two reasons: 1) \ourMethod only aims to generate structure and the exploration problem is not solved, 2) once the controller finds a plan that leads to an optimum, even if it is local, there is no incentive to destroy it, unless a plan that results in better regularity can be found within the planning horizon. There is no discrimination between ``boring'' and ``interesting'' patterns with respect to the model's current capabilities. This in turn means that the robot creates \eg a (spaced) line, which is a local optimum for \ourMethod, and then spends the rest of the episode, not touching any objects to keep the created alignment intact. With the injection of some disagreement in \ourMethodA, we observe improved interaction metrics throughout free play in terms of 2 or more object interactions and objects being in the air (either being lifted by the robot or being stacked on top of another block). In practice, since the ensemble of models tends to hallucinate due to imperfect predictions, even for pure \ourMethod we observe dynamic pattern generations, as reflected in the interaction metrics (more details in \supp{app:pure_rair}).  For the plain disagreement case with \oldMethod, more flipping behavior, and less air time are observed during free play, since the agent favors chaos. We also observe that the baselines RND and Disagreement (planning horizon 1) produce less interaction-rich data during free play. Especially for Disagreement, this further shows that planning for future novelty is an essential component for free play.
\looseness-1

\begin{wrapfigure}[10]{r}{0.27\textwidth}
    \centering
    \vspace{-1.em}
    \includegraphics[width=\linewidth]{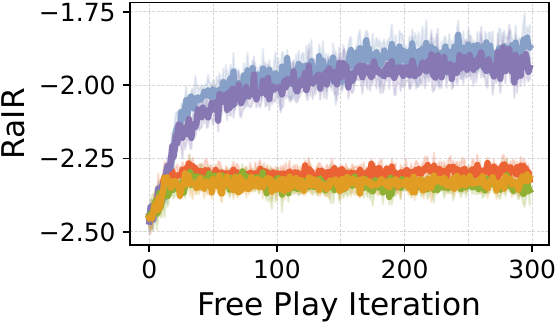}\vspace{-.2em}
    \caption{{\bf Highest \ourMethod value throughout free play} for pure {\color{ours_test_pure}\ourMethod}, {\color{ourviolet}\ourMethodA}, {\color{ourred}\oldMethod}, {\color{ourorange}RND} and  {\color{ourgreen}Dis}\looseness-1.}
    \label{fig:rair_best_values}
\end{wrapfigure}
Another reason why disagreement in \ourMethodA is helpful is due to the step-wise landscape of \ourMethod as shown in \fig{fig:rair_evolution}. Here, combining \ourMethod with ensemble disagreement effectively helps smoothen this reward function, making it easier to find plans with improvements in regularity with imperfect world models.

In \fig{fig:rair_best_values}, we report the highest achieved \ourMethod value in the collected rollouts throughout free play. We observe that pure \ourMethod and \ourMethodA indeed find more regular structures during play compared to the baselines. Some snapshots of regular structures observed during a \ourMethodA free play run are illustrated in \fig{fig:snapshots_freeplay}. %
Results for $\phi(s_i,s_j) =\nint{s_i - s_j}$ can be found in \supp{app:bidirectional}.\looseness-1

\subsection{Zero-shot Generalization to Assembly Downstream Tasks with \ourMethod in \FPP}
After the fully-intrinsic free-play phase, we evaluate zero-shot generalization performance on downstream tasks, where we perform model-based planning with the learned world models. Note that now instead of optimizing for intrinsic rewards, we are optimizing for extrinsic reward functions $r_{\text{task}}$ given by the environment (\supp{app:rewards}).
In \fig{fig:building_success}, we present the evolution of success rates of models checkpointed throughout free play on the following assembly tasks: singletower with 3 objects, 2 multitowers with 2 objects each, pyramid with 5 and 6 objects. %

The combination \ourMethodA yields significant improvements in the success rates of assembly tasks, as shown in \fig{fig:building_success} and \tab{tab:fpp:performance}. \ourMethod alone, outperforms \oldMethod in assembly tasks. As we are biasing exploration towards regularity, we see a decrease in more chaotic interactions during play time,
which is correlated with a decrease in performance for the more chaotic throwing and flipping tasks. For the generic Pick \& Place task, we observe comparable performance between \ourMethodA and \oldMethod. The decrease in performance for \ourMethod in non-assembly tasks shows the importance of an information-gain objective. The baselines RND and Disagreement exhibit very poor performance in the assembly tasks. In the other tasks, RND and \oldMethod are comparable (bold numbers show statistical indistinguishability from best with $p>0.05$) %
This showcases that guiding free play by the model's epistemic uncertainty as in the case with ensemble disagreement, helps learn robust and capable world models. The decrease in the zero-shot performance for Disagreement further proves the importance of planning for future novelty during free play. 
We also run free play in the custom \FPP environment for \ourMethodA and \oldMethod and observe improved zero-shot downstream performance in assembly tasks with \ourMethodA (see \supp{app:custom_fpp_perf}).\looseness-1

\begin{figure}[t]
     \vspace{.3em}
    \centering
  \captionof{table}{{\bf Zero-shot downstream task performance of \ourMethodA, \ourMethod, \oldMethod, RND and Dis} for assembly tasks as well as the generic pick \& place task and the more chaos-oriented throwing and flipping. Results are shown for five independent seeds. In the bottom row, we report the success rates achieved via planning with ground truth models. This is to provide a baseline for how hard the task is to solve with finite-horizon planning and potentially suboptimally designed task rewards.\looseness-1
}
  \vspace{-.3em}
    \label{tab:fpp:performance}
  \centering
    \renewcommand{\arraystretch}{1.03}
    \resizebox{1.\linewidth}{!}{ %
    \begin{tabular}{@{}lcccc|ccc@{}}
    \toprule
    & Singletower 3 & Multitower 2+2 & Pyramid  5& Pyramid  6& Pick\&Place 6& Throw 4& Flip 4\\
    \midrule
    \ourMethodA & $\mathbf{0.75 \pm 0.07}$  & $\mathbf{0.77 \pm 0.06}$ & $\mathbf{0.49 \pm 0.06}$ & $\mathbf{0.18 \pm 0.04}$ &
     $\mathbf{0.90 \pm 0.02}$ &$0.32 \pm 0.02$ & $0.63 \pm 0.08$ \\
    \ourMethod & $0.64 \pm 0.03$  & $0.62 \pm 0.03$ & $0.25 \pm 0.05$ &  $0.10 \pm 0.02$ &
     $0.74 \pm 0.05$ &$0.21 \pm 0.01$ & $0.65 \pm 0.1$ \\
     \oldMethod   & $0.40 \pm 0.12$ & $0.52 \pm 0.05$  & $0.14 \pm 0.09$  & $0.02 \pm 0.01$
     & $\mathbf{0.90 \pm 0.02}$ & $\mathbf{0.49 \pm 0.05}$ & $\mathbf{0.73 \pm 0.1}$ \\
     RND   & $0.07 \pm 0.07$ & $0.14 \pm 0.12$  & $0.02 \pm 0.02$  & $0.0 \pm 0.0$
     & $\mathbf{0.91 \pm 0.01}$ & $\mathbf{0.42 \pm 0.07}$ & $\mathbf{0.82 \pm 0.1}$ \\
     Dis   & $0.0 \pm 0.0$ & $0.01 \pm 0.01$  & $0.0 \pm 0.0$  & $0.0 \pm 0.0$
     & $\mathbf{0.89 \pm 0.01}$ & $0.30 \pm 0.04$ & $\mathbf{0.69 \pm 0.09}$ \\
    \midrule
    GT   & $0.99$ & $0.97$ & $0.82$  & $0.81$  & $0.99$ & $0.97$ & 1.0  \\
  \bottomrule
    \end{tabular}
  }

    \vspace{1em}
 	\centering
    \begin{subtable}[b]{0.01\textwidth}
    \notsotiny{\rotatebox{90}{\hspace{.5cm}\textsf{success rate}}}
    \end{subtable}
    \begin{subtable}[b]{0.24\textwidth}
         \centering
         {\scriptsize Singletower 3}\\[.1em]
         \quad\includegraphics[width=.6\textwidth]{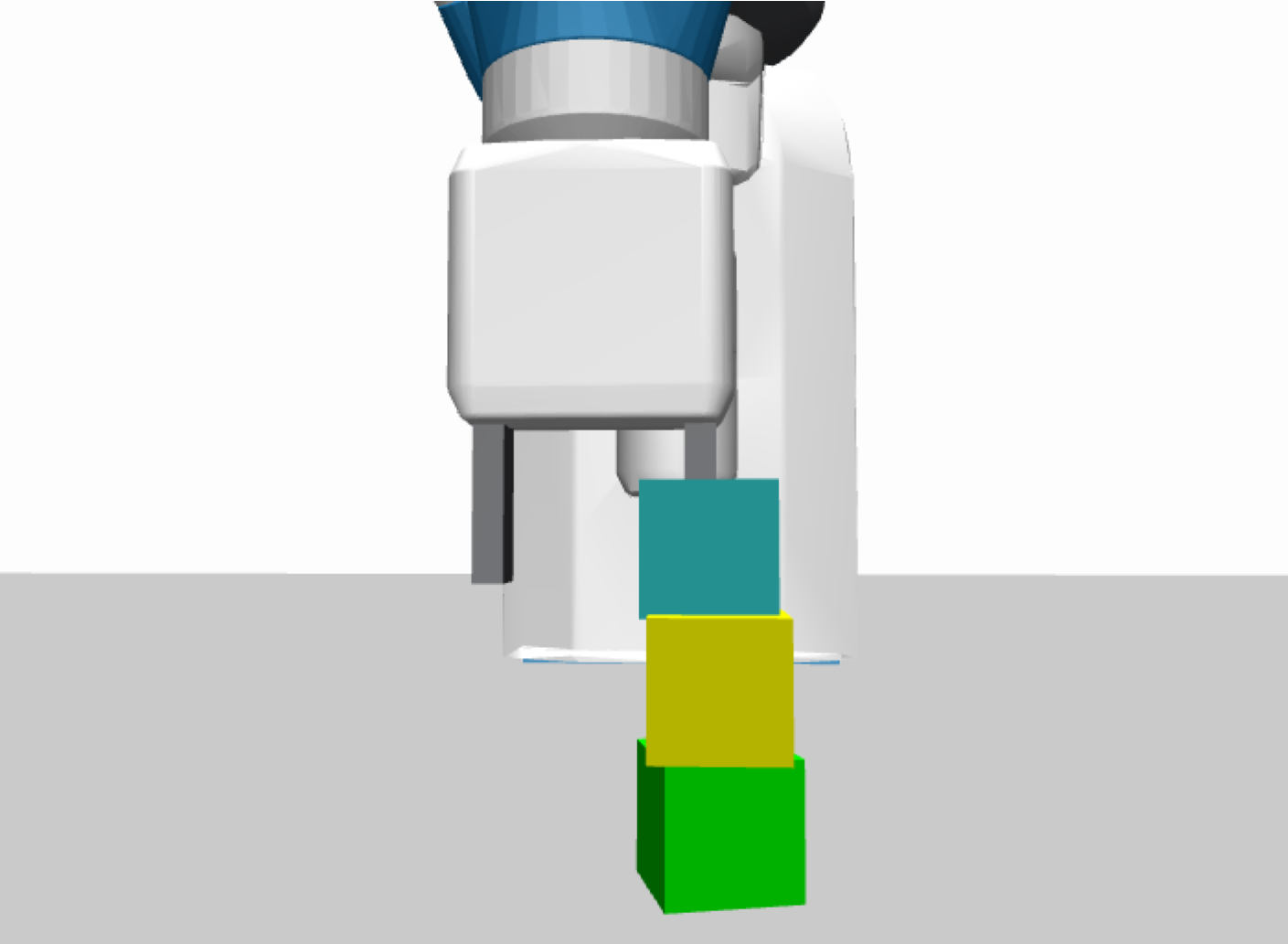}\\[.3em]
         \includegraphics[width=\linewidth]{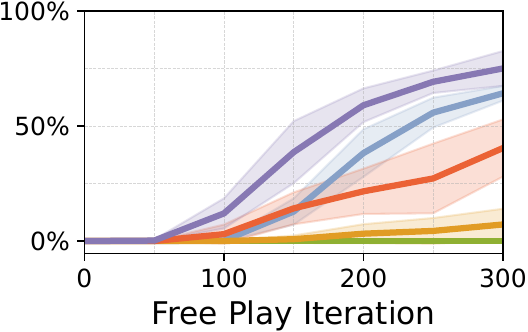}
     \end{subtable}
      \begin{subtable}[b]{0.24\textwidth}
         \centering
        {\scriptsize \phantom{g}Multitower 2+2\phantom{g}}\\[.1em]
        \quad\includegraphics[width=.6\textwidth]{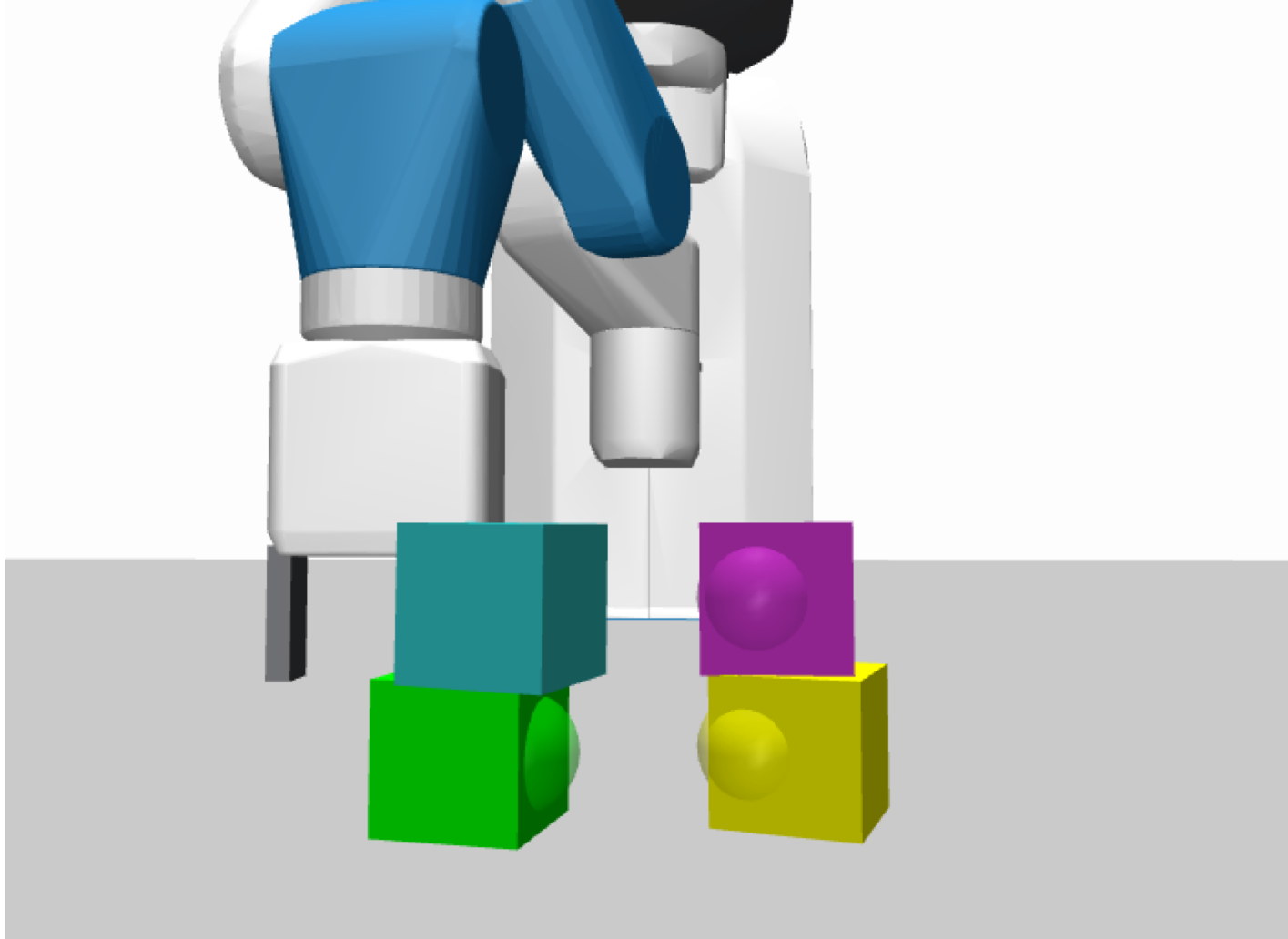}\\[.3em]
        \includegraphics[width=\linewidth]{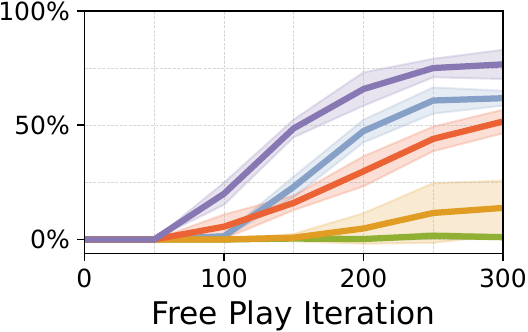}
     \end{subtable}
     \begin{subtable}[b]{0.24\textwidth}
         \centering
         {\scriptsize Pyramid 5}\\[.1em]
         \quad\includegraphics[width=.6\textwidth]{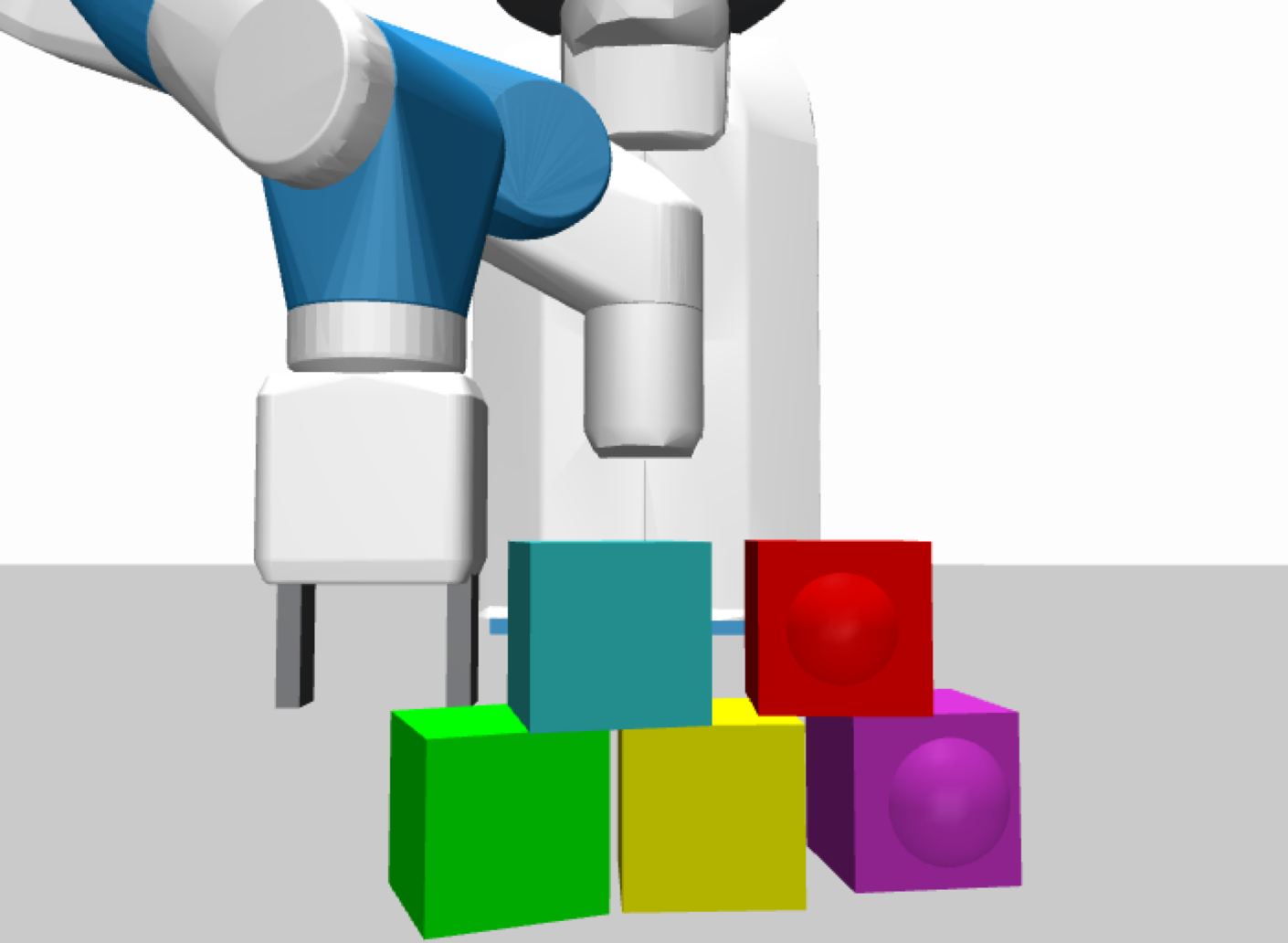}\\[.3em]
         \includegraphics[width=\linewidth]{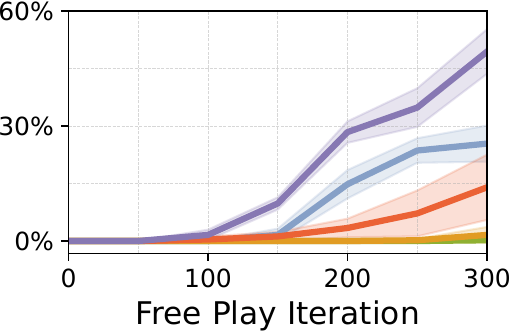}
     \end{subtable}
     \begin{subtable}[b]{0.24\textwidth}
       \centering
       {\scriptsize Pyramid 6}\\[.1em]
       \quad\includegraphics[width=.6\textwidth]{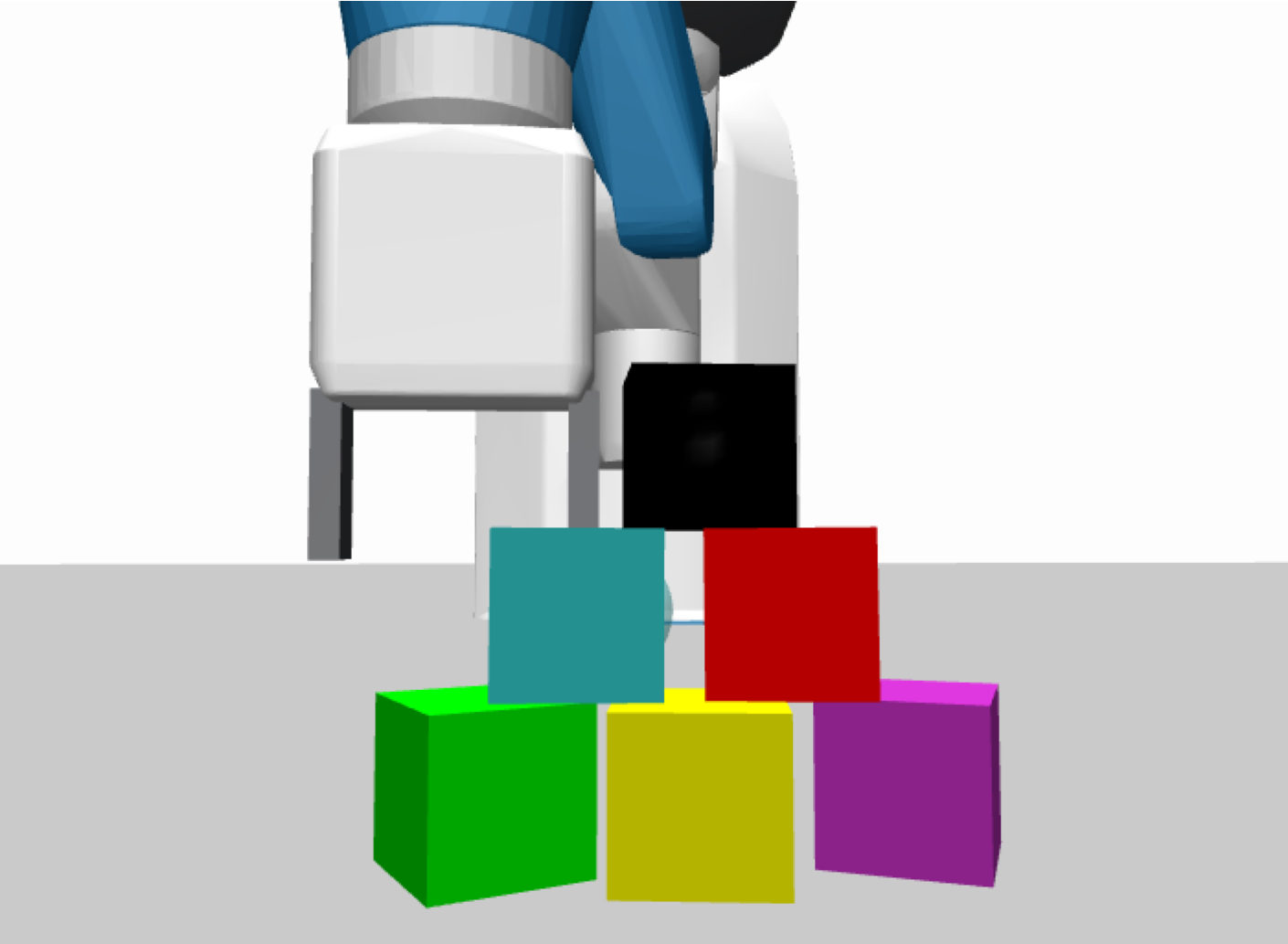}\\[.3em]
         \includegraphics[width=\linewidth]{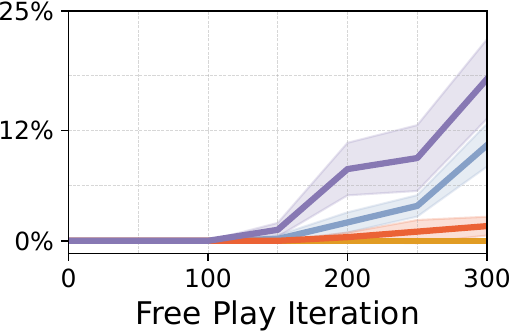}
       \end{subtable}
       \vspace{-.3em}
     \centering
    {\scriptsize
      \quad \textcolor{ours_test_pure}{\rule[2pt]{20pt}{1.2pt}} \ourMethod{}
      \quad \textcolor{ours}{\rule[2pt]{20pt}{1.2pt}} \ourMethodA{}
       \quad \textcolor{cee_us_test}{\rule[2pt]{20pt}{1.2pt}} \oldMethod
      \quad \textcolor{ourorange}{\rule[2pt]{20pt}{1.2pt}} RND
      \quad \textcolor{ourgreen}{\rule[2pt]{20pt}{1.2pt}} Dis
    }
\vspace{-.1em}
    \captionof{figure} {{\bf Success rates for zero-shot downstream task generalization for assembly tasks}
      in \FPP for model checkpoints over the course of free play.
      We compare \ourMethodA{} $(\lambda=0.1)$ and \ourMethod with \oldMethod, RND and Dis. %
      We used five independent seeds.} %
\label{fig:building_success}
\end{figure}

\subsection{Re-creating existing structures with \ourMethod}\label{sec:recreation}
\setlength{\intextsep}{1.0pt plus 2.0pt minus 2.0pt}
\setlength{\columnsep}{10pt}%
\begin{wrapfigure}{r}{0.38\textwidth}
\vspace{-.6em}
\addtocounter{figure}{-1} 
\begin{subfigure}[t]{.49\linewidth}
    \includegraphics[width=\linewidth, trim={0cm 0cm 0cm 3cm},clip]{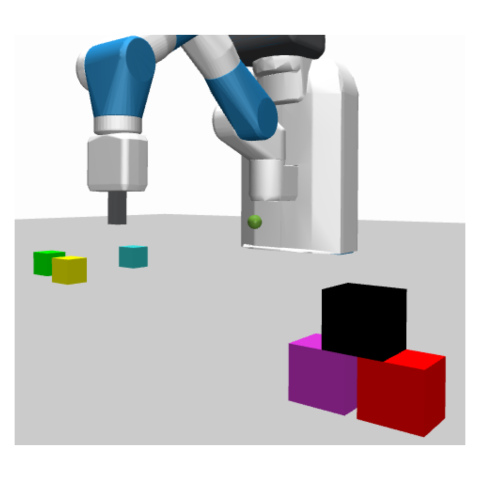}
    \vspace{-2em}
    \caption{$t=0$}
    \end{subfigure}
    \hfill
    \begin{subfigure}[t]{.49\linewidth}
    \includegraphics[width=\linewidth, trim={0cm 0cm 0cm 3cm},clip]{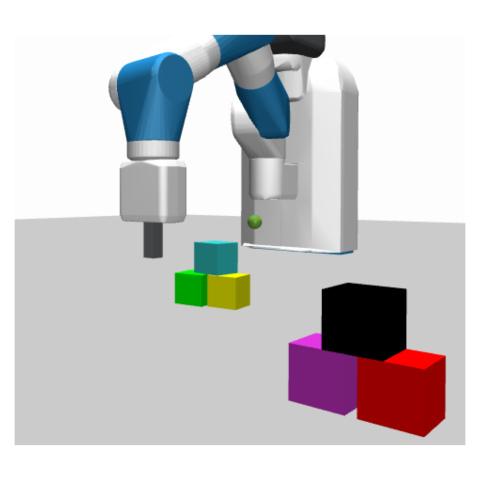}
    \vspace{-2em}
    \caption{$t=200$}
    \end{subfigure}\vspace{-.5em}
    \captionof{figure}{A pyramid initialized out of the robot's reach is re-created by optimizing for \ourMethod.\looseness-1}
    \label{fig:recreation}
\end{wrapfigure}
We test whether we can re-create existing arrangements in the environment with \ourMethod.
If there are regularities / sub-structures already present in the environment, then completing or re-creating these patterns naturally becomes an optimum for \ourMethod, as repeating this pattern introduces redundancy. %
We initialize pyramids, single- and multitowers out of the robot's manipulability range in \FPP. We then plan using iCEM to maximize \ourMethod with GT models. Doing so, the agent manages to re-create the existing structures in the environment with the blocks it has within reach. In \fig{fig:recreation}, this is showcased for a pyramid with 3 objects, where in 15 rollouts a pyramid is recreated in 73\% of the cases. Without the need to define any explicit reward functions, we can simply use our regularity objective to mimic existing ordered constellations. More details can be found in \supp{app:recreation}.\looseness-1

\section{\ourMethod in Locomotion Environments}

\begin{figure}[t]
  \centering
           \includegraphics[width=\textwidth]{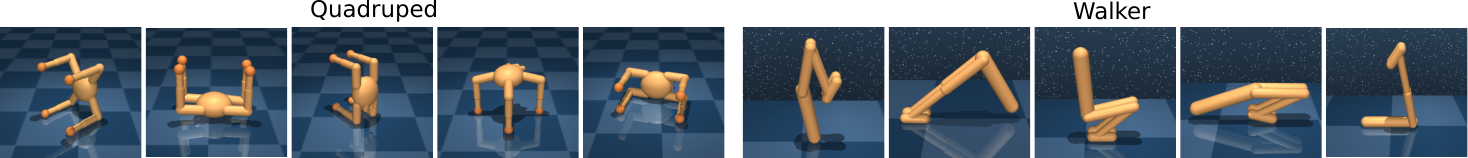}
    \vspace{-1.5em}
  \caption{{\bf{RaIR in Quadruped and Walker environments with GT models.}} We show generated poses when maximizing for regularity over the positions of different joints (\eg knees, toes).\looseness-1 
  }
  \vspace{-10pt}
       \label{fig:dmc_rair}
\end{figure}

The only requirement to incorporate \ourMethod in a given environment is an entity-based view of the world. We can easily apply this principle to locomotion environments, where we treat each joint as an entity. In \Fig{fig:dmc_rair}, we showcase the generated poses in the DeepMind Control Suite environments Quadruped and Walker, when we optimize for regularity using ground truth models. Here, regularity is computed over the $x$-$y$-$z$ positions of the knees and toes of the quadruped. For walker, we take the positions of the feet, legs and the torso. These poses also heavily overlap with the goal poses proposed in the RoboYoga benchmark \cite{Mendonca2021:LEXA} (see \fig{fig:quadruped_goals}), which further supports our hypothesis that regularities and our preference for them are ubiquitious.
We also apply free play with learned models in the Quadruped environment. More details and results can be found in \supp{app:roboyoga}.\looseness-1

\section{Related Work}\vspace{-.75em}
\textbf{Intrinsic motivation in RL} uses minimizing novelty/surprise to dissolve cognitive disequilibria as a prominent intrinsic reward signal definition \cite{pathak2017curiosity, schmidhuber1991possibility,Kim20:Active, groth2021curiosity, Storck1995Reinforcement-Driven-Information,BlaesVlastelicaZhuMartius2019:CWYC,PaoloEtAl2021:NoveltySearchSparseReward}.
As featured in this work, using the disagreement of an ensemble of world models as an estimate of expected information gain is a widely-used metric as it allows planning into the future \cite{pathak2019self, sekar2020planning, Sancaktaretal22}. Other prominent intrinsic rewards deployed in RL include learning progress \cite{schmidhuber1991possibility, BlaesVlastelicaZhuMartius2019:CWYC, Colas2019:CURIOUS}, empowerment \cite{KlyubinPolani2005:Empowerment,mohamed2015variational} and maximizing for state space coverage with count-based methods \cite{bellemare2016unifying, tang2017exploration} and RND \cite{burda2018exploration}. Another sub-category would be goal-conditioned unsupervised exploration methods combined with \eg ensemble disagreement \cite{Mendonca2021:LEXA,hu2023planning} or asymmetric self-play \cite{openai2021:assymmetric-selfplay}. In competence-based intrinsic motivation, unsupervised skill discovery methods aim to learn policies conditioned on a latent skill variable \cite{eysenbach2018diversity,sharma2019dynamics}. \citet{berseth2021smirl} propose surprise minimization as intrinsic reward to seek familiar states in unstable environments with an active source of entropy. Note that this differs from our work, as our notion of regularity is decoupled from surprise: \ourMethod aims to get to a state that is regular in itself.\looseness-1

\textbf{Compression}
and more specifically compression progress have been postulated as driving forces in human curiosity by \citet{schmidhuber2009:compression}. However, the focus has been on \emph{temporal} compression, where it is argued that short and simple explanations of the past make long-horizon planning easier. In our work, we don't focus on compression in the temporal dimension, \ie sequences of states. Instead, we perform compression as entropy minimization (in the relational case, equivalent to lossy compression) at a given timestep $t$, where we are interested in the relational redundancies in the current scene. More details on connections to compression can be found in \app{app:compression}.\looseness-1

\textbf{Assembly Tasks in RL} with 3+ objects pose an open challenge, where most methods achieve stacking via tailored learning curricula with more than 20 million environment steps \cite{li2020towards, lanier2019curiositydriven}, expert demonstrations \cite{nair2018overcoming}, also together with high-level actions \citep{biza2022factored}. \citet{hu2023planning} manage to solve 3-object stacking in an unsupervised setting with goal-conditioned RL (GCRL), using a very similar robotic setup to ours, but only with 30\% success rate. More discussion on GCRL in \supp{app:gcrl}.\looseness-1

\section{Discussion}\vspace{-.75em} \label{sec:discussion}
Although the search for regularity and symmetry has been studied extensively in developmental psychology, these concepts haven't been featured within reinforcement learning yet.
In this work, we propose a mathematical formulation of regularity as an intrinsic reward signal and operationalize it within model-based RL. We show that with our formulation of regularity, we indeed manage to create regular and symmetric patterns in a 2D grid environment as well as in a challenging compositional object manipulation environment. We also provide insights into the different components of \ourMethod and deepen the understanding of the types of regularities emerging from using different mappings $\phi$.
In the second part of the work, we incorporate \ourMethod within free play. Here, our goal is to bias information-search during exploration towards regularity.
We provide a proof-of-concept that augmenting epistemic uncertainty-based intrinsic rewards with \ourMethod helps exploration for symmetric and ordered arrangements. Finally, we also show that our regularity objective can simply be used to imitate existing regularities in the environment.\\ 
\textbf{Limitations and future work:} 
As we use finite-horizon planning, we don't necessarily converge to global optima. This can both be seen as a limitation and a feature, as it naturally allows us to obtain different levels of regularity in the generated patterns.
Currently, we are restricted to fully-observable MDPs. We embrace object-centric representations as a suitable inductive bias in RL, where the observations per object (consisting of poses and velocities) are naturally disentangled (more discussion in \app{app:object_centric}).
We also assume that this state space is interpretable such that we take, for instance, only the positions and color. 
The representational space, in which the \ourMethod measure is computed, is specified by the designer. Exciting future work would be to learn a representation under which the human relevant structures in the real-world (\eg towers, bridges) are indeed regular.\looseness-1 %

\begin{ack}
The authors thank Sebastian Blaes, Anselm Paulus, Marco Bagatella, Núria Armengol Urpí and Maximilian Seitzer for helpful discussions and for their help reviewing the manuscript. 
The authors thank the International Max Planck Research School for Intelligent Systems (IMPRS-IS) for supporting Cansu Sancaktar.
Georg Martius is a member of the Machine Learning Cluster of Excellence, EXC number 2064/1 -- Project number 390727645. 
We acknowledge the financial support from the German Federal Ministry of Education and Research (BMBF) through the Tübingen AI Center (FKZ: 01IS18039B).
This work was supported by the Volkswagen Stiftung (No 98 571).
\end{ack}

\bibliography{symmetry, curiosity}

\begin{thebibliography}{51}
\providecommand{\natexlab}[1]{#1}
\providecommand{\url}[1]{\texttt{#1}}
\expandafter\ifx\csname urlstyle\endcsname\relax
  \providecommand{\doi}[1]{doi: #1}\else
  \providecommand{\doi}{doi: \begingroup \urlstyle{rm}\Url}\fi

\bibitem[Liu et~al.(2010)Liu, Hel-Or, Kaplan, Van~Gool, et~al.]{liu2010computational}
Yanxi Liu, Hagit Hel-Or, Craig~S Kaplan, Luc Van~Gool, et~al.
\newblock Computational symmetry in computer vision and computer graphics.
\newblock \emph{Foundations and Trends{\textregistered} in Computer Graphics and Vision}, 5\penalty0 (1--2):\penalty0 1--195, 2010.

\bibitem[Attneave(1955)]{attneave1955symmetry}
Fred Attneave.
\newblock Symmetry, information, and memory for patterns.
\newblock \emph{The American journal of psychology}, 68\penalty0 (2):\penalty0 209--222, 1955.

\bibitem[Bornstein et~al.(1981)Bornstein, Ferdinandsen, and Gross]{bornstein1981perception}
Marc~H Bornstein, Kay Ferdinandsen, and Charles~G Gross.
\newblock Perception of symmetry in infancy.
\newblock \emph{Developmental psychology}, 17\penalty0 (1):\penalty0 82, 1981.

\bibitem[Bailey(1933)]{bailey1933scale}
Marjory~W Bailey.
\newblock A scale of block constructions for young children.
\newblock \emph{Child Development}, 4\penalty0 (2):\penalty0 121--139, 1933.

\bibitem[Zingrone(2014)]{zingrone2014construction}
William~A Zingrone.
\newblock The construction of symmetry in children and adults.
\newblock \emph{The Journal of genetic psychology}, 175\penalty0 (2):\penalty0 91--104, 2014.

\bibitem[Golomb(1987)]{golomb1987development}
Claire Golomb.
\newblock The development of compositional strategies in children's drawings.
\newblock \emph{Visual Arts Research}, pages 42--52, 1987.

\bibitem[Gibson(1988)]{gibson1988exploratory}
Eleanor~J Gibson.
\newblock Exploratory behavior in the development of perceiving, acting, and the acquiring of knowledge.
\newblock \emph{Annual review of psychology}, 39\penalty0 (1):\penalty0 1--42, 1988.

\bibitem[Langer(1986)]{langer1986origins}
Jonas Langer.
\newblock \emph{The origins of logic: One to two years}.
\newblock Academic Press, 1986.
\newblock ISBN 978-0124365551.

\bibitem[Pathak et~al.(2017)Pathak, Agrawal, Efros, and Darrell]{pathak2017curiosity}
Deepak Pathak, Pulkit Agrawal, Alexei~A. Efros, and Trevor Darrell.
\newblock Curiosity-driven exploration by self-supervised prediction.
\newblock In \emph{International Conference on Machine Learning (ICML)}, 2017.
\newblock URL \url{https://proceedings.mlr.press/v70/pathak17a.html}.

\bibitem[Pathak et~al.(2019)Pathak, Gandhi, and Gupta]{pathak2019self}
Deepak Pathak, Dhiraj Gandhi, and Abhinav Gupta.
\newblock Self-supervised exploration via disagreement.
\newblock In \emph{International Conference on Machine Learning (ICML)}, 2019.
\newblock URL \url{https://proceedings.mlr.press/v97/pathak19a.html}.

\bibitem[Sekar et~al.(2020)Sekar, Rybkin, Daniilidis, Abbeel, Hafner, and Pathak]{sekar2020planning}
Ramanan Sekar, Oleh Rybkin, Kostas Daniilidis, Pieter Abbeel, Danijar Hafner, and Deepak Pathak.
\newblock Planning to explore via self-supervised world models.
\newblock In \emph{International Conference on Machine Learning (ICML)}, 2020.
\newblock URL \url{https://proceedings.mlr.press/v119/sekar20a.html}.

\bibitem[Sancaktar et~al.(2022)Sancaktar, Blaes, and Martius]{Sancaktaretal22}
Cansu Sancaktar, Sebastian Blaes, and Georg Martius.
\newblock Curious exploration via structured world models yields zero-shot object manipulation.
\newblock In \emph{Advances in Neural Information Processing Systems (NeurIPS)}, 2022.

\bibitem[Lin(1996)]{lin1996correlation}
Shu-Kun Lin.
\newblock Correlation of entropy with similarity and symmetry.
\newblock \emph{Journal of Chemical Information and Computer Sciences}, 36\penalty0 (3):\penalty0 367--376, 1996.

\bibitem[Lazarev(2022)]{lazarev2022information}
Daniel Lazarev.
\newblock Information measures for entropy and symmetry.
\newblock \emph{arXiv preprint arXiv:2211.14857}, 2022.

\bibitem[Bormashenko(2020)]{bormashenko2020entropy}
Edward Bormashenko.
\newblock Entropy, information, and symmetry; ordered is symmetrical, ii: system of spins in the magnetic field.
\newblock \emph{Entropy}, 22\penalty0 (2):\penalty0 235, 2020.

\bibitem[Schmidhuber(2009)]{schmidhuber2009:compression}
J{\"u}rgen Schmidhuber.
\newblock Driven by compression progress: A simple principle explains essential aspects of subjective beauty, novelty, surprise, interestingness, attention, curiosity, creativity, art, science, music, jokes.
\newblock In \emph{Anticipatory Behavior in Adaptive Learning Systems: From Psychological Theories to Artificial Cognitive Systems 4}, pages 48--76. Springer, 2009.
\newblock Longer preprint: \url{https://arxiv.org/abs/0812.4360}.

\bibitem[Schmidhuber(2013)]{schmidhuber2013powerplay}
J{\"u}rgen Schmidhuber.
\newblock Powerplay: Training an increasingly general problem solver by continually searching for the simplest still unsolvable problem.
\newblock \emph{Frontiers in Psychology}, 4:\penalty0 313, 2013.

\bibitem[Chua et~al.(2018)Chua, Calandra, McAllister, and Levine]{chua2018}
Kurtland Chua, Roberto Calandra, Rowan McAllister, and Sergey Levine.
\newblock Deep reinforcement learning in a handful of trials using probabilistic dynamics models.
\newblock In \emph{Advances in Neural Information Processing Systems (NeurIPS)}, 2018.
\newblock URL \url{https://proceedings.neurips.cc/paper_files/paper/2018/file/3de568f8597b94bda53149c7d7f5958c-Paper.pdf}.

\bibitem[Pinneri et~al.(2020)Pinneri, Sawant, Blaes, Achterhold, Stueckler, Rolinek, and Martius]{pinneri2020:iCEM}
Cristina Pinneri, Shambhuraj Sawant, Sebastian Blaes, Jan Achterhold, Joerg Stueckler, Michal Rolinek, and Georg Martius.
\newblock Sample-efficient cross-entropy method for real-time planning.
\newblock In \emph{Conference on Robot Learning (CoRL)}, 2020.
\newblock URL \url{https://proceedings.mlr.press/v155/pinneri21a.html}.

\bibitem[Shannon(1948)]{shannon1948mathematical}
Claude~E Shannon.
\newblock A mathematical theory of communication.
\newblock \emph{The Bell system technical journal}, 27\penalty0 (3):\penalty0 379--423, 1948.

\bibitem[Burda et~al.(2019)Burda, Edwards, Storkey, and Klimov]{burda2018exploration}
Yuri Burda, Harrison Edwards, Amos Storkey, and Oleg Klimov.
\newblock Exploration by random network distillation.
\newblock In \emph{International Conference on Learning Representations (ICLR)}, 2019.
\newblock URL \url{https://openreview.net/forum?id=H1lJJnR5Ym}.

\bibitem[Battaglia et~al.(2018)Battaglia, Hamrick, Bapst, Sanchez-Gonzalez, Zambaldi, Malinowski, Tacchetti, Raposo, Santoro, Faulkner, Gulcehre, Song, Ballard, Gilmer, Dahl, Vaswani, Allen, Nash, Langston, Dyer, Heess, Wierstra, Kohli, Botvinick, Vinyals, Li, and Pascanu]{battaglia2018relational}
Peter~W. Battaglia, Jessica~B. Hamrick, Victor Bapst, Alvaro Sanchez-Gonzalez, Vinicius Zambaldi, Mateusz Malinowski, Andrea Tacchetti, David Raposo, Adam Santoro, Ryan Faulkner, Caglar Gulcehre, Francis Song, Andrew Ballard, Justin Gilmer, George Dahl, Ashish Vaswani, Kelsey Allen, Charles Nash, Victoria Langston, Chris Dyer, Nicolas Heess, Daan Wierstra, Pushmeet Kohli, Matt Botvinick, Oriol Vinyals, Yujia Li, and Razvan Pascanu.
\newblock Relational inductive biases, deep learning, and graph networks.
\newblock \emph{arXiv:1806.01261}, 2018.
\newblock URL \url{https://arxiv.org/abs/1806.01261}.

\bibitem[Plappert et~al.(2018)Plappert, Andrychowicz, Ray, McGrew, Baker, Powell, Schneider, Tobin, Chociej, Welinder, Kumar, and Zaremba]{Plappert2018MultiGoalRL}
Matthias Plappert, Marcin Andrychowicz, Alex Ray, Bob McGrew, Bowen Baker, Glenn Powell, J.~Schneider, Joshua Tobin, Maciek Chociej, P.~Welinder, V.~Kumar, and W.~Zaremba.
\newblock Multi-goal reinforcement learning: Challenging robotics environments and request for research.
\newblock \emph{arXiv preprint arXiv:1802.09464}, 2018.
\newblock URL \url{https://arxiv.org/abs/1802.09464}.

\bibitem[Li et~al.(2020)Li, Jabri, Darrell, and Agrawal]{li2020towards}
Richard Li, Allan Jabri, Trevor Darrell, and Pulkit Agrawal.
\newblock Towards practical multi-object manipulation using relational reinforcement learning.
\newblock In \emph{2020 IEEE International Conference on Robotics and Automation (ICRA)}, pages 4051--4058. IEEE, 2020.
\newblock URL \url{https://arxiv.org/abs/1912.11032}.

\bibitem[Mendonca et~al.(2021)Mendonca, Rybkin, Daniilidis, Hafner, and Pathak]{Mendonca2021:LEXA}
Russell Mendonca, Oleh Rybkin, Kostas Daniilidis, Danijar Hafner, and Deepak Pathak.
\newblock Discovering and achieving goals via world models.
\newblock In \emph{Advances in Neural Information Processing Systems (NeurIPS)}, 2021.
\newblock URL \url{https://openreview.net/forum?id=6vWuYzkp8d}.

\bibitem[Schmidhuber(1991)]{schmidhuber1991possibility}
J{\"u}rgen Schmidhuber.
\newblock A possibility for implementing curiosity and boredom in model-building neural controllers.
\newblock In \emph{Proceedings of the International Conference on Simulation of Adaptive Behavior: From Animals to Animats}, 1991.
\newblock URL \url{https://dl.acm.org/doi/10.5555/116517.116542}.

\bibitem[Kim et~al.(2020)Kim, Sano, De~Freitas, Haber, and Yamins]{Kim20:Active}
Kuno Kim, Megumi Sano, Julian De~Freitas, Nick Haber, and Daniel Yamins.
\newblock Active world model learning with progress curiosity.
\newblock In \emph{International Conference on Machine Learning (ICML)}, 2020.
\newblock URL \url{https://arxiv.org/abs/2007.07853}.

\bibitem[Groth et~al.(2021)Groth, Wulfmeier, Vezzani, Dasagi, Hertweck, Hafner, Heess, and Riedmiller]{groth2021curiosity}
Oliver Groth, Markus Wulfmeier, Giulia Vezzani, Vibhavari Dasagi, Tim Hertweck, Roland Hafner, Nicolas Heess, and Martin Riedmiller.
\newblock Is curiosity all you need? on the utility of emergent behaviours from curious exploration.
\newblock \emph{arXiv preprint arXiv:2109.08603}, 2021.
\newblock URL \url{https://arxiv.org/abs/2109.08603}.

\bibitem[Storck et~al.(1995)Storck, Hochreiter, and Schmidhuber]{Storck1995Reinforcement-Driven-Information}
J.~Storck, S.~Hochreiter, and J.~Schmidhuber.
\newblock Reinforcement driven information acquisition in non-deterministic environments.
\newblock In \emph{Proceedings of the International Conference on Artificial Neural Networks}, pages 159--164, Paris, 1995. EC2 \& Cie.
\newblock URL \url{https://people.idsia.ch/~juergen/icann95new.pdf}.

\bibitem[Blaes et~al.(2019)Blaes, Vlastelica, Zhu, and Martius]{BlaesVlastelicaZhuMartius2019:CWYC}
Sebastian Blaes, Marin Vlastelica, Jia-Jie Zhu, and Georg Martius.
\newblock Control {W}hat {Y}ou {C}an: {I}ntrinsically motivated task-planning agent.
\newblock In \emph{Advances in Neural Information Processing Systems (NeurIPS)}, 2019.
\newblock URL \url{https://proceedings.neurips.cc/paper/2019/hash/b6f97e6f0fd175613910d613d574d0cb-Abstract.html}.

\bibitem[Paolo et~al.(2021)Paolo, Coninx, Doncieux, and Laflaqui\`{e}re]{PaoloEtAl2021:NoveltySearchSparseReward}
Giuseppe Paolo, Alexandre Coninx, Stephane Doncieux, and Alban Laflaqui\`{e}re.
\newblock Sparse reward exploration via novelty search and emitters.
\newblock In \emph{Proceedings of the Genetic and Evolutionary Computation Conference}, page 154–162, 2021.
\newblock URL \url{https://doi.org/10.1145/3449639.3459314}.

\bibitem[Colas et~al.(2019)Colas, Fournier, Chetouani, Sigaud, and Oudeyer]{Colas2019:CURIOUS}
C{\'e}dric Colas, Pierre Fournier, Mohamed Chetouani, Olivier Sigaud, and Pierre-Yves Oudeyer.
\newblock {CURIOUS}: Intrinsically motivated modular multi-goal reinforcement learning.
\newblock In \emph{International Conference on Machine Learning (ICML)}, 2019.
\newblock URL \url{https://proceedings.mlr.press/v97/colas19a.html}.

\bibitem[Klyubin et~al.(2005)Klyubin, Polani, and Nehaniv]{KlyubinPolani2005:Empowerment}
A.S. Klyubin, D.~Polani, and C.L. Nehaniv.
\newblock Empowerment: a universal agent-centric measure of control.
\newblock In \emph{IEEE Congress on Evolutionary Computation}, volume~1, pages 128--135 Vol.1, 2005.
\newblock URL \url{https://ieeexplore.ieee.org/document/1554676}.

\bibitem[Mohamed and Jimenez~Rezende(2015)]{mohamed2015variational}
Shakir Mohamed and Danilo Jimenez~Rezende.
\newblock Variational information maximisation for intrinsically motivated reinforcement learning.
\newblock In \emph{Advances in Neural Information Processing Systems (NeurIPS)}, 2015.
\newblock URL \url{https://arxiv.org/abs/1509.08731}.

\bibitem[Bellemare et~al.(2016)Bellemare, Srinivasan, Ostrovski, Schaul, Saxton, and Munos]{bellemare2016unifying}
Marc Bellemare, Sriram Srinivasan, Georg Ostrovski, Tom Schaul, David Saxton, and Remi Munos.
\newblock Unifying count-based exploration and intrinsic motivation.
\newblock \emph{Advances in Neural Information Processing Systems (NeurIPS)}, 2016.

\bibitem[Tang et~al.(2017)Tang, Houthooft, Foote, Stooke, Xi~Chen, Duan, Schulman, DeTurck, and Abbeel]{tang2017exploration}
Haoran Tang, Rein Houthooft, Davis Foote, Adam Stooke, OpenAI Xi~Chen, Yan Duan, John Schulman, Filip DeTurck, and Pieter Abbeel.
\newblock \# exploration: A study of count-based exploration for deep reinforcement learning.
\newblock \emph{Advances in Neural Information Processing Systems (NeurIPS)}, 2017.

\bibitem[Hu et~al.(2023)Hu, Chang, Rybkin, and Jayaraman]{hu2023planning}
Edward~S. Hu, Richard Chang, Oleh Rybkin, and Dinesh Jayaraman.
\newblock Planning goals for exploration.
\newblock In \emph{International Conference on Learning Representations (ICLR)}, 2023.
\newblock URL \url{https://openreview.net/forum?id=6qeBuZSo7Pr}.

\bibitem[OpenAI et~al.(2021)OpenAI, Plappert, Sampedro, Xu, Akkaya, Kosaraju, Welinder, D'Sa, Petron, de~Oliveira~Pinto, Paino, Noh, Weng, Yuan, Chu, and Zaremba]{openai2021:assymmetric-selfplay}
OpenAI, Matthias Plappert, Raul Sampedro, Tao Xu, Ilge Akkaya, Vineet Kosaraju, Peter Welinder, Ruben D'Sa, Arthur Petron, Henrique~Ponde de~Oliveira~Pinto, Alex Paino, Hyeonwoo Noh, Lilian Weng, Qiming Yuan, Casey Chu, and Wojciech Zaremba.
\newblock Asymmetric self-play for automatic goal discovery in robotic manipulation.
\newblock \emph{arXiv:2101.04882}, 2021.
\newblock URL \url{https://arxiv.org/abs/2101.04882}.

\bibitem[Eysenbach et~al.(2019)Eysenbach, Gupta, Ibarz, and Levine]{eysenbach2018diversity}
Benjamin Eysenbach, Abhishek Gupta, Julian Ibarz, and Sergey Levine.
\newblock Diversity is all you need: Learning skills without a reward function.
\newblock In \emph{International Conference on Learning Representations (ICLR)}, 2019.
\newblock URL \url{https://openreview.net/forum?id=SJx63jRqFm}.

\bibitem[Sharma et~al.(2020)Sharma, Gu, Levine, Kumar, and Hausman]{sharma2019dynamics}
Archit Sharma, Shixiang Gu, Sergey Levine, Vikash Kumar, and Karol Hausman.
\newblock Dynamics-aware unsupervised discovery of skills.
\newblock In \emph{International Conference on Learning Representations (ICLR)}, 2020.
\newblock URL \url{https://openreview.net/forum?id=HJgLZR4KvH}.

\bibitem[Berseth et~al.(2021)Berseth, Geng, Devin, Rhinehart, Finn, Jayaraman, and Levine]{berseth2021smirl}
Glen Berseth, Daniel Geng, Coline~Manon Devin, Nicholas Rhinehart, Chelsea Finn, Dinesh Jayaraman, and Sergey Levine.
\newblock {\{}SM{\}}irl: Surprise minimizing reinforcement learning in unstable environments.
\newblock In \emph{International Conference on Learning Representations (ICLR)}, 2021.
\newblock URL \url{https://openreview.net/forum?id=cPZOyoDloxl}.

\bibitem[Lanier et~al.(2019)Lanier, McAleer, and Baldi]{lanier2019curiositydriven}
John~B. Lanier, Stephen McAleer, and Pierre Baldi.
\newblock Curiosity-driven multi-criteria hindsight experience replay, 2019.

\bibitem[Nair et~al.(2018)Nair, McGrew, Andrychowicz, Zaremba, and Abbeel]{nair2018overcoming}
Ashvin Nair, Bob McGrew, Marcin Andrychowicz, Wojciech Zaremba, and Pieter Abbeel.
\newblock Overcoming exploration in reinforcement learning with demonstrations.
\newblock In \emph{2018 IEEE international conference on robotics and automation (ICRA)}, pages 6292--6299. IEEE, 2018.

\bibitem[Biza et~al.(2022)Biza, Kipf, Klee, Platt, van~de Meent, and Wong]{biza2022factored}
Ondrej Biza, Thomas Kipf, David Klee, Robert Platt, Jan-Willem van~de Meent, and Lawson~LS Wong.
\newblock Factored world models for zero-shot generalization in robotic manipulation.
\newblock \emph{arXiv preprint arXiv:2202.05333}, 2022.
\newblock URL \url{https://arxiv.org/abs/2202.05333}.

\bibitem[Tassa et~al.(2018)Tassa, Doron, Muldal, Erez, Li, Casas, Budden, Abdolmaleki, Merel, Lefrancq, et~al.]{tassa2018deepmind}
Yuval Tassa, Yotam Doron, Alistair Muldal, Tom Erez, Yazhe Li, Diego de~Las Casas, David Budden, Abbas Abdolmaleki, Josh Merel, Andrew Lefrancq, et~al.
\newblock Deepmind control suite.
\newblock \emph{arXiv preprint arXiv:1801.00690}, 2018.

\bibitem[Locatello et~al.(2020)Locatello, Weissenborn, Unterthiner, Mahendran, Heigold, Uszkoreit, Dosovitskiy, and Kipf]{locatello2020object}
Francesco Locatello, Dirk Weissenborn, Thomas Unterthiner, Aravindh Mahendran, Georg Heigold, Jakob Uszkoreit, Alexey Dosovitskiy, and Thomas Kipf.
\newblock Object-centric learning with slot attention.
\newblock \emph{Advances in Neural Information Processing Systems (NeurIPS)}, 2020.

\bibitem[Tsividis et~al.(2021)Tsividis, Loula, Burga, Foss, Campero, Pouncy, Gershman, and Tenenbaum]{tsividis2021human}
Pedro~A. Tsividis, Joao Loula, Jake Burga, Nathan Foss, Andres Campero, Thomas Pouncy, Samuel~J. Gershman, and Joshua~B. Tenenbaum.
\newblock Human-level reinforcement learning through theory-based modeling, exploration, and planning.
\newblock \emph{arXiv:2107.12544}, 2021.
\newblock URL \url{https://arxiv.org/abs/2107.12544}.

\bibitem[Battaglia et~al.(2016)Battaglia, Pascanu, Lai, Jimenez~Rezende, et~al.]{battaglia2016interaction}
Peter Battaglia, Razvan Pascanu, Matthew Lai, Danilo Jimenez~Rezende, et~al.
\newblock Interaction networks for learning about objects, relations and physics.
\newblock \emph{Advances in Neural Information Processing Systems (NeurIPS)}, 2016.

\bibitem[Spelke(1990)]{spelke1990principles}
Elizabeth~S Spelke.
\newblock Principles of object perception.
\newblock \emph{Cognitive science}, 14\penalty0 (1):\penalty0 29--56, 1990.

\bibitem[Spelke et~al.(1993)Spelke, Breinlinger, Jacobson, and Phillips]{spelke1993gestalt}
Elizabeth~S Spelke, Karen Breinlinger, Kristen Jacobson, and Ann Phillips.
\newblock Gestalt relations and object perception: A developmental study.
\newblock \emph{Perception}, 22\penalty0 (12):\penalty0 1483--1501, 1993.

\bibitem[Peters and Kriegeskorte(2021)]{peters2021capturing}
Benjamin Peters and Nikolaus Kriegeskorte.
\newblock Capturing the objects of vision with neural networks.
\newblock \emph{Nature human behaviour}, 5\penalty0 (9):\penalty0 1127--1144, 2021.

\end{thebibliography}
\newpage

\appendix
\renewcommand{\figurename}{\bf Figure}
\renewcommand{\tablename}{\bf Table}

\renewcommand{\thefigure}{S\arabic{figure}}
\renewcommand{\thetable}{S\arabic{table}}
\renewcommand{\thealgorithm}{S\arabic{algorithm}}
\renewcommand{\theequation}{S\arabic{equation}}

\setcounter{figure}{0}
\setcounter{table}{0}
\setcounter{algorithm}{0}
\setcounter{equation}{0}

\hrule height 4pt
\vskip -\parskip%
{\center \textbf{\Large Supplementary Material for \\
Regularity as Intrinsic Reward for Free Play \\}}
\vskip 0.29in
\vskip -\parskip
\hrule height 1pt
\vskip 0.09in%

Code and videos are available at \url{https://sites.google.com/view/rair-project}.

\section{Experiment Results with Ground Truth Models} \label{app:3dim_compression}
\subsection{Experiment Results for \ourMethod in \FPP with $x$-$y$-$z$}

\begin{figure}[!hb]
  \centering
    \begin{subfigure}[b]{0.15\textwidth}
         \centering
         \includegraphics[width=\textwidth]{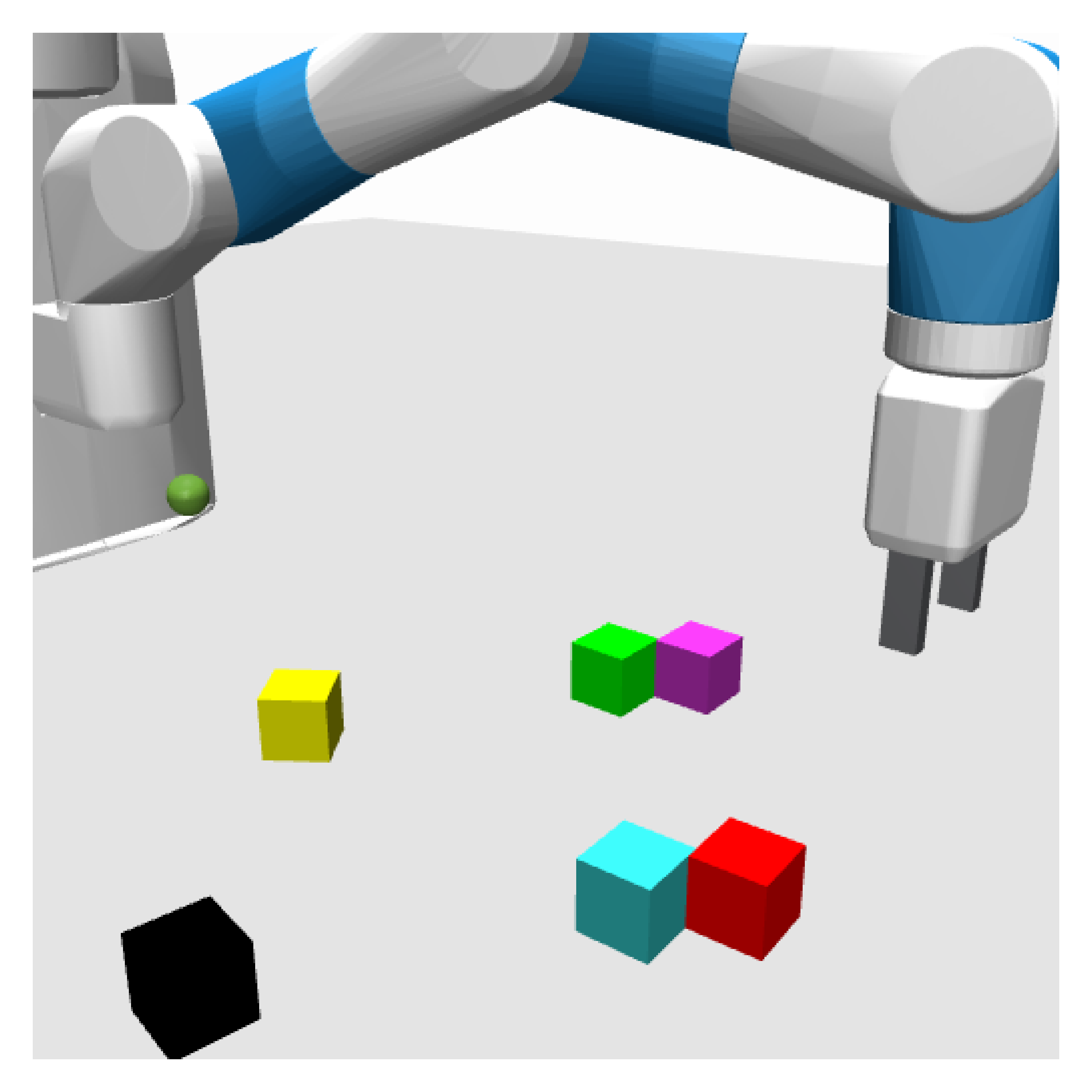}
     \end{subfigure}
      \begin{subfigure}[b]{0.15\textwidth}
         \centering
         \includegraphics[width=\textwidth]{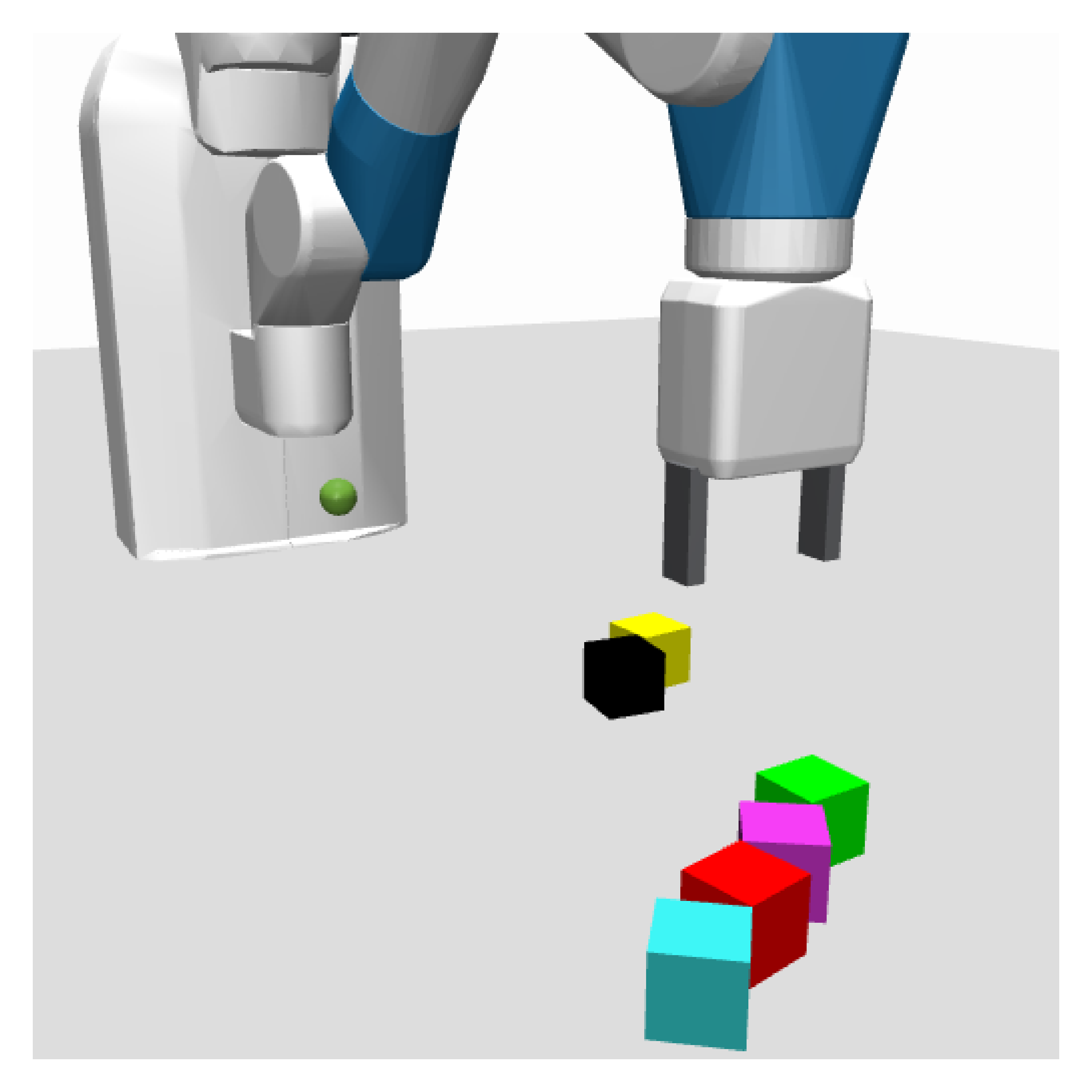}
     \end{subfigure}
     \begin{subfigure}[b]{0.15\textwidth}
         \centering
         \includegraphics[width=\textwidth]{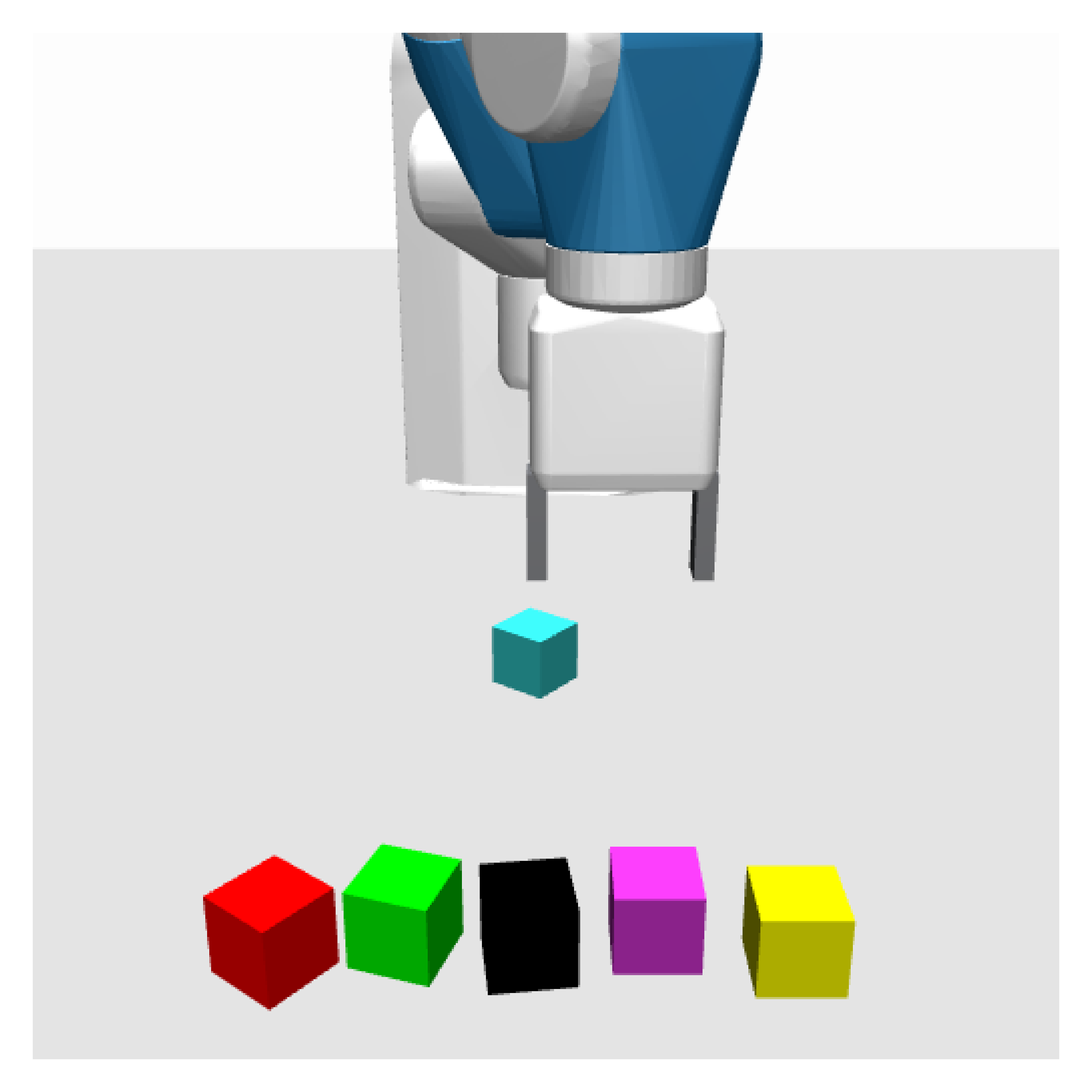}
     \end{subfigure}
 \begin{subfigure}[b]{0.15\textwidth}
         \centering
         \includegraphics[width=\textwidth]{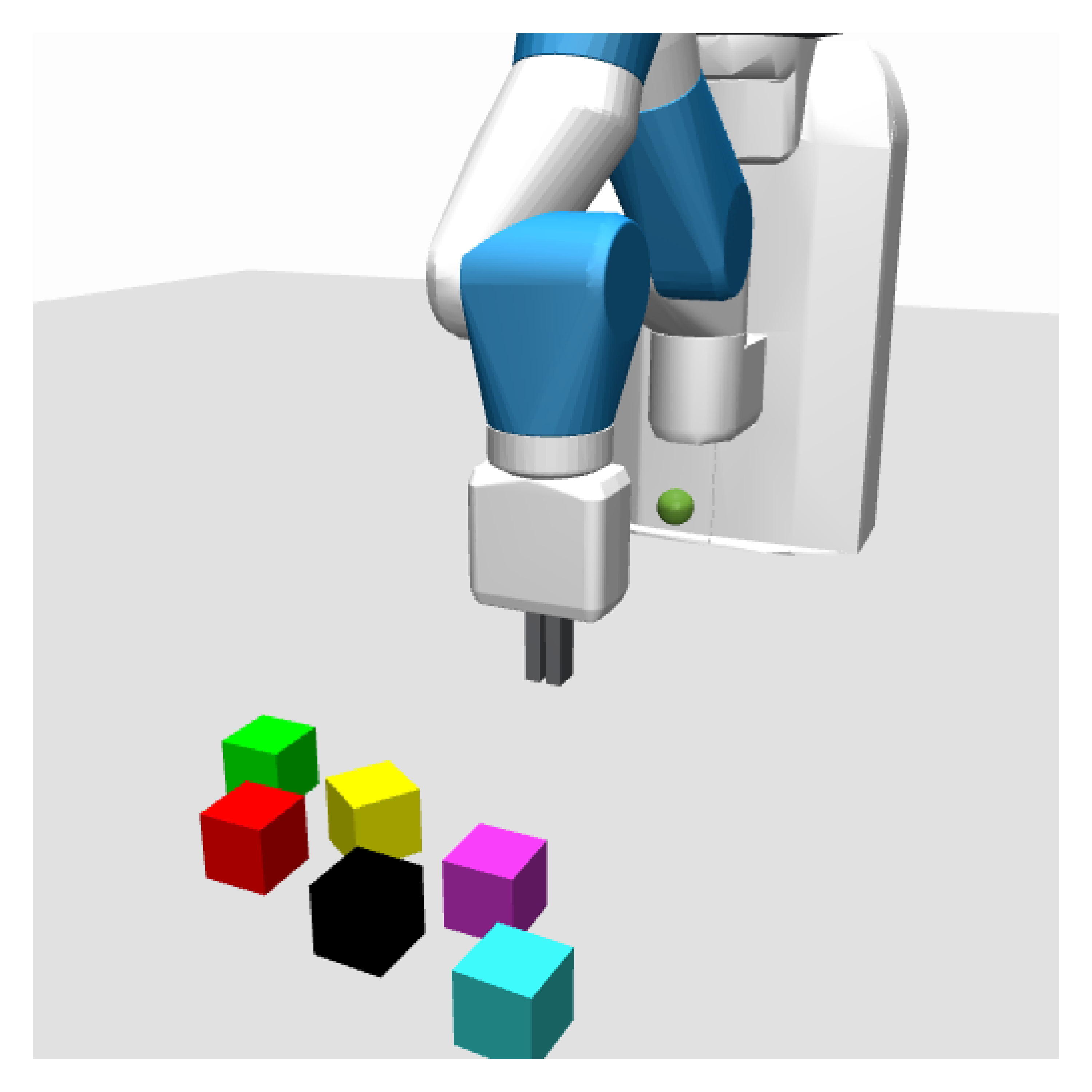}
     \end{subfigure}
    \begin{subfigure}[b]{0.15\textwidth}
         \centering
         \includegraphics[width=\textwidth]{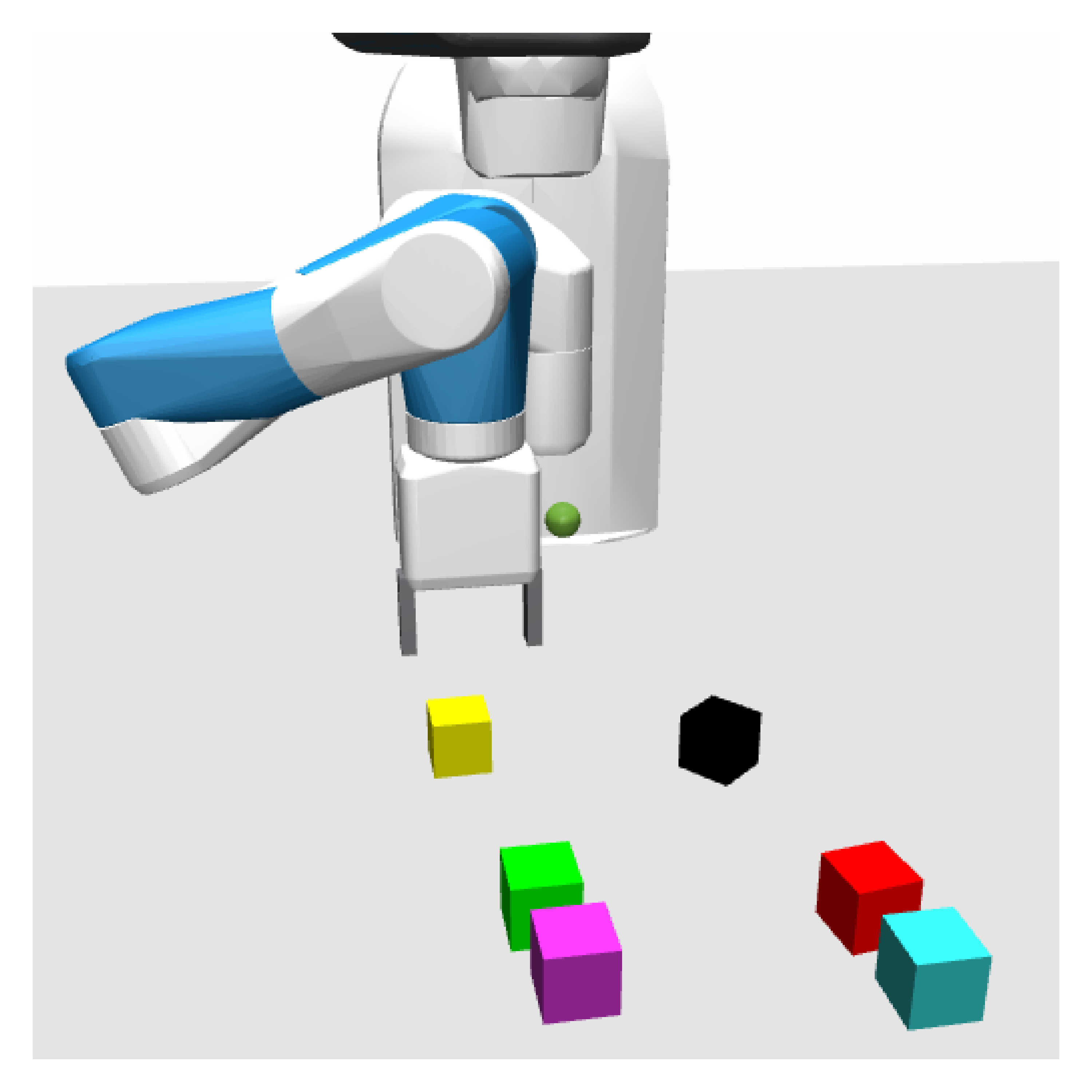}
     \end{subfigure}
    \begin{subfigure}[b]{0.15\textwidth}
         \centering
         \includegraphics[width=\textwidth]{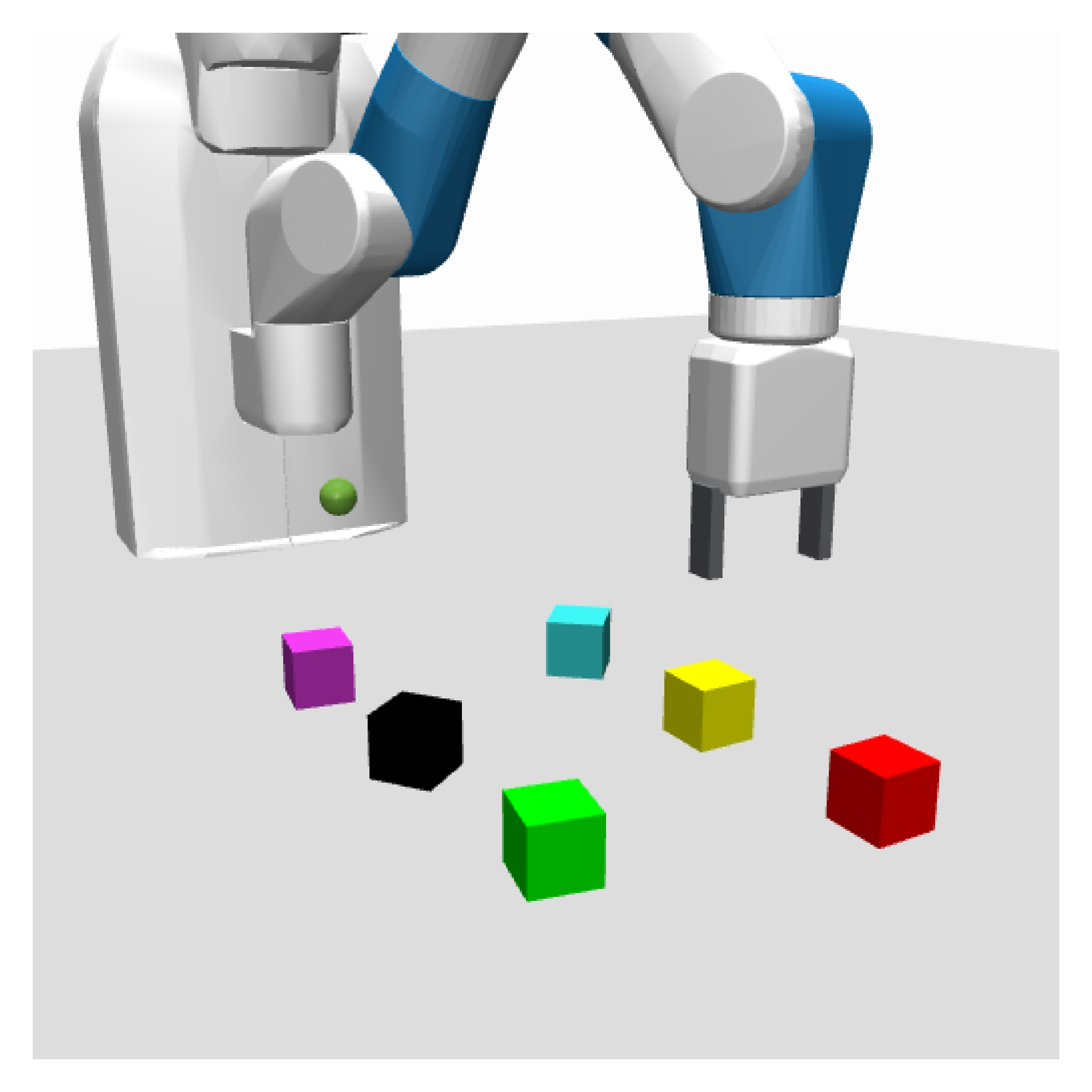}
     \end{subfigure}
     \vspace{-.3em}
  \caption{{\bf{Emerging patterns with \ourMethod{} on the $x$-$y$-$z$ subspace with GT models}}, where we use absolute relational $\phi$. }
       \label{fig:patterns_xyz}
\end{figure}

As discussed in \sec{sec:emerging_gt}, in our experiments we compute \ourMethod on the $x$-$y$ subspace of the object positions in \FPP to inject a bias towards vertical alignments. 
Examples of patterns generated when optimizing for \ourMethod using 
\begin{equation*}
    \phi(s_i,s_j) = \{(|\nint{s_{i,x} - s_{j,x}}|, |\nint{s_{i,y} - s_{j,y}}|,  |\nint{s_{i,z} - s_{j,z}}|)\}
\label{eqn:rair_xyz}
\end{equation*}
are showcased in \fig{fig:patterns_xyz}.
When we also include the $z$-positions of the objects in the \ourMethod computation, patterns and constellations on the ground are preferred. In this case, there is no difference between a horizontal line on the ground vs. a vertical line, \ie a stack. Since creating a stack, however, is a more sparse solution, in practice the zero-order trajectory optimizer converges already to regular structures on the ground and vertical constellations don't emerge. 
Starting from a randomly initialized scene with all objects on the ground, the regularity metric for $x$-$y$-$z$ only starts increasing when multiple objects are in the stack, which would require a very long planning horizon to find this solution.

\section{Experiment Results for Relational Case without Absolute Value} \label{app:bidirectional}
\begin{figure}[!h]
    \centering
    \vspace{1em}
    {\scriptsize
      \textcolor{ours_test_pure}{\rule[2pt]{20pt}{1.2pt}} \ourMethod{}
      \quad \textcolor{ours}{\rule[2pt]{20pt}{1.2pt}} \ourMethodA{} $(\lambda=0.1)$
      \quad \textcolor{ours_test_dis}{\rule[2pt]{20pt}{1.2pt}} \ourMethodA{} $(\lambda=1)$
       \quad \textcolor{cee_us_test}{\rule[2pt]{20pt}{1.2pt}} \oldMethod
    }\\
    \begin{subfigure}[b]{0.01\textwidth}
        \notsotiny{\rotatebox{90}{\hspace{.5cm}\textsf{relative time}}}
    \end{subfigure}
    \begin{subfigure}[t]{.24\linewidth}
    \includegraphics[width=\linewidth]{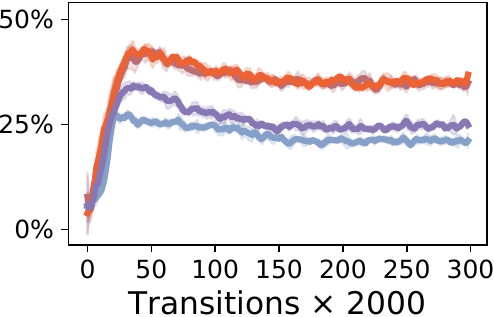}\vspace{-.3em}
    \caption{1 object moves}
    \end{subfigure}
    \hfill
    \begin{subfigure}[t]{.24\linewidth}
    \includegraphics[width=\linewidth]{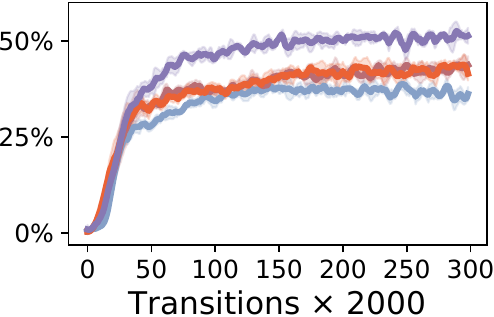}\vspace{-.3em}
    \caption{2 or more objects move}
    \end{subfigure}\hfill
    \begin{subfigure}[t]{.24\linewidth}
    \includegraphics[width=\linewidth]{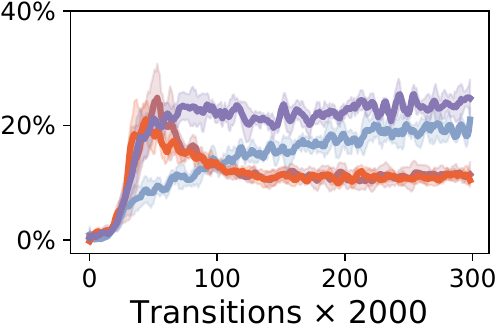}\vspace{-.3em}
    \caption{object(s) in air}
    \end{subfigure}
    \begin{subfigure}[t]{.24\linewidth}
    \includegraphics[width=\linewidth]{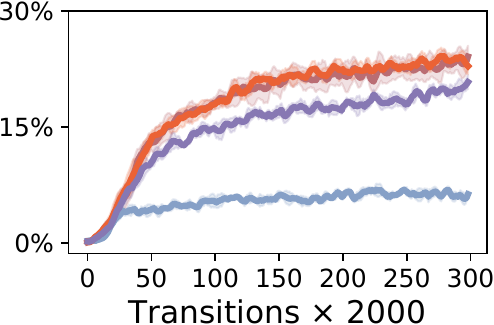}\vspace{-.3em}
    \caption{object(s) flipped}
    \end{subfigure}
    \caption{{\bf Comparison of interactions during free play in \FPP when combining ensemble disagreement with \ourMethod for different augmentation weights $\lambda$ with relational $\phi$}.
     Interaction metrics of free-play exploration count the relative amount of time steps spent in moving one object (a), moving two and more objects (b), moving objects in the air (c), and flipping object(s) (d). We used 5 independent seeds.
    }
    \label{fig:interactions_bidir:FPP}
\vspace{2em}
  \centering
    \begin{subfigure}[b]{0.136\textwidth}
         \centering
         \includegraphics[width=\textwidth]{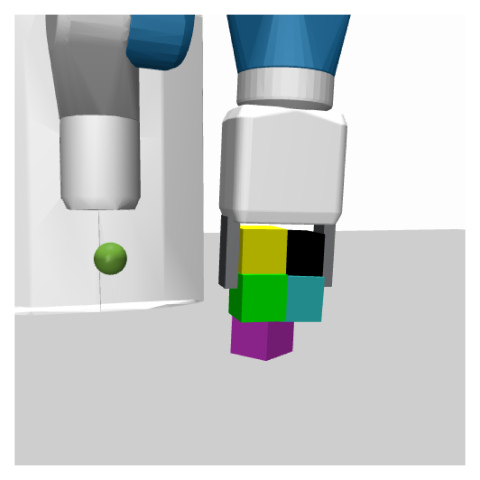}
        \centering{\scriptsize Iteration 245}
     \end{subfigure}
      \begin{subfigure}[b]{0.136\textwidth}
         \centering
         \includegraphics[width=\textwidth]{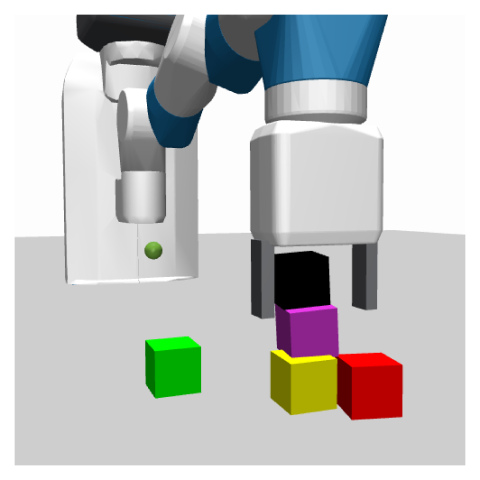}
        \centering{\scriptsize Iteration 246}
     \end{subfigure}
     \begin{subfigure}[b]{0.136\textwidth}
         \centering
         \includegraphics[width=\textwidth]{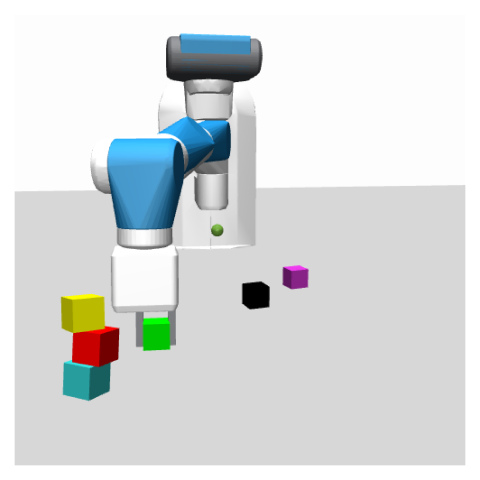}
        \centering{\scriptsize Iteration 247}
     \end{subfigure}
 \begin{subfigure}[b]{0.136\textwidth}
         \centering
         \includegraphics[width=\textwidth]{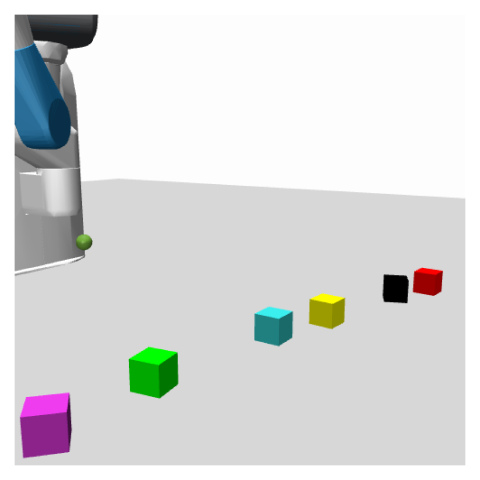}
        \centering{\scriptsize Iteration 248}
     \end{subfigure}
    \begin{subfigure}[b]{0.136\textwidth}
         \centering
         \includegraphics[width=\textwidth]{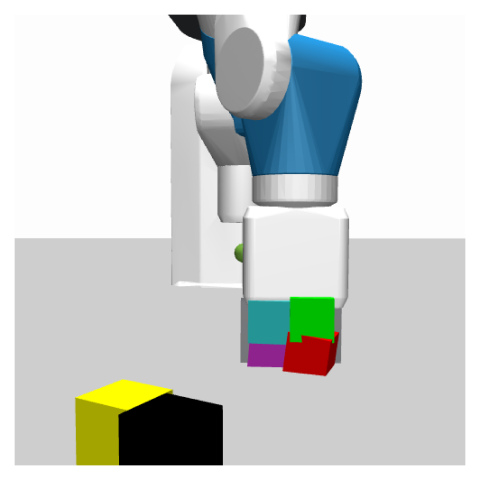}
        \centering{\scriptsize Iteration 270}
     \end{subfigure}
     \begin{subfigure}[b]{0.136\textwidth}
        \centering
         \includegraphics[width=\textwidth]{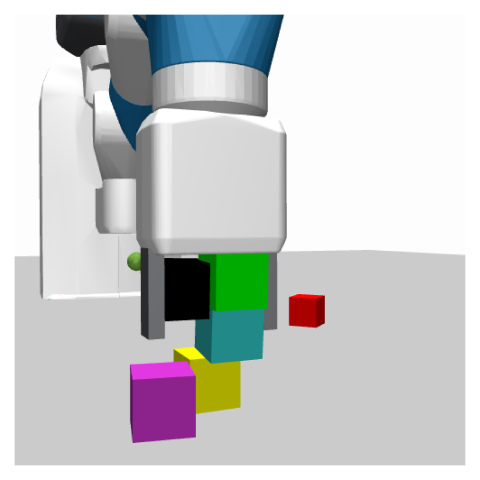}
        \centering{\scriptsize Iteration 275}
     \end{subfigure}
     \begin{subfigure}[b]{0.136\textwidth}
         \centering
         \includegraphics[width=\textwidth]{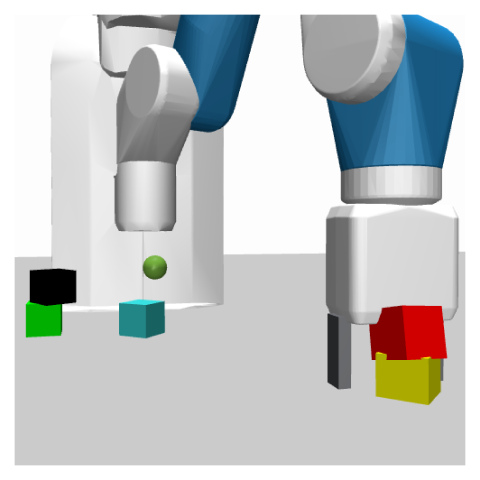}
        \centering{\scriptsize Iteration 295}
     \end{subfigure}\vspace{-.3em}
  \caption{{\bf{Snapshots from free play with \ourMethodA{} and relational $\phi$.}} We showcase snapshots of lowest entropy from exemplary rollouts at different iterations of free play. Following the regularity objective, stacks and alignments are generated. These snapshots come from a run with $\lambda=0.1$.}
       \label{fig:snapshots_freeplay_bidir}
\end{figure}

We present the interaction metrics observed during free play with \ourMethodA in the case of relational $\phi$ with $\phi(s_i,s_j) = \{\nint{s_{i} - s_{j}}\}=\{(\nint{s_{i,x} - s_{j,x}}, \nint{s_{i,y} - s_{j,y}})\}$. We find the interaction metrics to be comparable to the absolute relational case presented in the main paper with $\phi(s_i,s_j) = \{(|\nint{s_{i,x} - s_{j,x}}|, |\nint{s_{i,y} - s_{j,y}}|)\}$. In \fig{fig:interactions_bidir:FPP}, we also include the results for \ourMethodA with $\lambda=1$. In the case of the increased weighting on the ensemble disagreement term, the free-play behavior indeed collapses back onto \oldMethod. 
This means we have more flipping behavior and less air time. 
In the case of \ourMethodA with smaller $\lambda=0.1$, we seek regular states, which include vertical alignments, such that the air time doesn't go down. Note that in this case, there is still an incentive to ``destroy'' and lean towards chaos due to the ensemble disagreement reward term, such that the constellations showcased in \fig{fig:snapshots_freeplay} (snapshots for the absolute relational $\phi$) and \fig{fig:snapshots_freeplay_bidir} are not necessarily preserved.

We also evaluate the success rates for zero-shot downstream task generalization using the models trained in free-play runs with relational (R) $\phi$ and present them in \tab{tab:all_versions}. We find the performance in this case to be comparable to the absolute relational (AR) $\phi$ case.
 Note that in comparison to previous work \cite{Sancaktaretal22}, we use controller mode ``best'', not ``sum'' and a shorter planning horizon of 20 timesteps instead of 30. The environment is also intialized with 6 objects instead of 4. That's why the reported success rates differ from previous work. We also include the success rates for \oldMethod with mode ``sum'', and see that using mode ``best'' for free play also results in improved performance in assembly tasks.

\begin{table}[h]
    \centering
  \caption{{\bf Zero-shot downstream task generalization performance of \ourMethodA for different $\phi$ and $\lambda$} for assembly tasks as well as the generic pick \& place task and the more chaos-oriented throwing and flipping. Results are shown for five independent seeds. AR: Absolute relative $\phi$, R: Relative $\phi$. In the row \oldMethod (sum), we report success rates when we use the controller mode \emph{sum} instead of \emph{best}, which is the default for \FPP in this work. In the bottom row, we report the success rates achieved via planning with ground truth models. This is to provide a baseline for how hard the task is to solve with finite-horizon planning and potentially suboptimally designed task rewards.}
  \centering
    \renewcommand{\arraystretch}{.99}
    \resizebox{1.\linewidth}{!}{ %
    \begin{tabular}{@{}lcccc|ccc@{}}
    \toprule
    & Singletower& Multitower & Pyramid  & Pyramid  & Pick\&Place & Throw & Flip \\
    & 3 & 2+2 &  5 &  6 &  6 &  4 &  4\\
    \midrule
    \ourMethodA (R) & $\mathbf{0.80 \pm 0.07}$  & $\mathbf{0.77 \pm 0.03}$ & $\mathbf{0.47 \pm 0.04}$ & $\mathbf{0.17 \pm 0.05}$ &
     $\mathbf{0.90 \pm 0.01}$ &$0.38 \pm 0.02$ & $0.63 \pm 0.05$ \\
    \ourMethodA (AR) & $\mathbf{0.75 \pm 0.07}$  & $\mathbf{0.77 \pm 0.06}$ & $\mathbf{0.49 \pm 0.06}$ & $\mathbf{0.18 \pm 0.04}$ &
     $\mathbf{0.90 \pm 0.02}$ &$0.32 \pm 0.02$ & $0.63 \pm 0.08$ \\
    \ourMethod (AR) ($\lambda=0$) & $0.64 \pm 0.03$  & $0.62 \pm 0.03$ & $0.25 \pm 0.05$ &  $0.10 \pm 0.02$ &
     $0.74 \pm 0.05$ &$0.21 \pm 0.01$ & $0.65 \pm 0.1$ \\
     \oldMethod   & $0.40 \pm 0.12$ & $0.52 \pm 0.05$  & $0.14 \pm 0.09$  & $0.02 \pm 0.01$
     & $\mathbf{0.90 \pm 0.02}$ & $\mathbf{0.49 \pm 0.05}$ & $0.73 \pm 0.1$ \\
    \midrule
     \oldMethod (sum)  & $0.26 \pm 0.11$ & $0.48 \pm 0.13$  & $0.10 \pm 0.10$  & $0.0 \pm 0.0$
     & $\mathbf{0.91 \pm 0.02}$ & $\mathbf{0.51 \pm 0.04}$ & $0.76 \pm 0.15$ \\
    \midrule
    GT   & $0.99$ & $0.97$ & $0.82$  & $0.81$  & $0.99$ & $0.97$ & 1.0  \\
  \bottomrule
    \end{tabular}
  }
    \label{tab:all_versions}
\end{table}

\section{Experiment Results for Free Play with pure \ourMethod} \label{app:pure_rair}

In this section, we further discuss the zero-shot downstream task generalization performance for free play with pure \ourMethod and the role of the information-gain term in our intrinsic reward combination used in free play, as specified in \eqn{eqn:combination}.
As discussed in \sec{sec:free_play_w_rair}, adding ensemble disagreement to our regularity objective leads to 1) more interaction-rich free play and 2) more robust world models which yield higher success rates for zero-shot downstream task generalization. For both the absolute relational case presented in \fig{fig:interactions:FPP} and the relational case in \fig{fig:interactions_bidir:FPP}, \ourMethod with no disagreement term yields less interactions in terms of object(s) being moved, being in air and being flipped. This is because the exploration problem is not solved by \ourMethod alone. When we use ensemble disagreement as an intrinsic reward, the discovery of different types of interactions is accelerated. When one of the models in the ensemble learns a new type of dynamics, such as an object moving, the ensemble disagreement goes up, incentivizing the agent to repeat this behavior until it is learned by all models such that disagreement goes down. In the case of \ourMethod, this only happens implicitly: with some models in the ensemble learning a certain type of dynamics in the environment, during planning, the models can hallucinate objects being aligned and creating a regular pattern with high \ourMethod such that these actions are executed by the controller. These false attempts also help exploration.

As the models produce better predictions, especially after free play iteration 200, we observe that more stable patterns are generated with \ourMethod compared to \ourMethodA (\fig{fig:rair_best_values}) and the amount of time objects are moving starts decreasing {\fig{fig:interactions:FPP}}. This is because in this case when the models get better and hallucinate less, there is no reason to leave local optima such as a spaced line unless a pattern that yields a higher regularity value can be found within the planning horizon.

The challenge of exploration with pure \ourMethod is also reflected in the interaction time for object(s) in air. Starting to create regular patterns such as stacks takes longer, as exploring to lift objects happens later without the disagreement reward. This is also connected to the step-wise landscape of \ourMethod as discussed in \sec{sec:free_play_w_rair} such that explicit exploration via ensemble disagreement is beneficial.

\begin{figure}[t]
  \centering
    \begin{subfigure}[b]{0.136\textwidth}
         \centering
         \includegraphics[width=\textwidth]{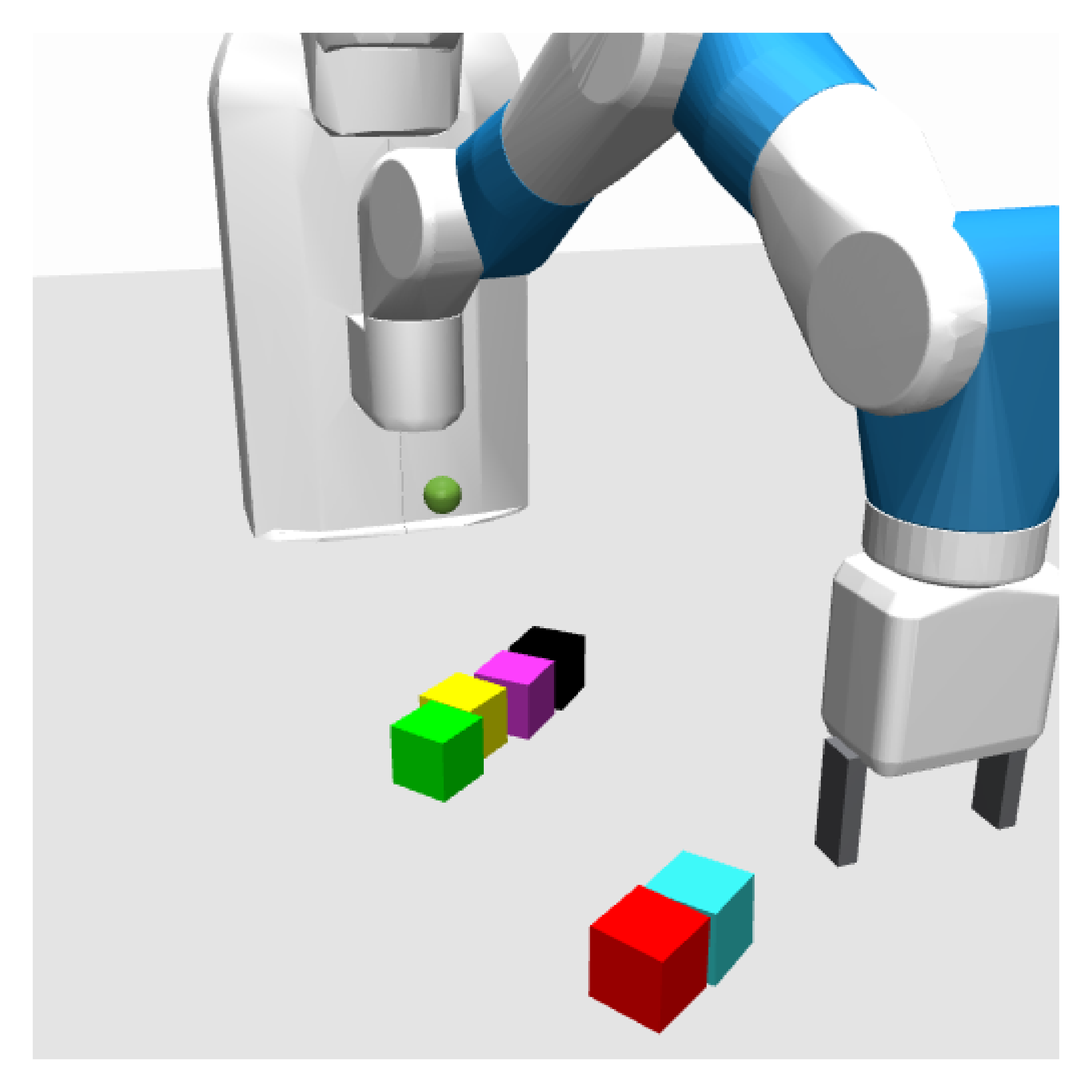}
        \centering{\scriptsize Iteration 270}
     \end{subfigure}
      \begin{subfigure}[b]{0.136\textwidth}
         \centering
         \includegraphics[width=\textwidth]{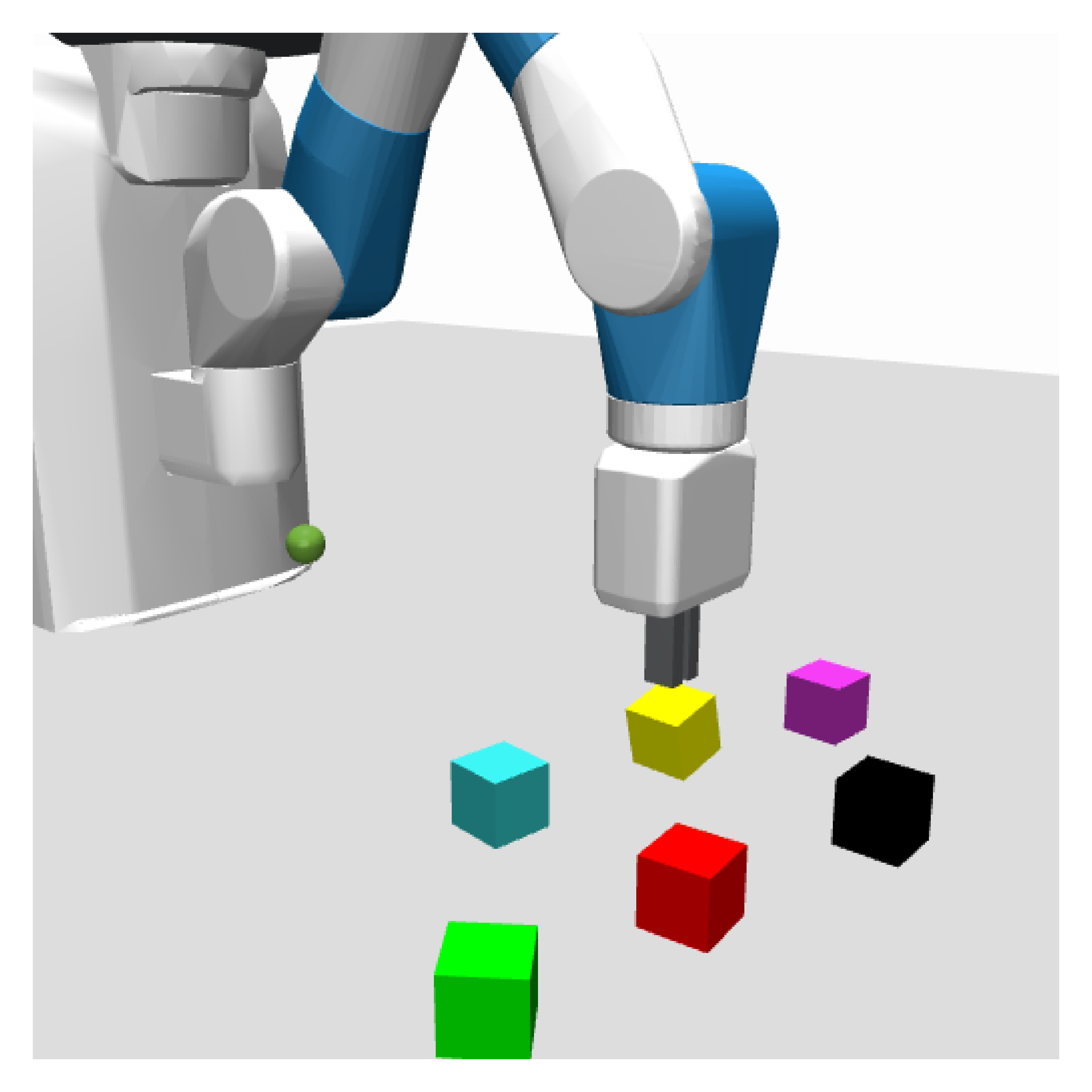}
        \centering{\scriptsize Iteration 273}
     \end{subfigure}
     \begin{subfigure}[b]{0.136\textwidth}
         \centering
         \includegraphics[width=\textwidth]{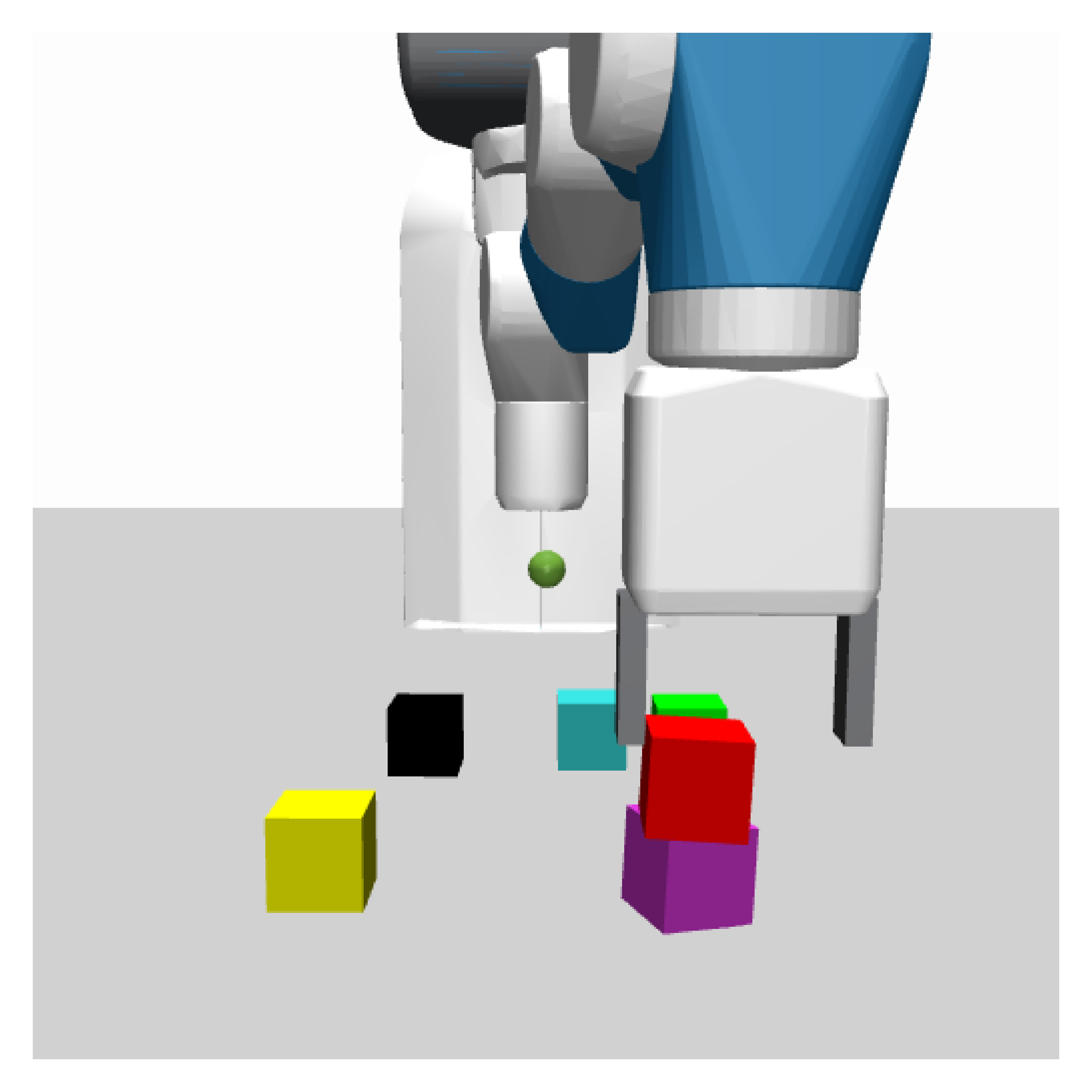}
        \centering{\scriptsize Iteration 276}
     \end{subfigure}
 \begin{subfigure}[b]{0.136\textwidth}
         \centering
         \includegraphics[width=\textwidth]{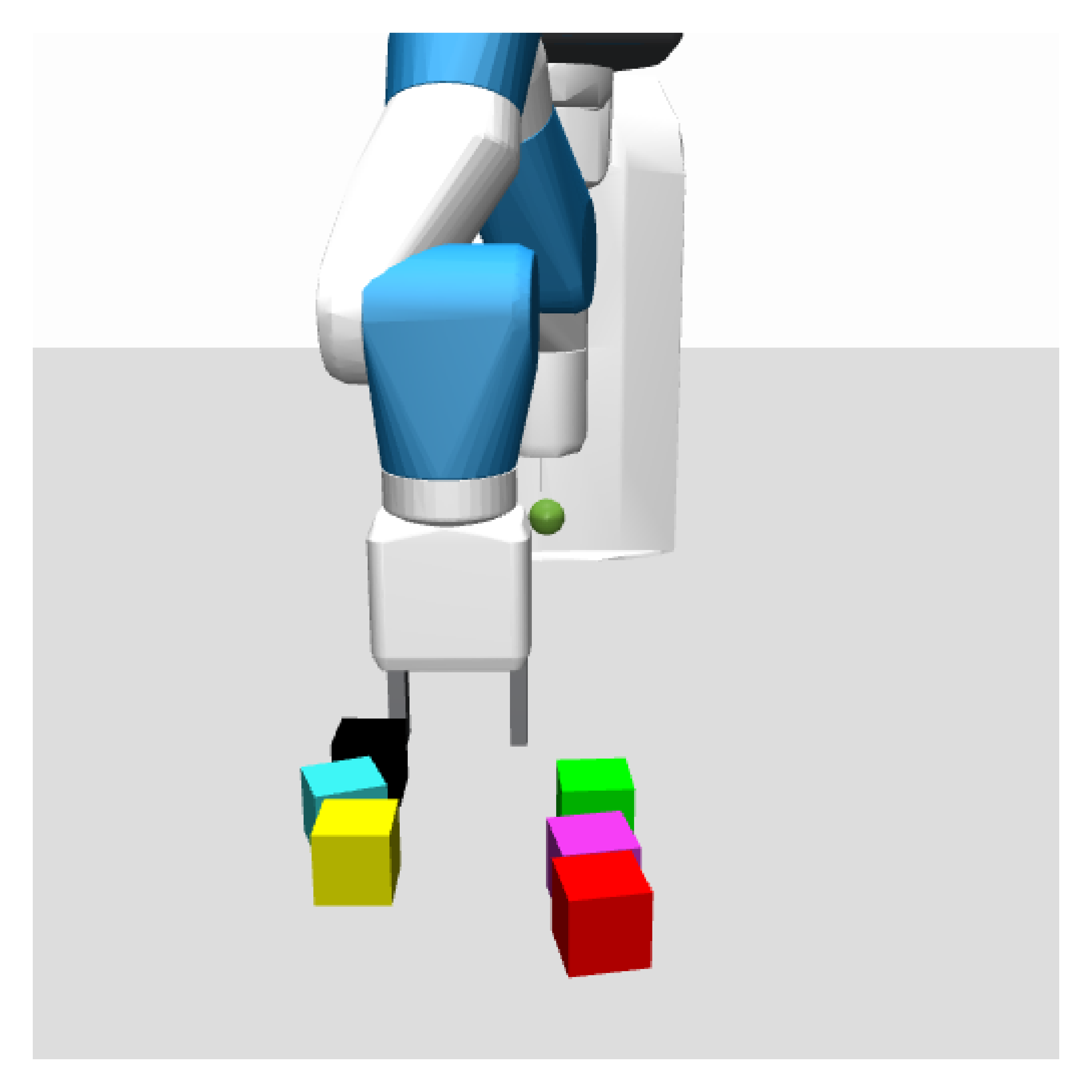}
        \centering{\scriptsize Iteration 295}
     \end{subfigure}
    \begin{subfigure}[b]{0.136\textwidth}
         \centering
         \includegraphics[width=\textwidth]{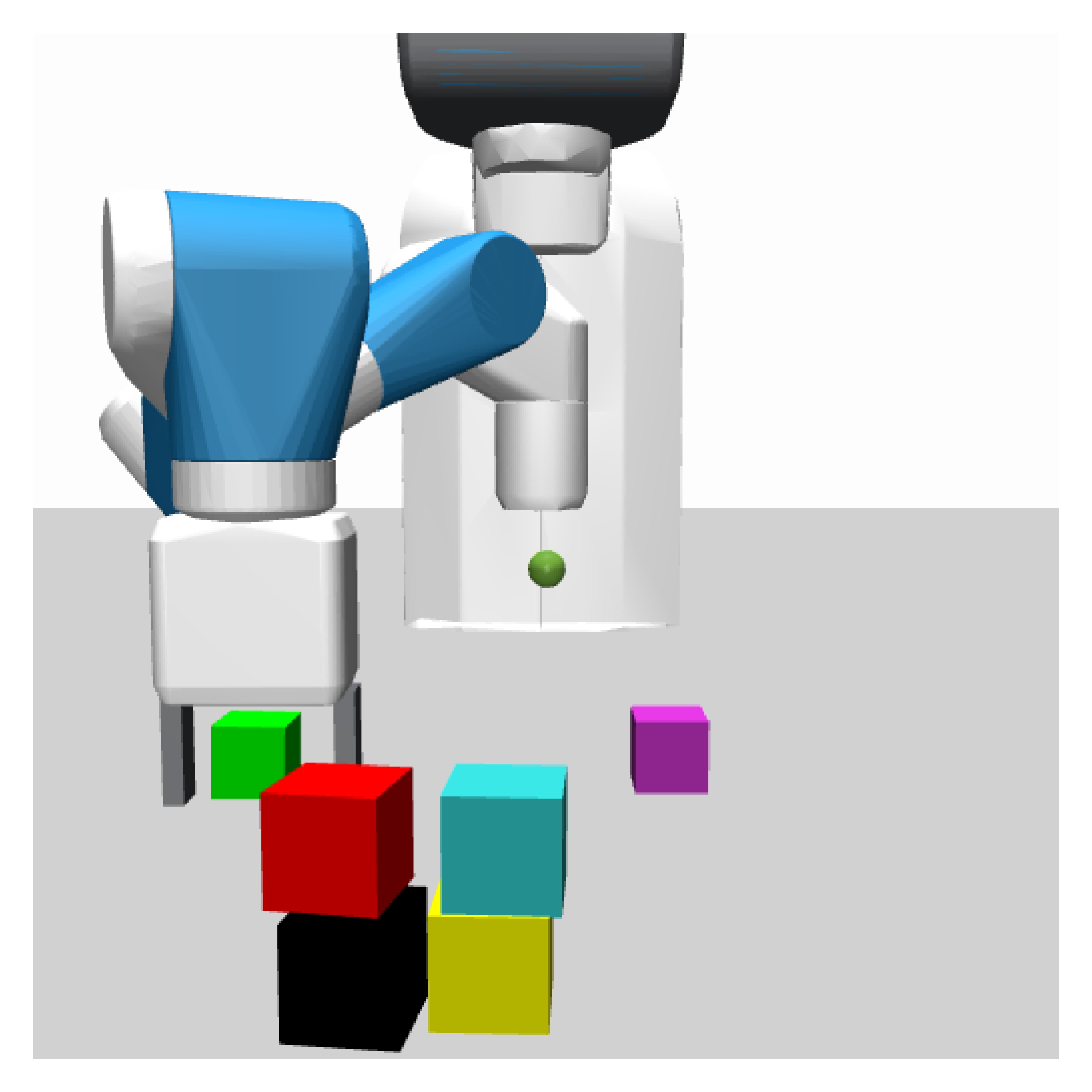}
        \centering{\scriptsize Iteration 296}
     \end{subfigure}
     \begin{subfigure}[b]{0.136\textwidth}
        \centering
         \includegraphics[width=\textwidth]{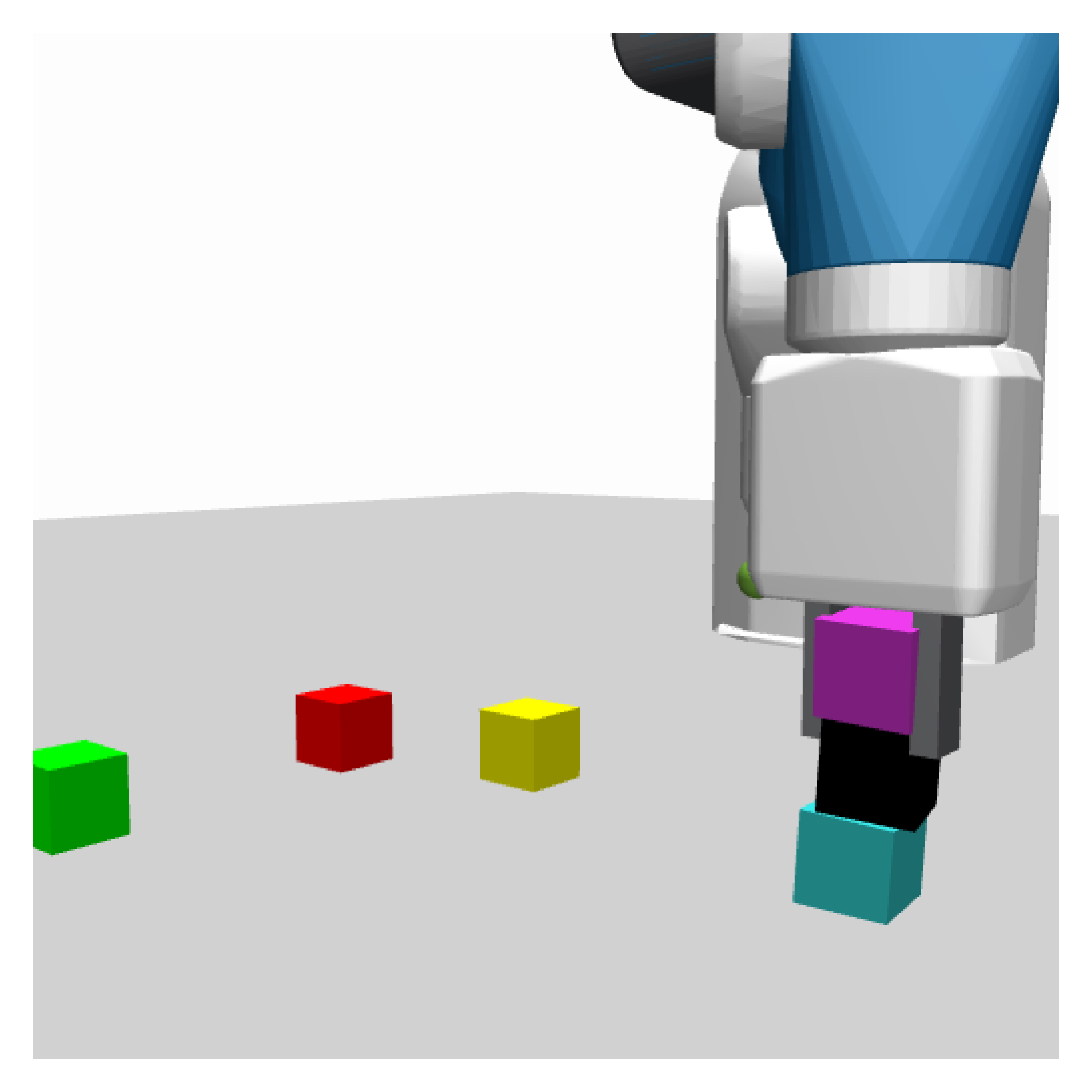}
        \centering{\scriptsize Iteration 298}
     \end{subfigure}
     \begin{subfigure}[b]{0.136\textwidth}
         \centering
         \includegraphics[width=\textwidth]{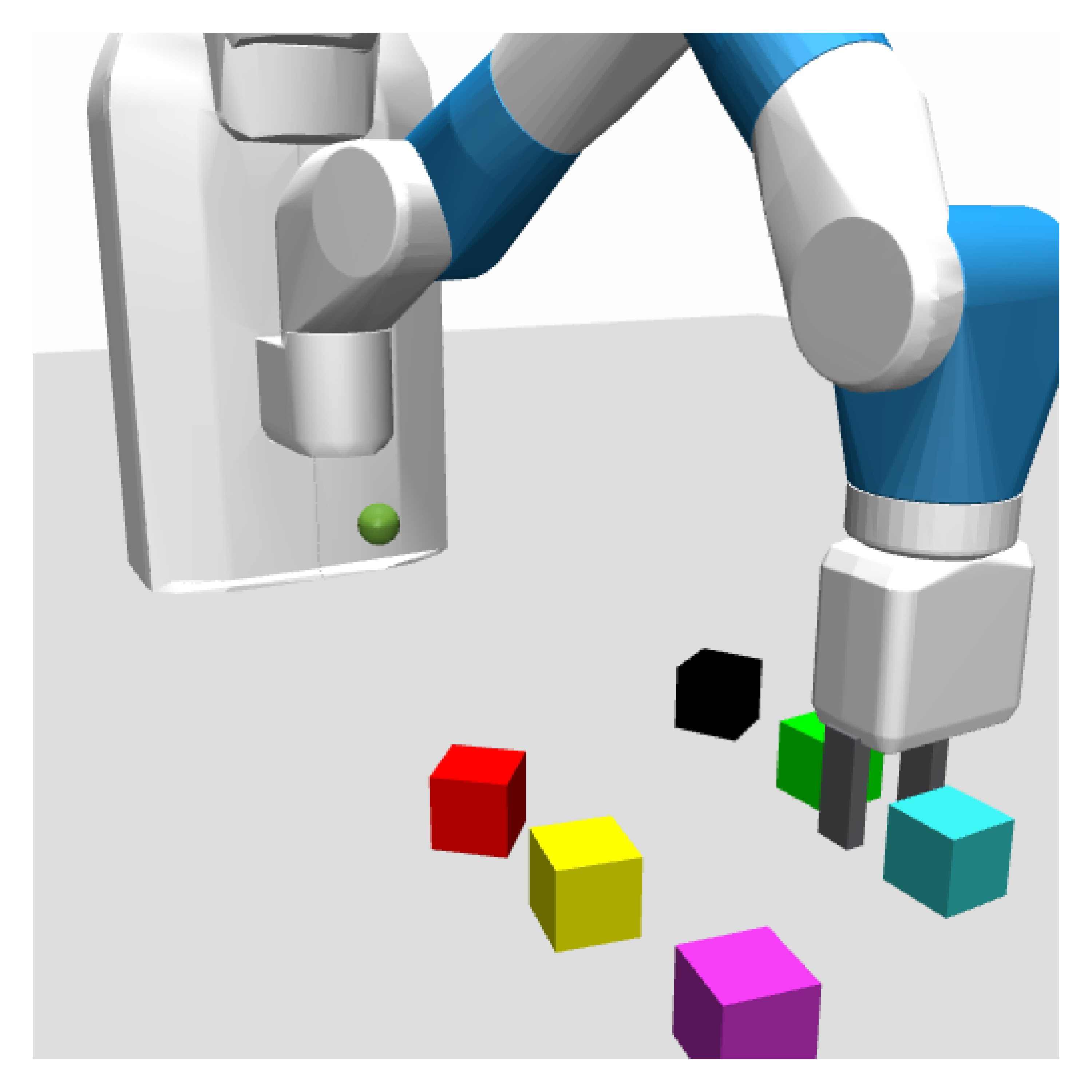}
        \centering{\scriptsize Iteration 299}
     \end{subfigure}\vspace{-.3em}
  \caption{{\bf{Snapshots from free play with pure \ourMethod{}.}} We showcase snapshots of lowest entropy from exemplary rollouts at different iterations of free play. These snapshots come from a run with absolute relational $\phi$.}
       \label{fig:snapshots_freeplay_pure}
\end{figure}
As showcased in \fig{fig:snapshots_freeplay_pure}, we still observe the stable generation of patterns such as spaced lines later on in training as well, as these are local optima of \ourMethod. However, we start to see more stacks generated in the later stages of free play. In \fig{fig:rair_best_values}, the highest \ourMethod value achieved for the different variants are showcased throughout training. Pure \ourMethod, achieves slightly higher regularity then \ourMethodA. This is also because pure \ourMethod, tends to generate more regular patterns that feature all objects, \ie all objects are in-line or build a square. With \ourMethodA, as some chaotic behavior is injected to free play via ensemble disagreement, more local regularities such as a stack of 2, with the rest of the objects in disorder, are likely to emerge.

Through injecting ensemble disagreement into free play, the robustness of the learned world models is also increased as they are guided by their own epistemic uncertainty \cite{Sancaktaretal22}. During free play, data is actively collected from regions where the models are uncertain, acting as their own adversary. This in turn makes the models more robust for deployment in model-based planning in the follow-up extrinsic phase, where the accuracy of model predictions is paramount for good performance. This is reflected in the downstream task performance evaluations in \fig{fig:building_success}, where \ourMethodA consistently outperforms both \ourMethod and \oldMethod in the assembly tasks. Note that as regularity explicitly favours alignments such as stacks, unlike \oldMethod, these dynamics are explored better, leading to higher success rates. This also showcases the importance of guiding free play towards regularity.
In \fig{fig:chaotic_tasks_pure_rair}, the results for the pick \& place, throwing and flipping tasks are shown. Due to the increased robustness of the model with the disagreement term, we indeed observe better performance for \ourMethodA and \oldMethod for the Pick \& Place task compared to pure \ourMethod. This is also true for the throwing task. However, another contributing factor here is that models with disagreement favor more chaotic behaviors and perform more ``throwing''-like behaviors during free play. As \oldMethod has no bias towards regularity, it performs best, whereas pure \ourMethod performs worse than \ourMethodA. Interestingly, for the flipping 4 objects task we found performance for \ourMethod and \ourMethodA to be comparable despite the significantly reduced amount of time spent flipping objects in the case of pure \ourMethod, as can be seen in \fig{fig:interactions:FPP}. Upon inspecting the data generated during free-play, we hypothesize this is because unlike \ourMethodA, which flips and rolls objects together in a chaotic way, we found \ourMethod to produce more isolated flipping of individual objects.
\begin{figure}[!h]
    \centering
    {\scriptsize
      \textcolor{ours_test_pure}{\rule[2pt]{20pt}{1.2pt}} \ourMethod{}
      \quad \textcolor{ours}{\rule[2pt]{20pt}{1.2pt}} \ourMethodA{} %
       \quad \textcolor{cee_us_test}{\rule[2pt]{20pt}{1.2pt}} \oldMethod
      \quad \textcolor{ourorange}{\rule[2pt]{20pt}{1.2pt}} RND
      \quad \textcolor{ourgreen}{\rule[2pt]{20pt}{1.2pt}} Dis
    }\vspace{.1em}\\
    \begin{subfigure}[b]{0.01\textwidth}
        \notsotiny{\rotatebox{90}{\hspace{.5cm}\textsf{success rate}}}
    \end{subfigure} 
    \begin{subfigure}[t]{.24\linewidth}
    \includegraphics[width=\linewidth]{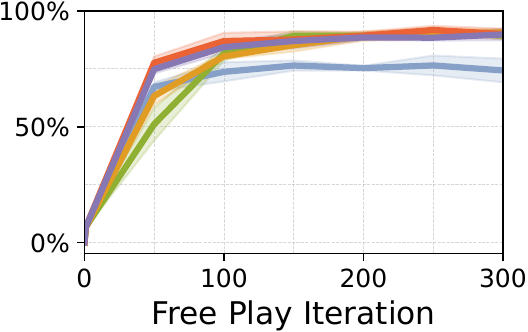}\vspace{-.3em}
    \caption{Pick \& Place 6}
    \end{subfigure}
    \begin{subfigure}[t]{.24\linewidth}
    \includegraphics[width=\linewidth]{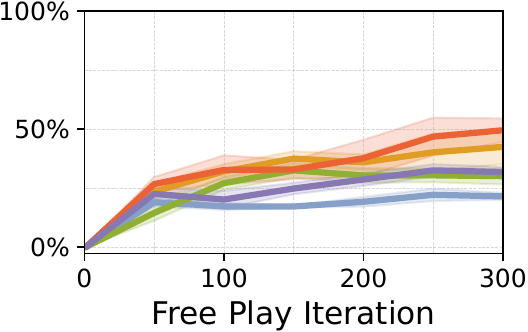}\vspace{-.3em}
    \caption{Throw 4}
    \end{subfigure}
    \begin{subfigure}[t]{.24\linewidth}
    \includegraphics[width=\linewidth]{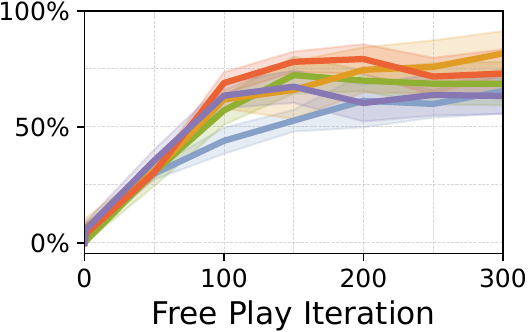}\vspace{-.3em}
    \caption{Flip 4}
    \end{subfigure}
    \caption{{\bf Downstream Task Performance for Pick \& Place and the more chaotic tasks of throwing and flipping} with only \ourMethod ($\lambda=0$), \ourMethodA ($\lambda=0.1$), \oldMethod, RND and Disagreement. We use absolute relational $\phi$ for \ourMethod computations. Results are shown for 5 independent seeds.
    }
    \label{fig:chaotic_tasks_pure_rair}
\end{figure}

\section{Experiment Results for Re-creating Existing Patterns} \label{app:recreation}
As presented in \sec{sec:recreation}, we test whether we can re-create existing regularities in the environment by simply optimizing for \ourMethod with iCEM, using ground truth models. As for the pyramid with 3 objects in \sec{sec:recreation}, we initialize different regular structures outside of the robot's manipulability range and test whether these regular patterns can be re-created, merely by maximizing for \ourMethod. We test for the re-creation of a singletower with 3 and 4 objects, 2 towers with 2 objects each (referred to as multitower 2+2), as well as a spaced line and a rhombus with 4 objects. We test this with ground truth models for 15 independent rollouts for each structure and report the re-creation rates. Example rollouts are illustrated in \fig{fig:recreation_more}. Note that due to the limited sample-budget with iCEM and the finite-horizon, we don't necessarily converge to the global minima, which corresponds to the full recreation of the structure. However, in all of the tested cases, the generated structures repeat at least one prominent sub-structure present in the underlying regular constellation by optimizing for \ourMethod.

For \emph{Singletower 3}, the entire stack of 3 gets recreated 73\% of the time. A partial recreation with a stack of 2 blocks is observed in all but one of the remaining cases.

When a \emph{Singletower 4} is initialized outside of the robot's range, the full tower with 4 blocks gets recreated 40\% of the time. In the remaining cases, either a tower of 3 (33\%) or towers of 2 (27\%) are built.

For the challenging \emph{Multitower 2+2} case, the two towers are built, with the same distance to each other as in the original pattern, 20\% of the time. An example of this ``complete'' recreation is illustrated in \fig{fig:multitower_full}. Otherwise, 53\% of the time a stack of 2 is built (\fig{fig:multitower_partial}) or a spaced line repeating the relative position of the two towers in the original pattern.

For the patterns on the ground, namely \emph{Spaced Line} and \emph{Rhombus}, the recreation rates are higher since the exploration problem is less prominent. At least 75\% of the original pattern is re-created at each rollout, \ie for the case of 4 objects, at least 3 objects follow the original pattern.
The complete \emph{Spaced Line} is recreated 80\% and the entire rhombus 73\% of the test rollouts. 

In these experiments, we use \ourMethod with $\phi(s_i,s_j) = \{(|\nint{s_{i,x} - s_{j,x}}|, |\nint{s_{i,y} - s_{j,y}}|, |\nint{s_{i,z} - s_{j,z}}|)\}$. This is because in the case of existing structures in the scene, we don't need/want to inject any biases, \eg towards vertical alignments, into the optimization. As the existing pattern is outside of the manipulability range of the robot, re-creating the pattern becomes a direct global optimum for \ourMethod, as all regularities reoccur. However, if we restrict ourselves to the $x$-$y$ subspace, this is no longer the case: even for a rhombus, a stack built with the blocks in-reach becomes the global optimum. This is because all the blocks in the stack then have the same $x$-$y$ relation to the blocks in the rhombus.

\begin{figure}
\begin{subfigure}[t]{.48\linewidth}
    \centering
    {\scriptsize
      Start
       \hspace{10em} End
    }
    \includegraphics[width=.47\linewidth]{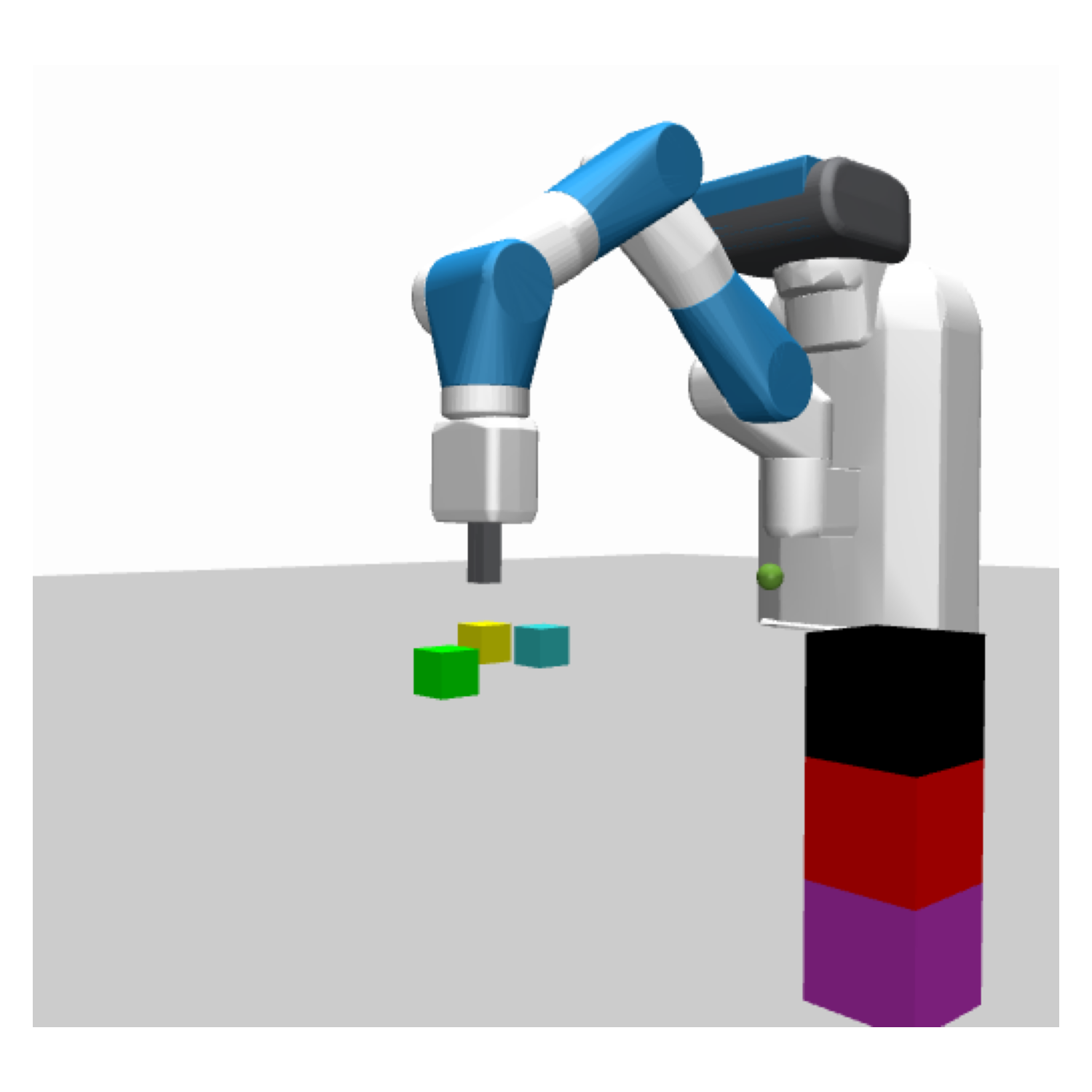}
    \includegraphics[width=.47\linewidth]{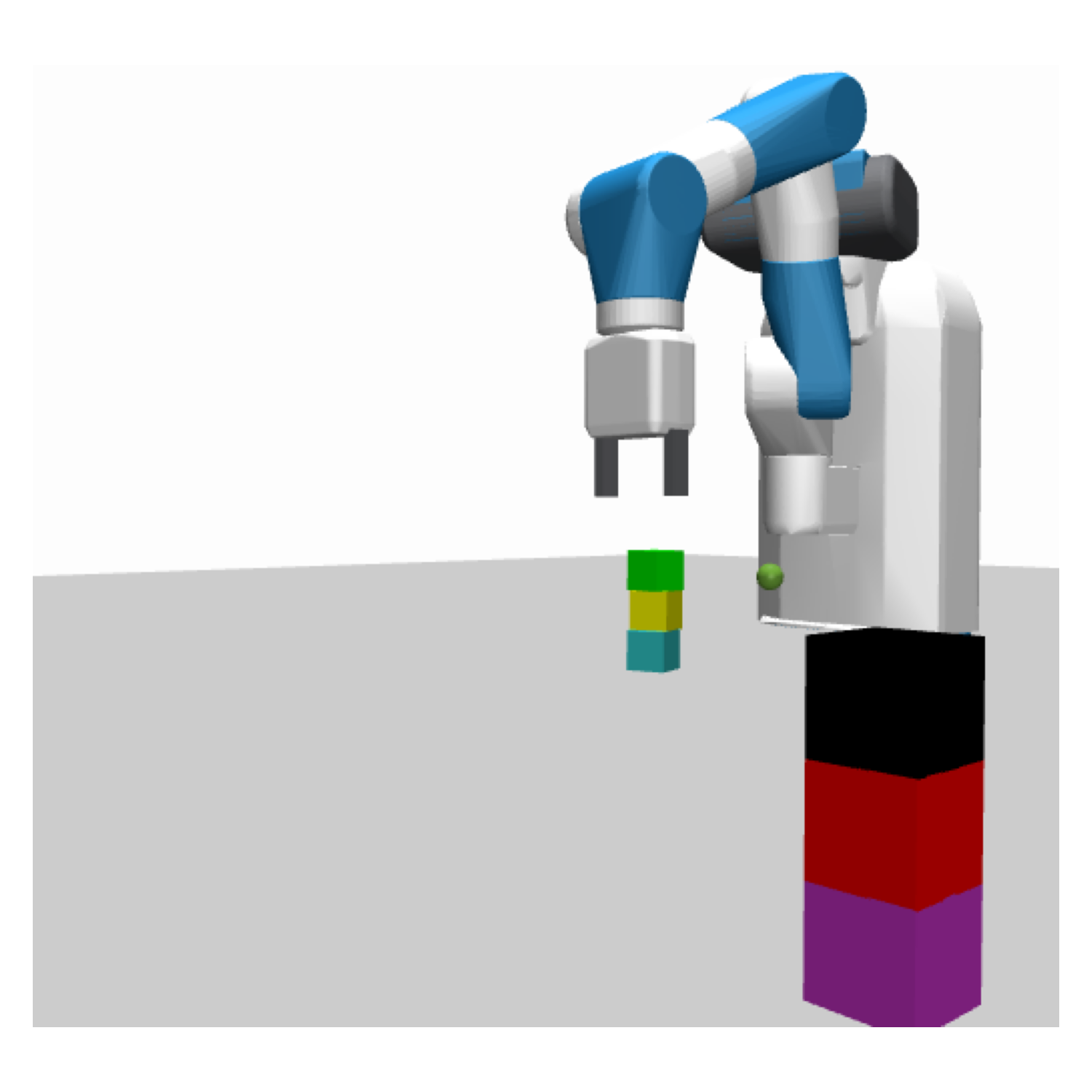}
\vspace{-.3em}
\caption{Singletower 3}
 \end{subfigure}
\begin{subfigure}[t]{.48\linewidth}
    \centering
    {\scriptsize
      Start
       \hspace{10em} End
    }
    \includegraphics[width=.47\linewidth]{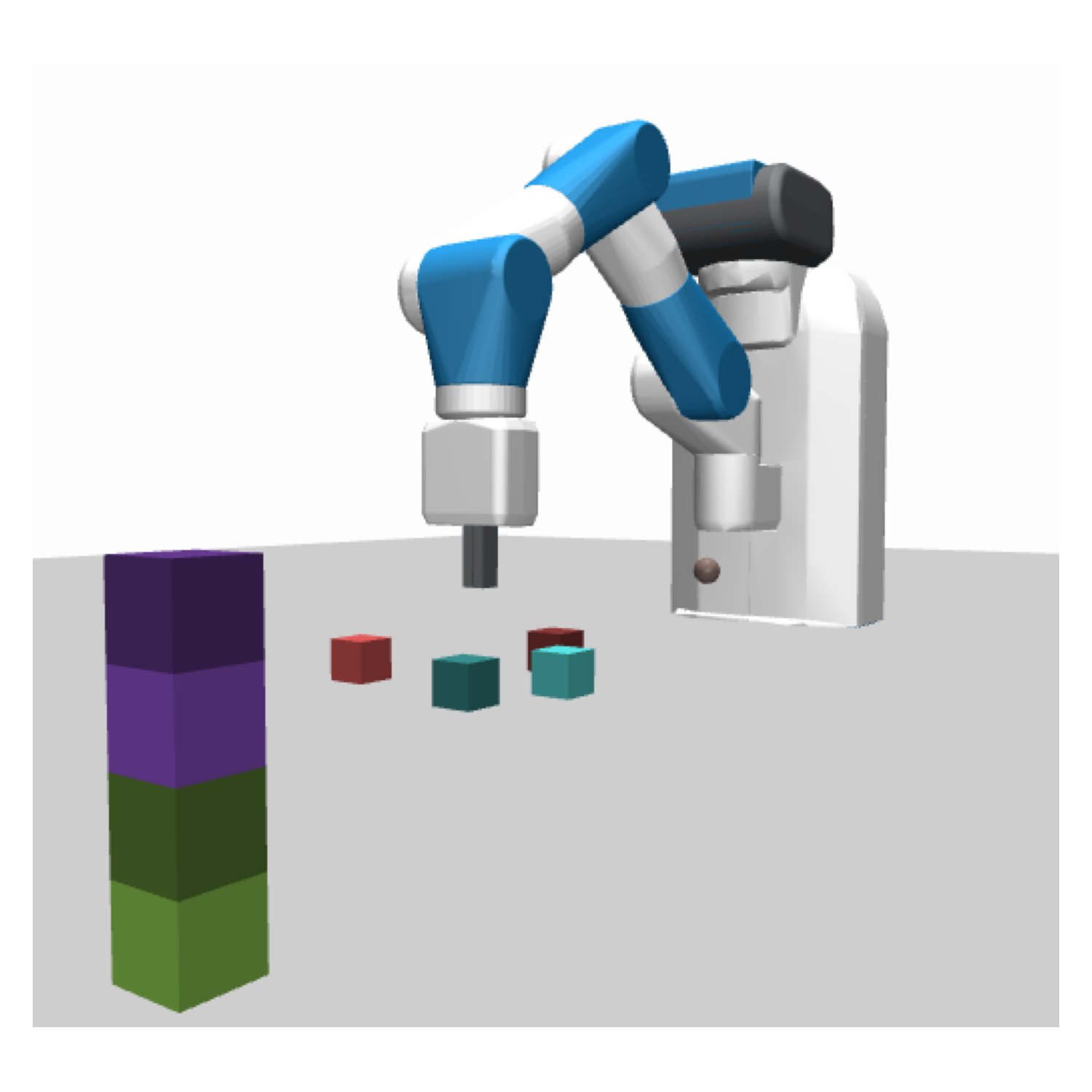}
    \includegraphics[width=.47\linewidth]{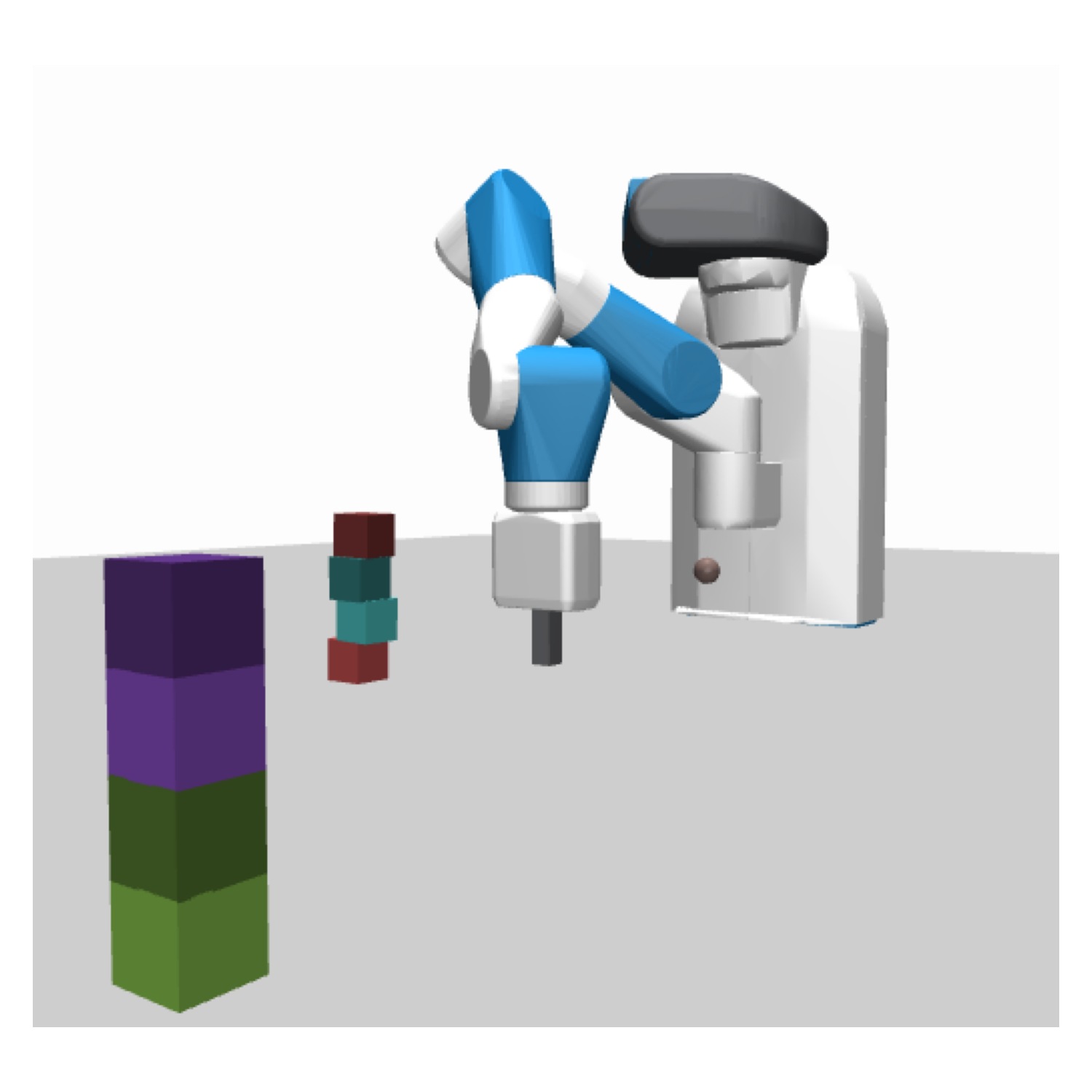}
    \vspace{-.6em}
\caption{Singletower 4}
\end{subfigure}
\\
\\
\begin{subfigure}[t]{.48\linewidth}
    \centering
    {\scriptsize
      Start
       \hspace{10em} End
    }
    \includegraphics[width=.47\linewidth]{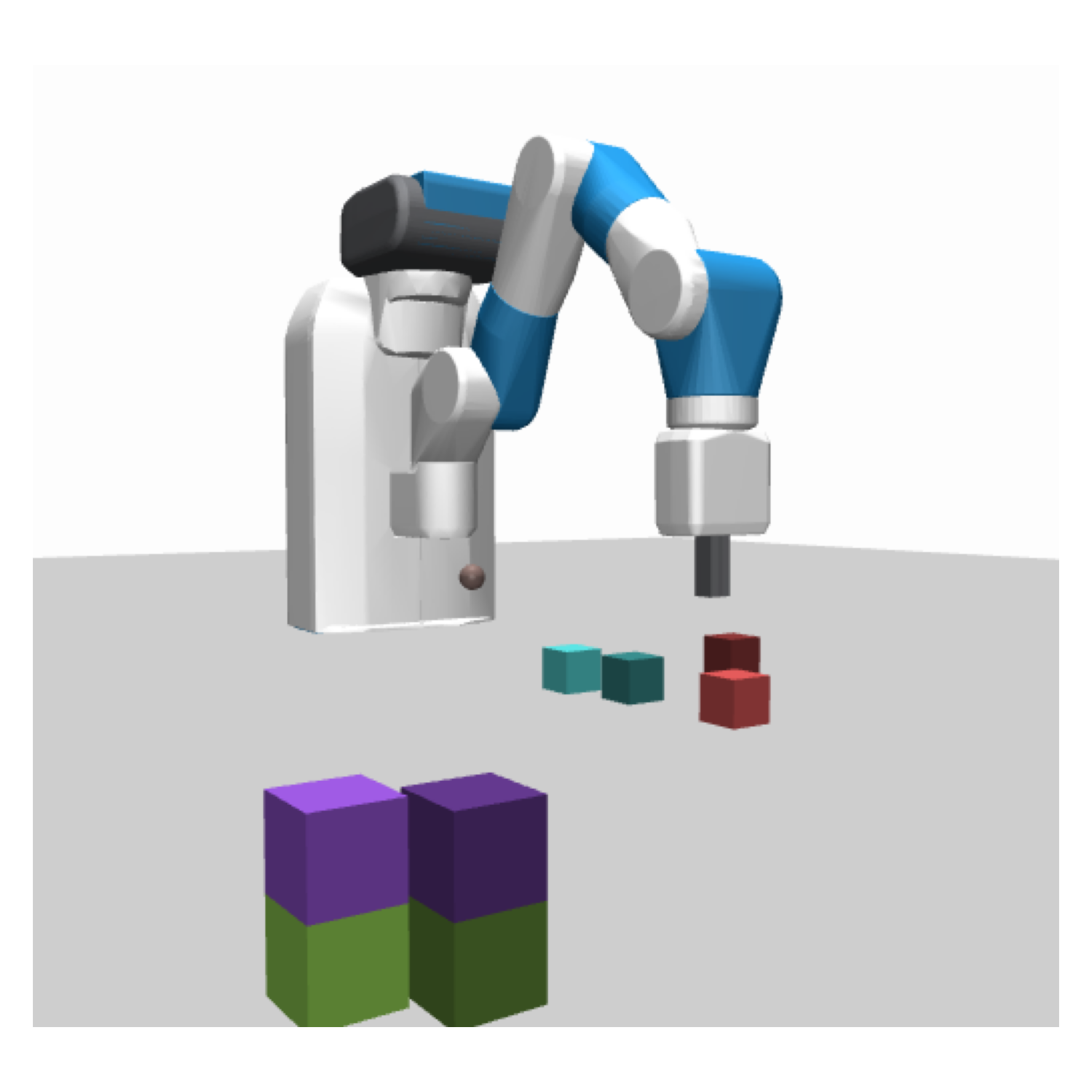}
    \includegraphics[width=.47\linewidth]{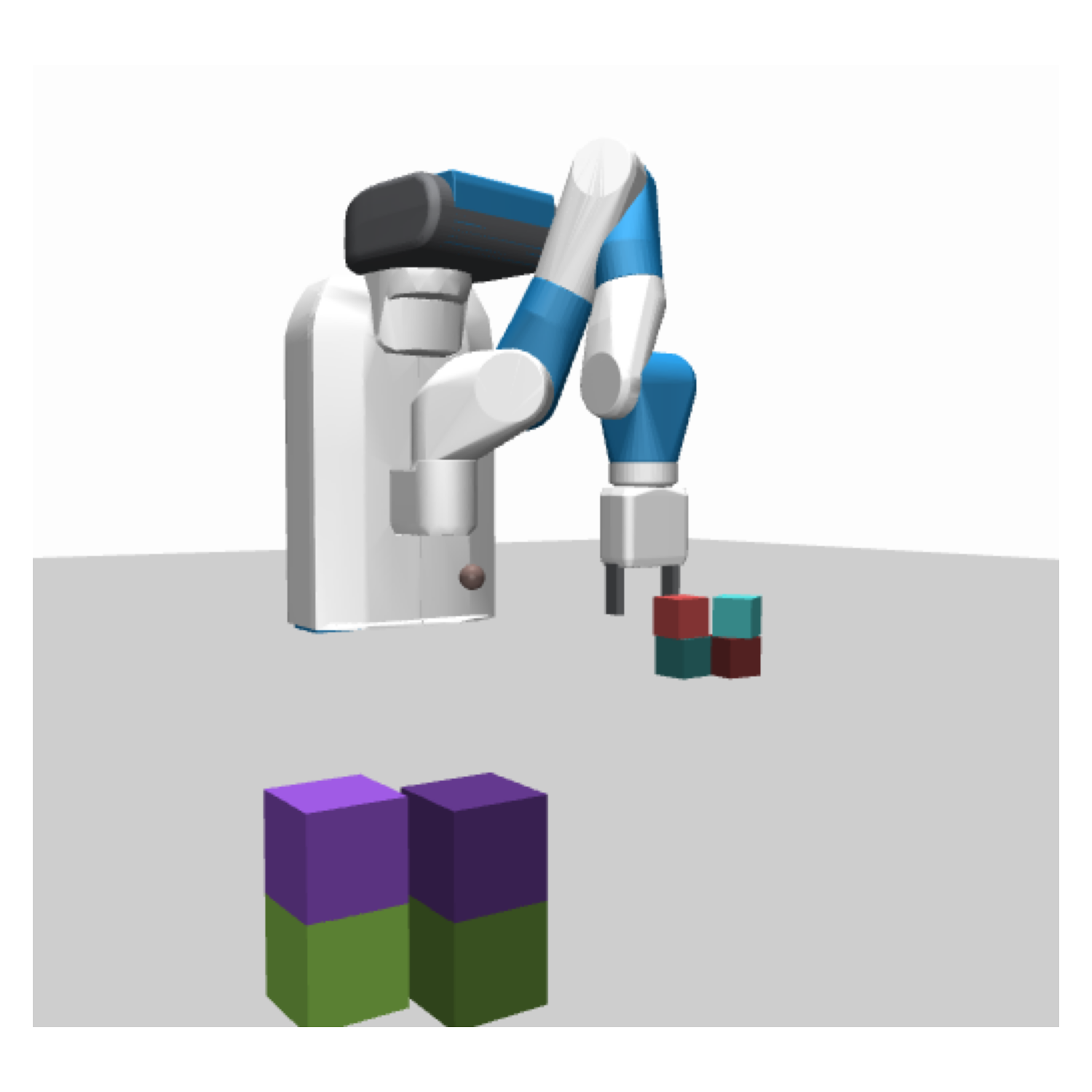}
    \vspace{-.6em}
\caption{Multitower 2+2}
\label{fig:multitower_full}
 \end{subfigure}%
\begin{subfigure}[t]{.48\linewidth}
    \centering
    {\scriptsize
      Start
       \hspace{10em} End
    }
    \includegraphics[width=.47\linewidth]{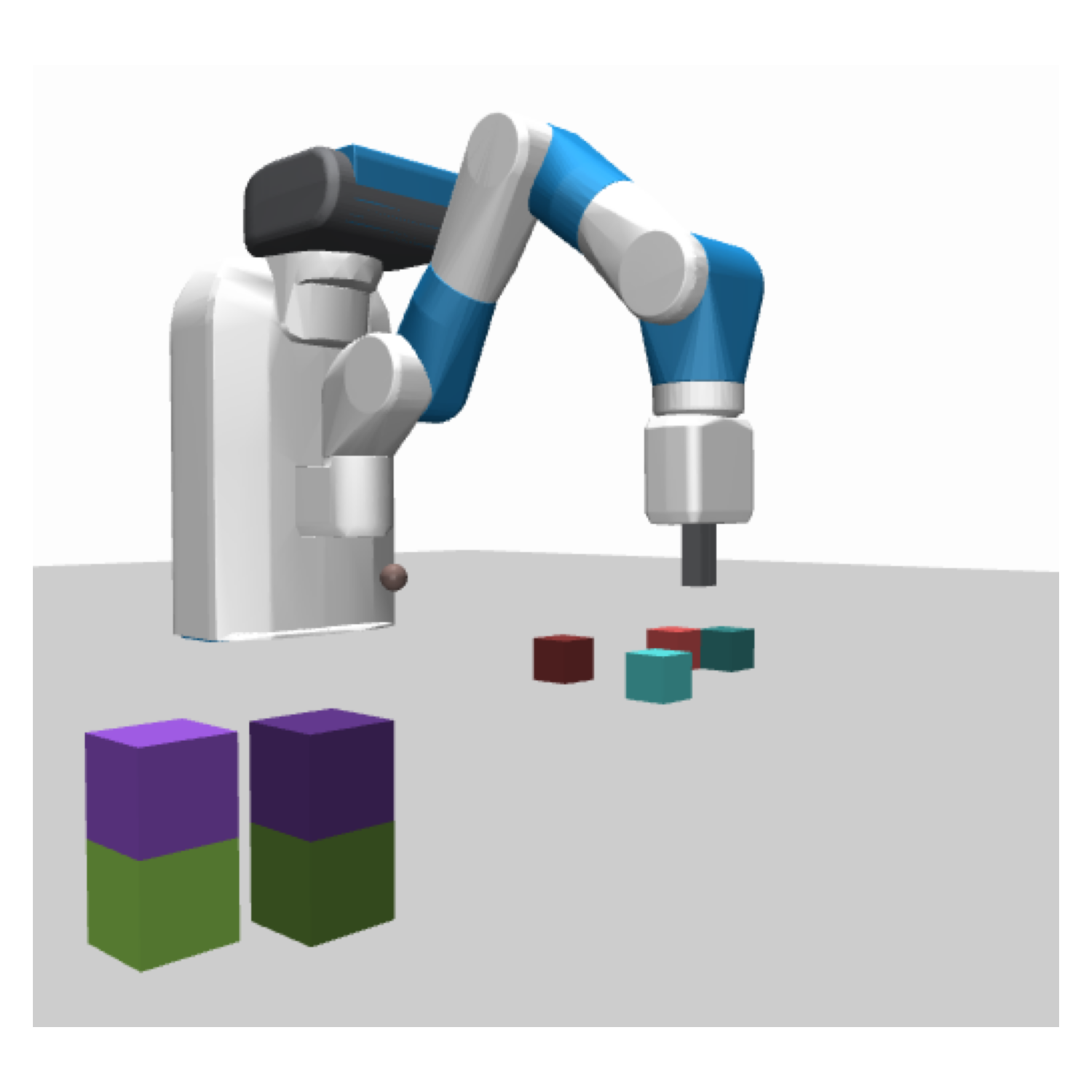}
    \includegraphics[width=.47\linewidth]{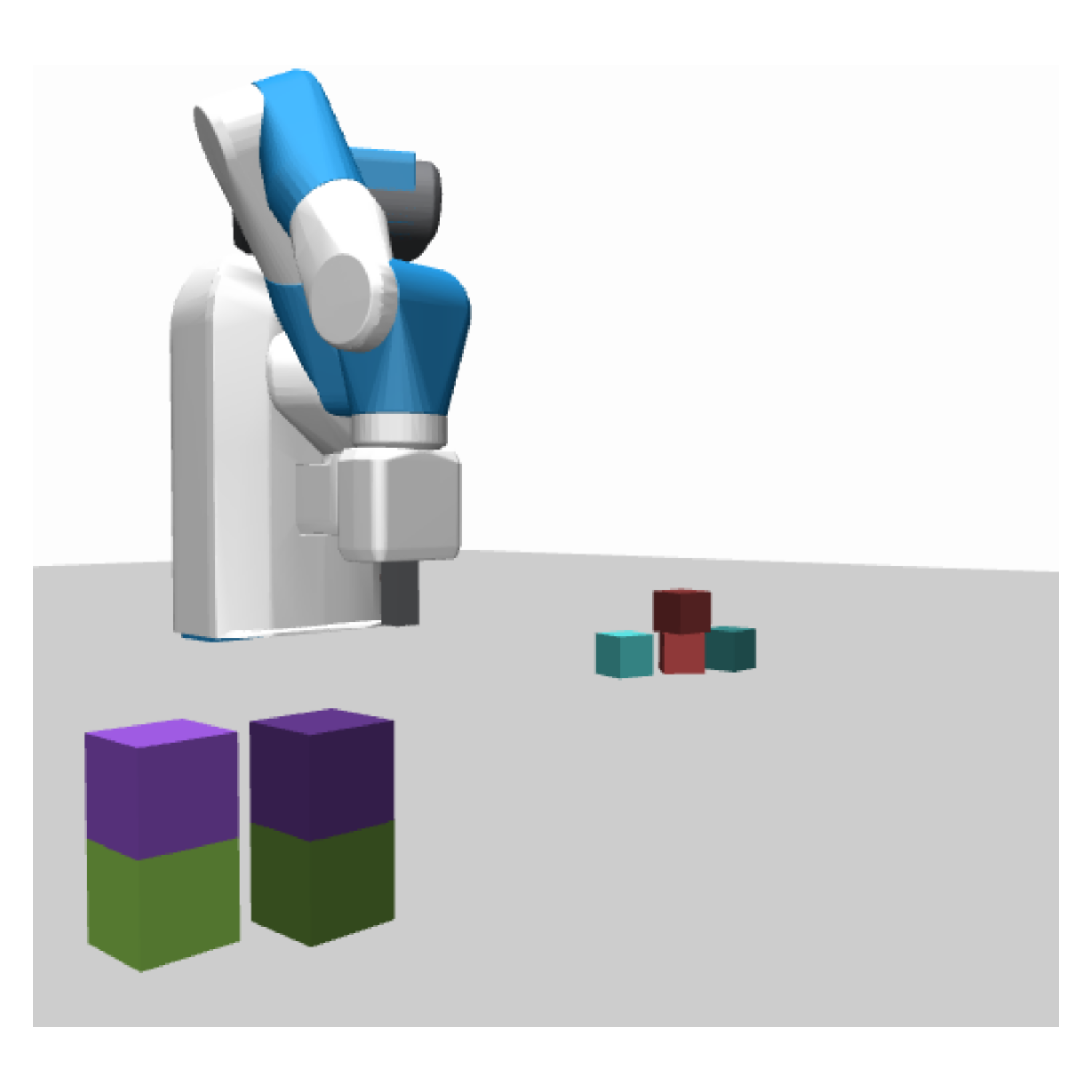}
\vspace{-.6em}
\caption{Multitower 2+2 (partial recreation)}
\label{fig:multitower_partial}
\end{subfigure}
\\
\\
\begin{subfigure}[t]{.48\linewidth}
    \centering
    {\scriptsize
      Start
       \hspace{10em} End
    }
    \includegraphics[width=.47\linewidth]{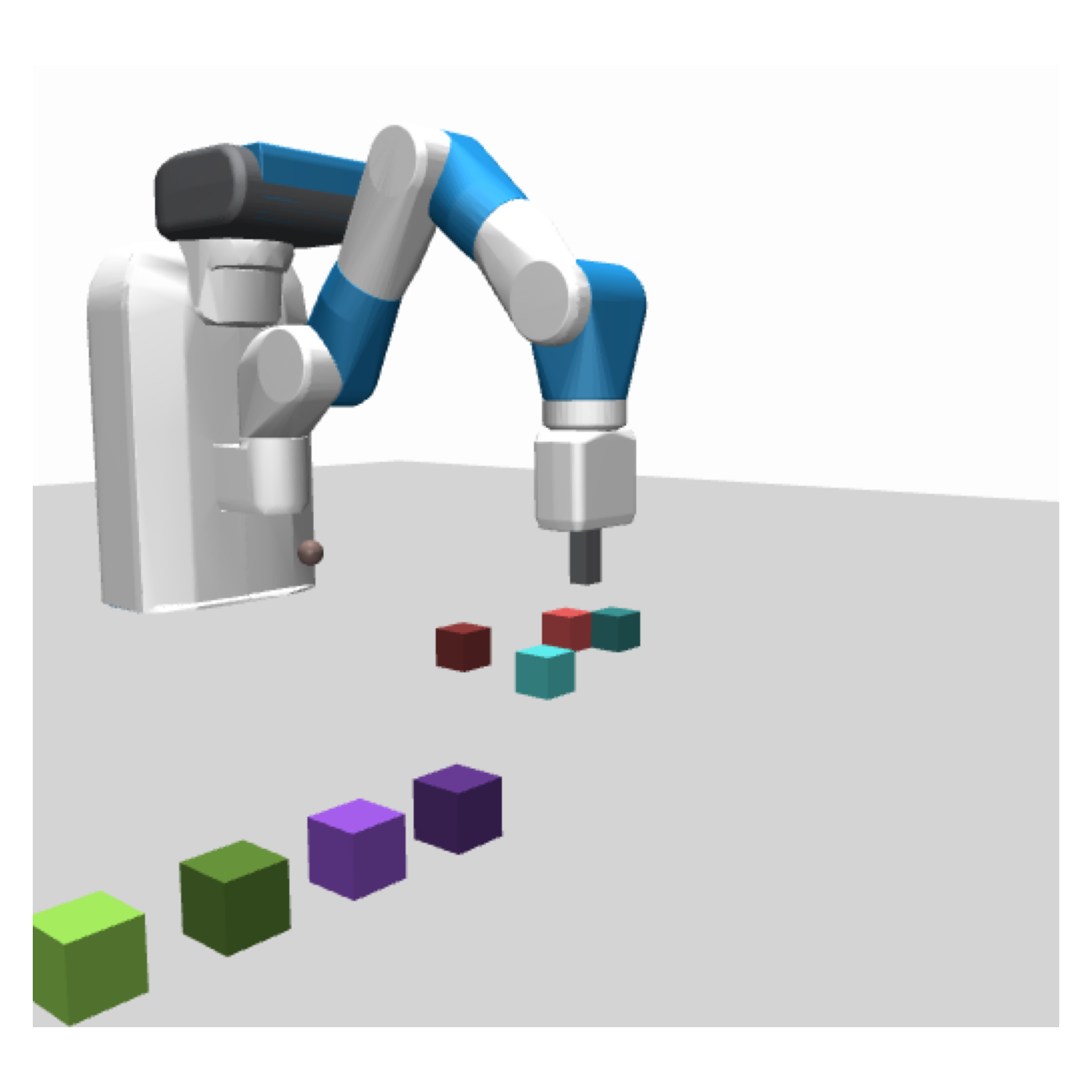}
    \includegraphics[width=.47\linewidth]{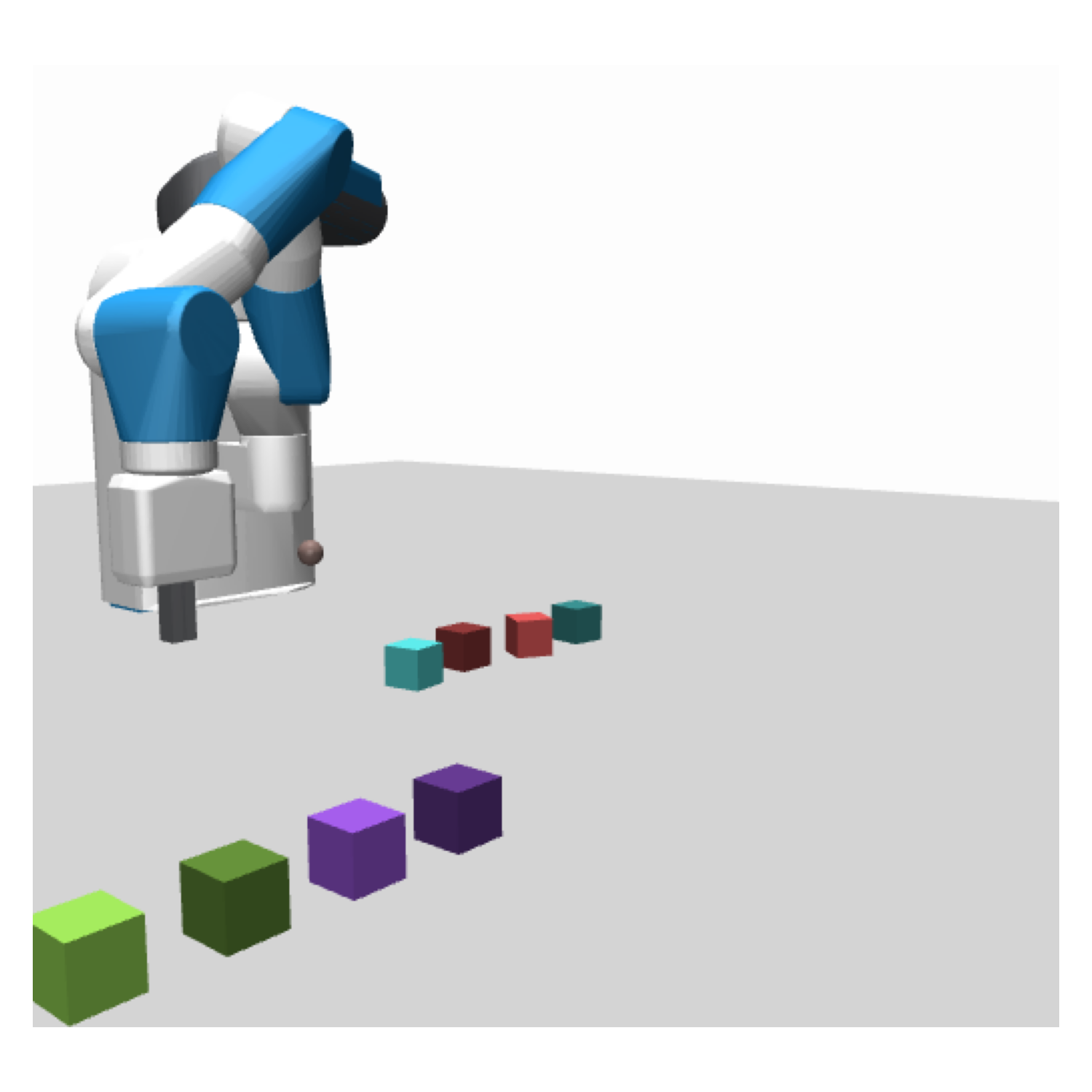}
\vspace{-.6em}
\caption{Spaced line 4}
 \end{subfigure}%
\begin{subfigure}[t]{.48\linewidth}
    \centering
    {\scriptsize
      Start
       \hspace{10em} End
    }
    \includegraphics[width=.47\linewidth]{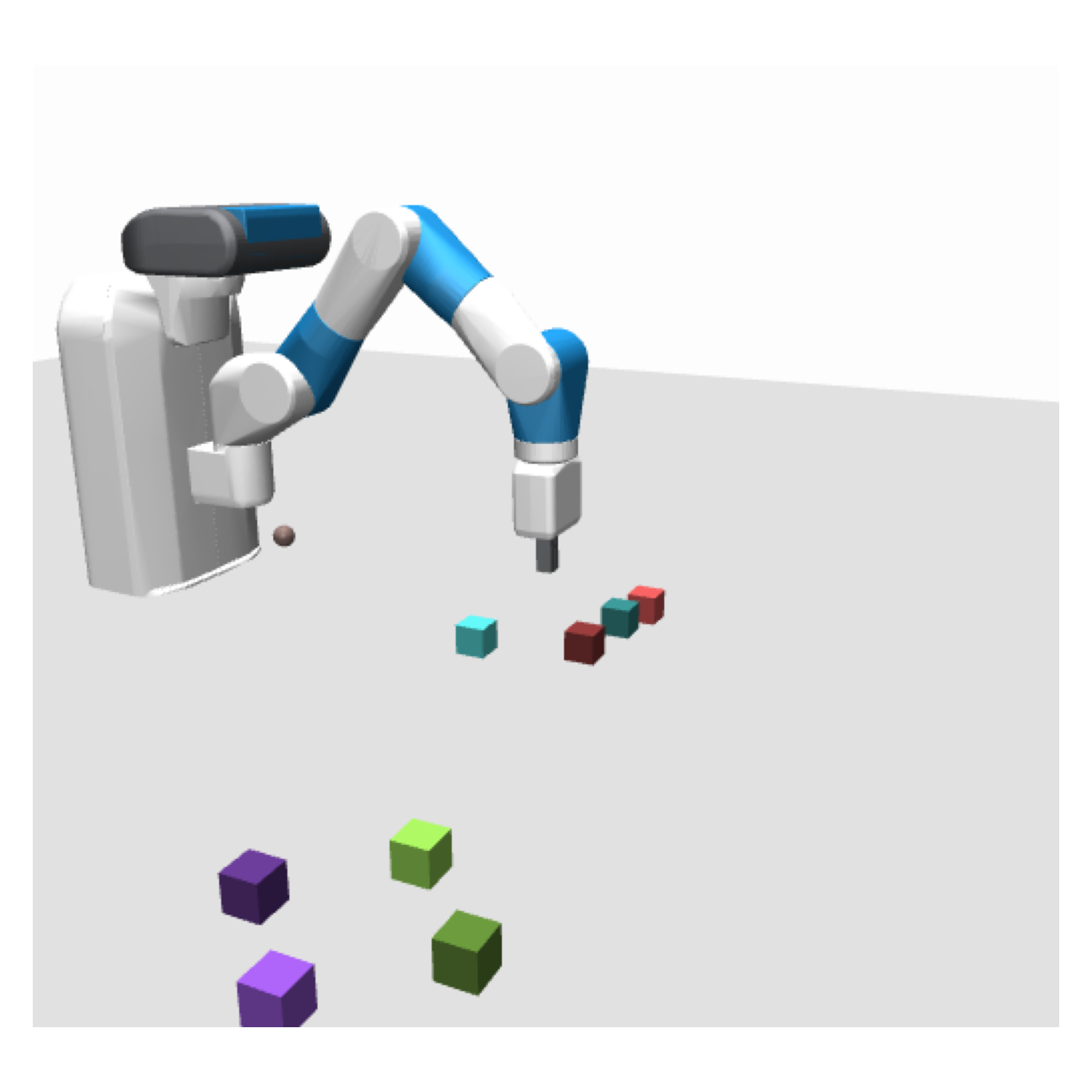}
    \includegraphics[width=.47\linewidth]{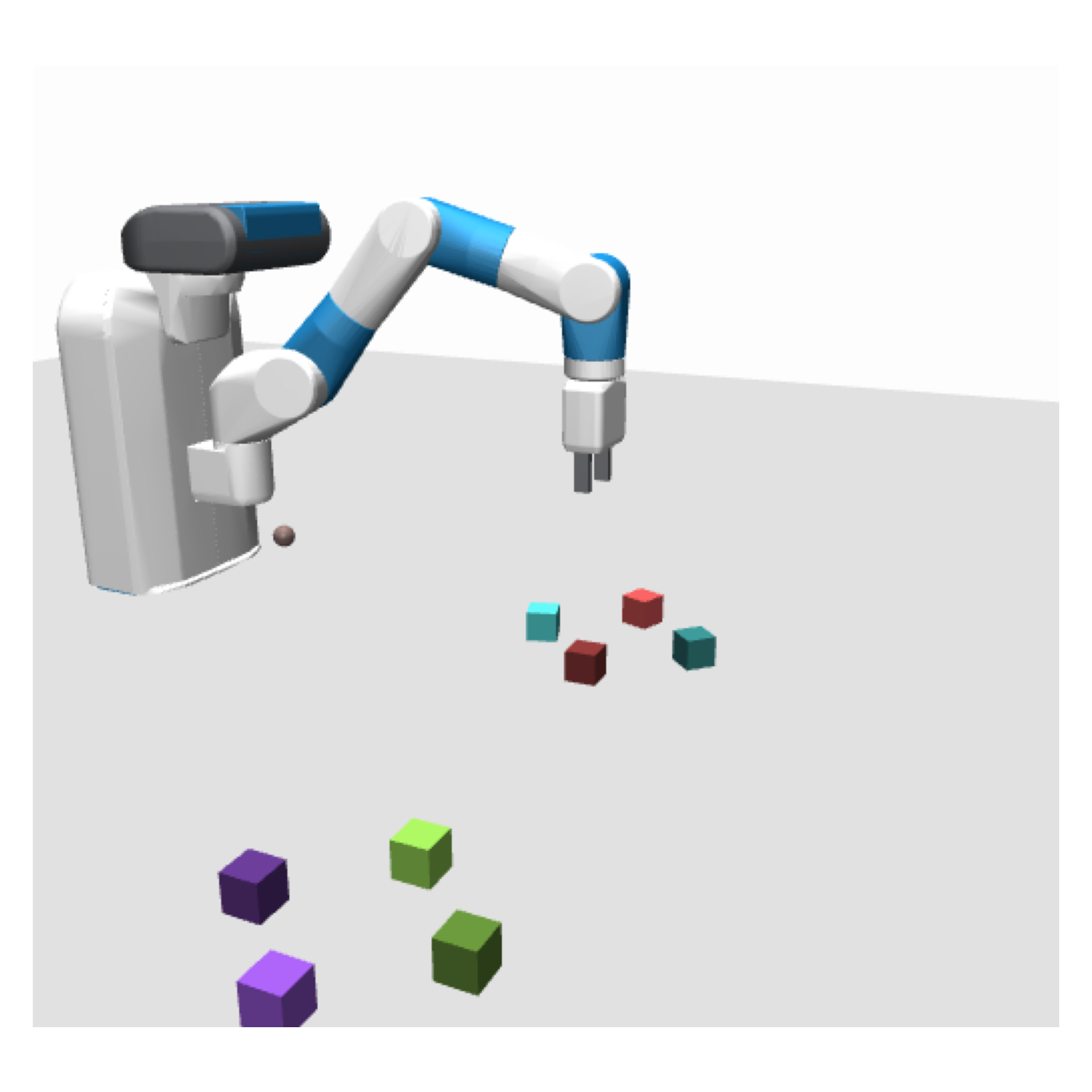}
\vspace{-.6em}
\caption{Rhombus 4}
\end{subfigure}
\addtocounter{figure}{-1}
    \captionof{figure}{Different regular structures initialized outside of the robot's reach at the start of the episode ($t=0$) and re-created by optimizing for \ourMethod with GT models. Showcased here for the end of the episode ($t=200$).}
    \label{fig:recreation_more}
\end{figure}

\section{Experiment Results for \ourMethod in Colored \Grid with GT Models} \label{app:colored_grid}
\begin{figure}[h]
  \centering
    \begin{subfigure}[b]{0.16\textwidth}
         \centering
         \includegraphics[width=\textwidth]{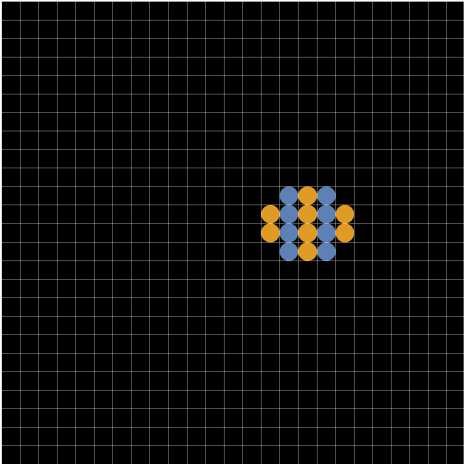}
     \end{subfigure}
    \begin{subfigure}[b]{0.16\textwidth}
         \centering
         \includegraphics[width=\textwidth]{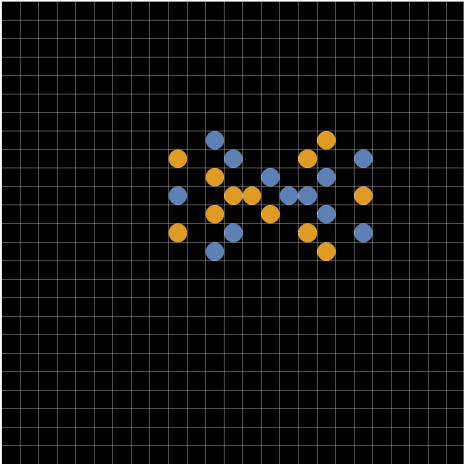}
     \end{subfigure}
     \begin{subfigure}[b]{0.16\textwidth}
        \centering
         \includegraphics[width=\textwidth]{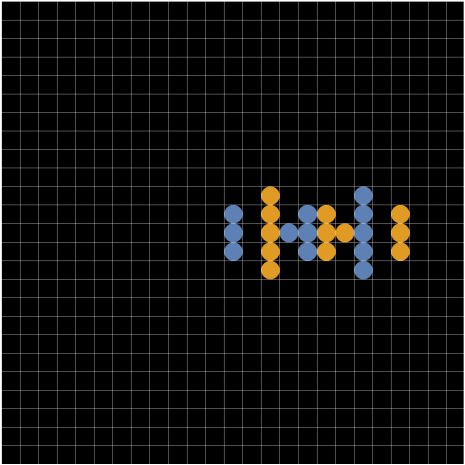}
     \end{subfigure}
      \begin{subfigure}[b]{0.16\textwidth}
         \centering
         \includegraphics[width=\textwidth]{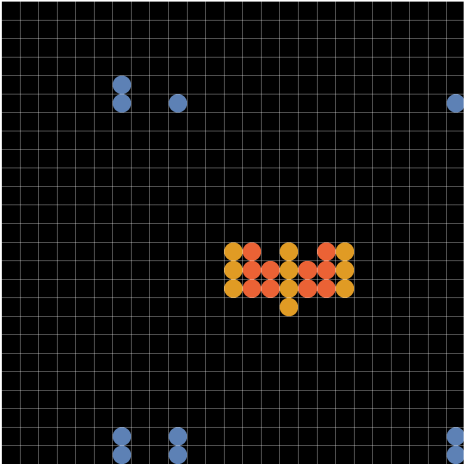}
     \end{subfigure}
 \begin{subfigure}[b]{0.16\textwidth}
         \centering
         \includegraphics[width=\textwidth]{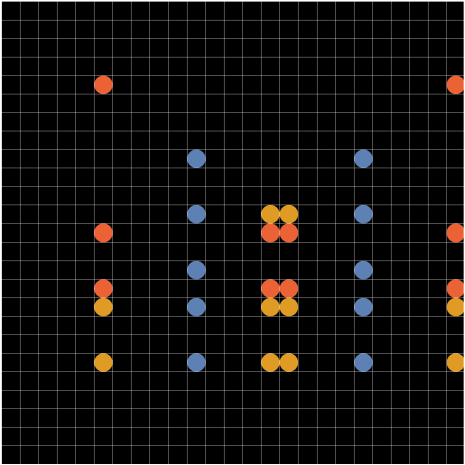}
     \end{subfigure}
  \caption{{\bf{Including color into the regularity computations in \Grid.}} We show generated patterns when maximizing for \ourMethod, with x-y dimensions and color as an extra symbol. In this case, \ourMethod tries to find patterns that are not just spatially symmetric but also regular in color.}
       \label{fig:colored_grid_snapshots}
\end{figure}
In colored \Grid, $S_{\text{obj}}$ includes not only the $x$-$y$ positions, but also a binary variable encoding the different classes of colors. For $c$ colors, we have an encoding of length $\operatorname{ceil}(\log_2c)$. So when we are computing relational \ourMethod, we also take into account the relations between colors. Take as example the case where we have two color classes blue (\textcolor{ourblue}{0}) and orange (\textcolor{ourorange}{1}) and use absolute relational $\phi$: if all blue objects are in a horizontal spaced line, we have repetitions of $|[x_i+\delta_x, 0, \textcolor{ourblue}{0}]-[x_i, 0, \textcolor{ourblue}{0}]|=[\delta_x, 0, 0]$ between neighboring objects with spacing $\delta_x$. If we have a line with an alternating pattern of colors in sequence, then we again have redundancies. For instance, for neighboring entities orange-blue we have: $|[x_i+\delta_x, 0, \textcolor{ourorange}{1}]-[x_i, 0, \textcolor{ourblue}{0}]|=[\delta_x, 0, 1]$. Since we have absolute relational $\phi$, we get the same symbol for the blue-orange neighbors $|[x_i+\delta_x, 0, \textcolor{ourblue}{0}]-[x_i, 0, \textcolor{ourorange}{1}]|=[\delta_x, 0, 1]$. As a result, a blue-orange-blue-orange-{$\ldots$} alternating line is also an optimum for \ourMethod. Additional patterns generated with \ourMethod in colored \Grid with 2 and 3 colors using GT models are shown in \fig{fig:colored_grid_snapshots}. 

\section{Experiment Results for \ourMethod in Custom \FPP} \label{app:custom_fpp}
We test regularity on a variant of \FPP with diverse object shapes and masses (more details in \app{app:envs}).
\subsection{\ourMethod with GT models}
We apply \ourMethod to custom \FPP with different shapes and masses, where we compute regularity between the $x$-$y$ positions of the center of mass of objects. Here our aim is to show the generality of \ourMethod and that we are not constrained to identical entities. The generated constellations when optimizing for \ourMethod using GT models are showcased in \fig{fig:custom_shapes}.

\begin{figure}[!h]
  \centering
      \begin{subfigure}[b]{0.16\textwidth}
         \centering
         \includegraphics[width=\textwidth]{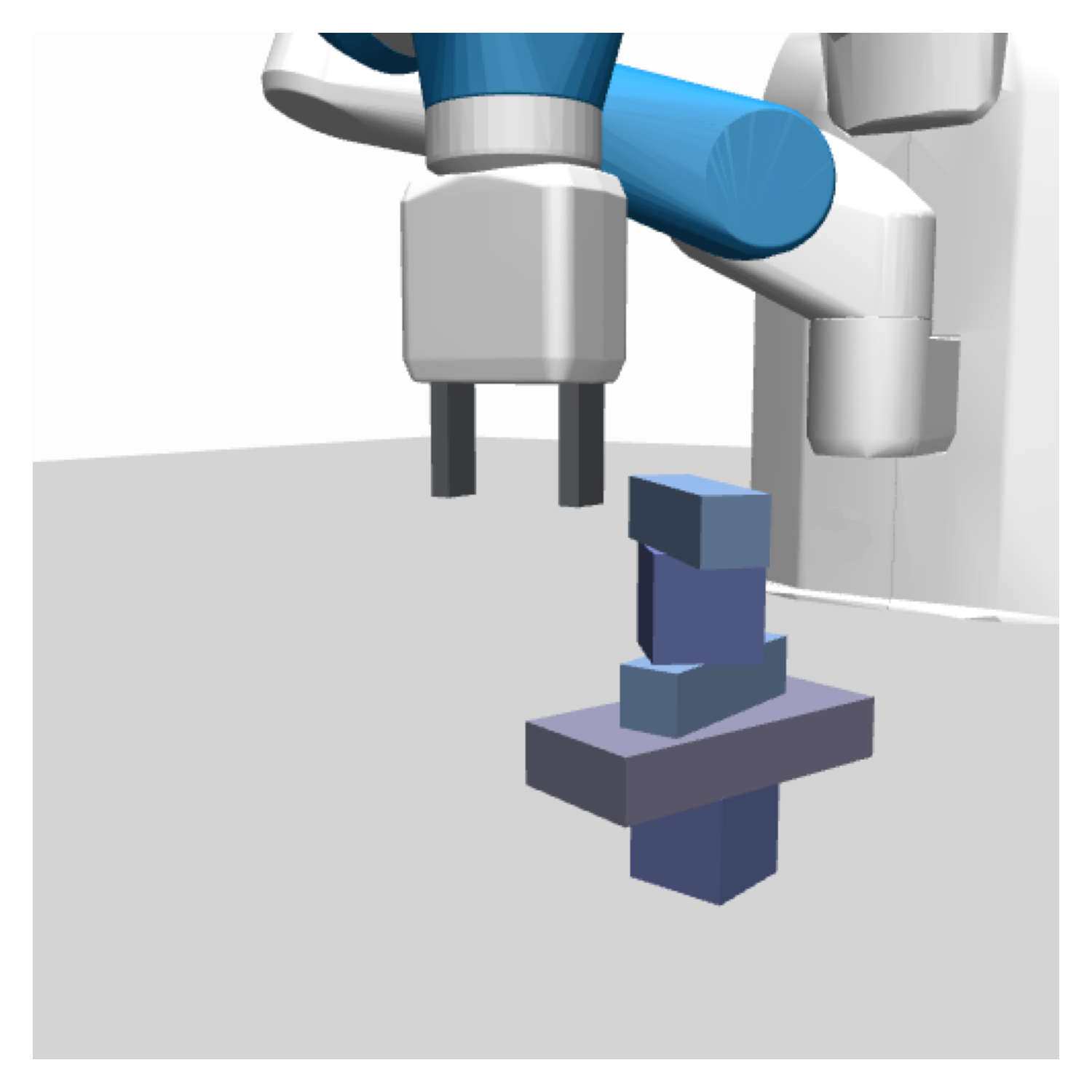}
     \end{subfigure}
      \begin{subfigure}[b]{0.16\textwidth}
         \centering
         \includegraphics[width=\textwidth]{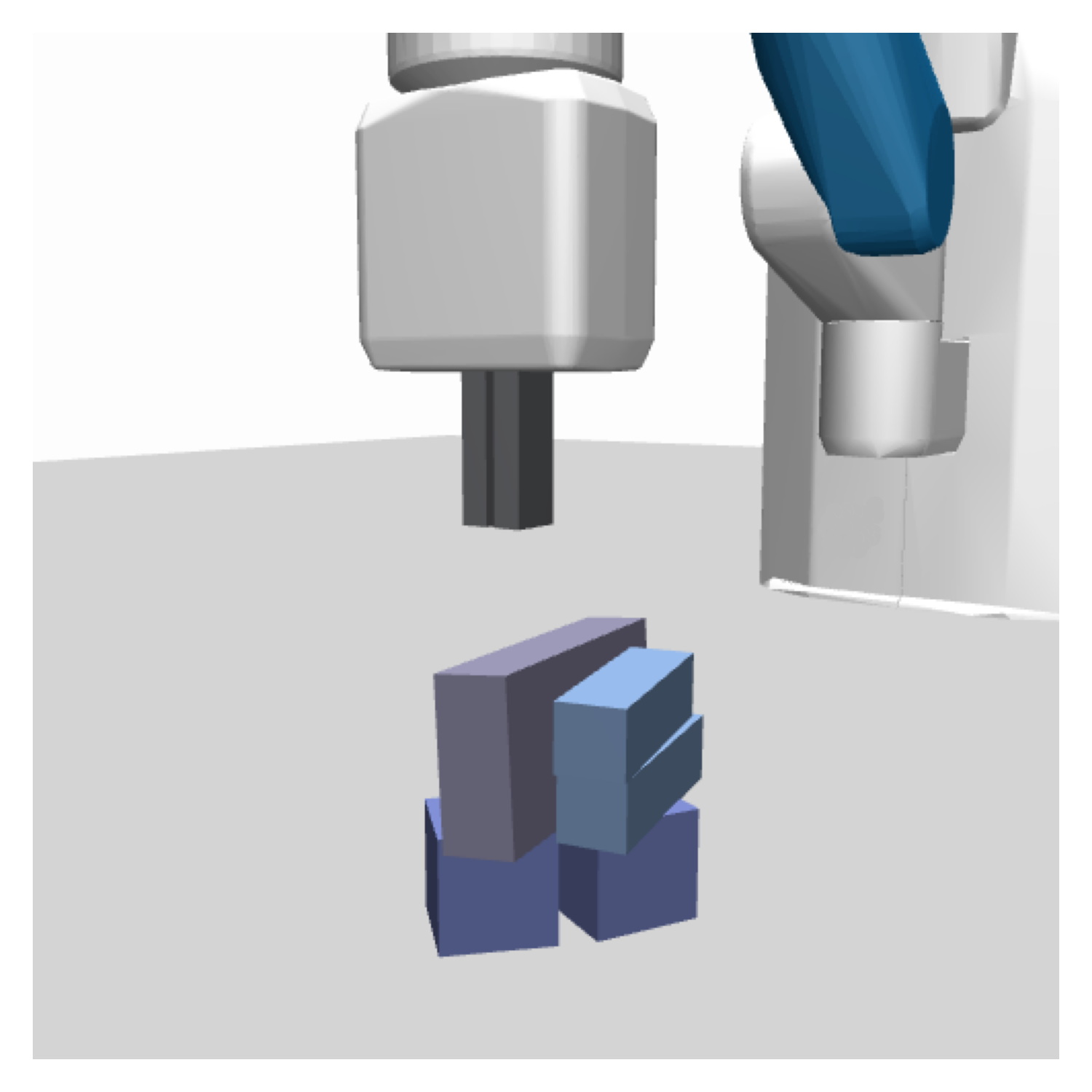}
     \end{subfigure}
    \begin{subfigure}[b]{0.16\textwidth}
         \centering
         \includegraphics[width=\textwidth]{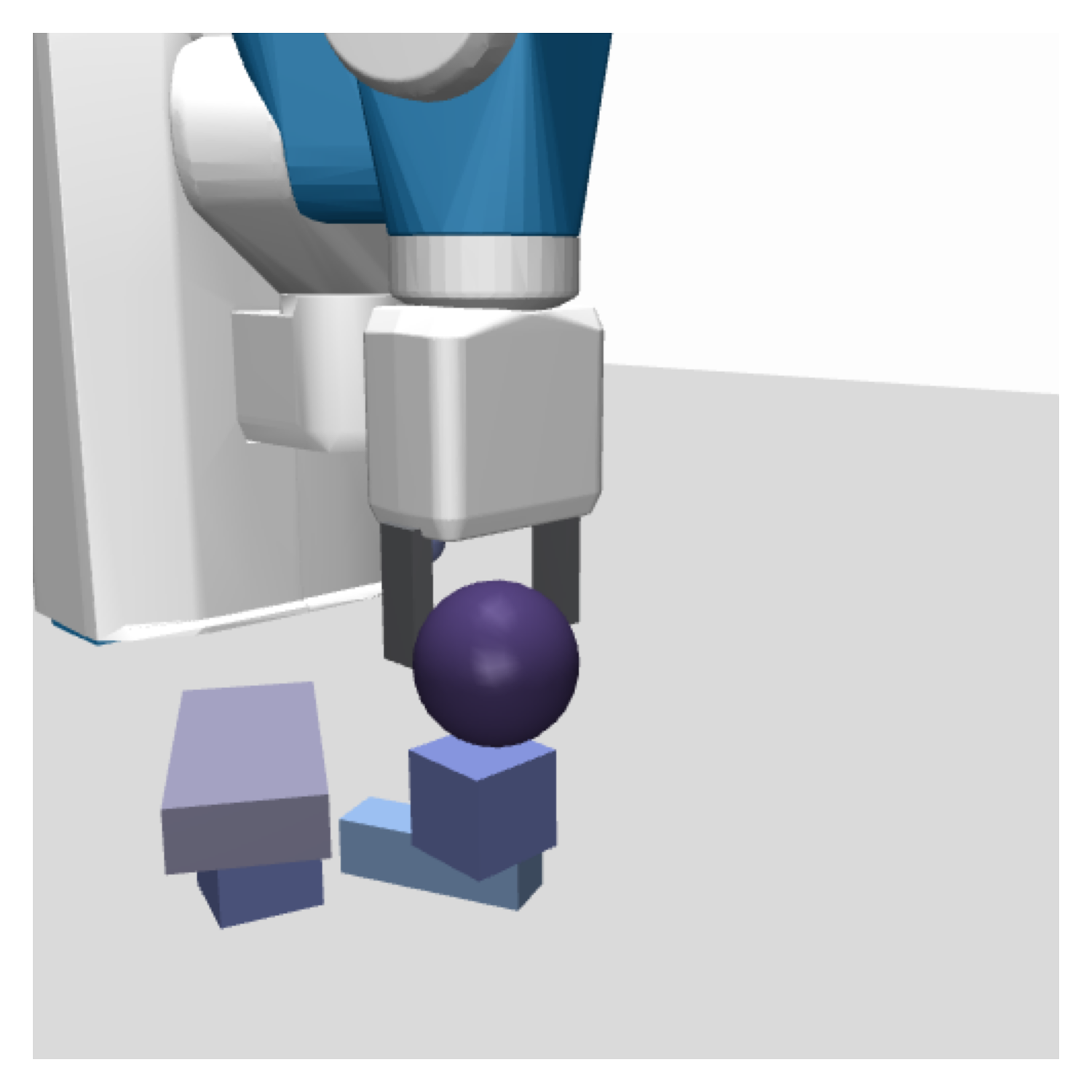}
     \end{subfigure}
    \begin{subfigure}[b]{0.16\textwidth}
         \centering
         \includegraphics[width=\textwidth]{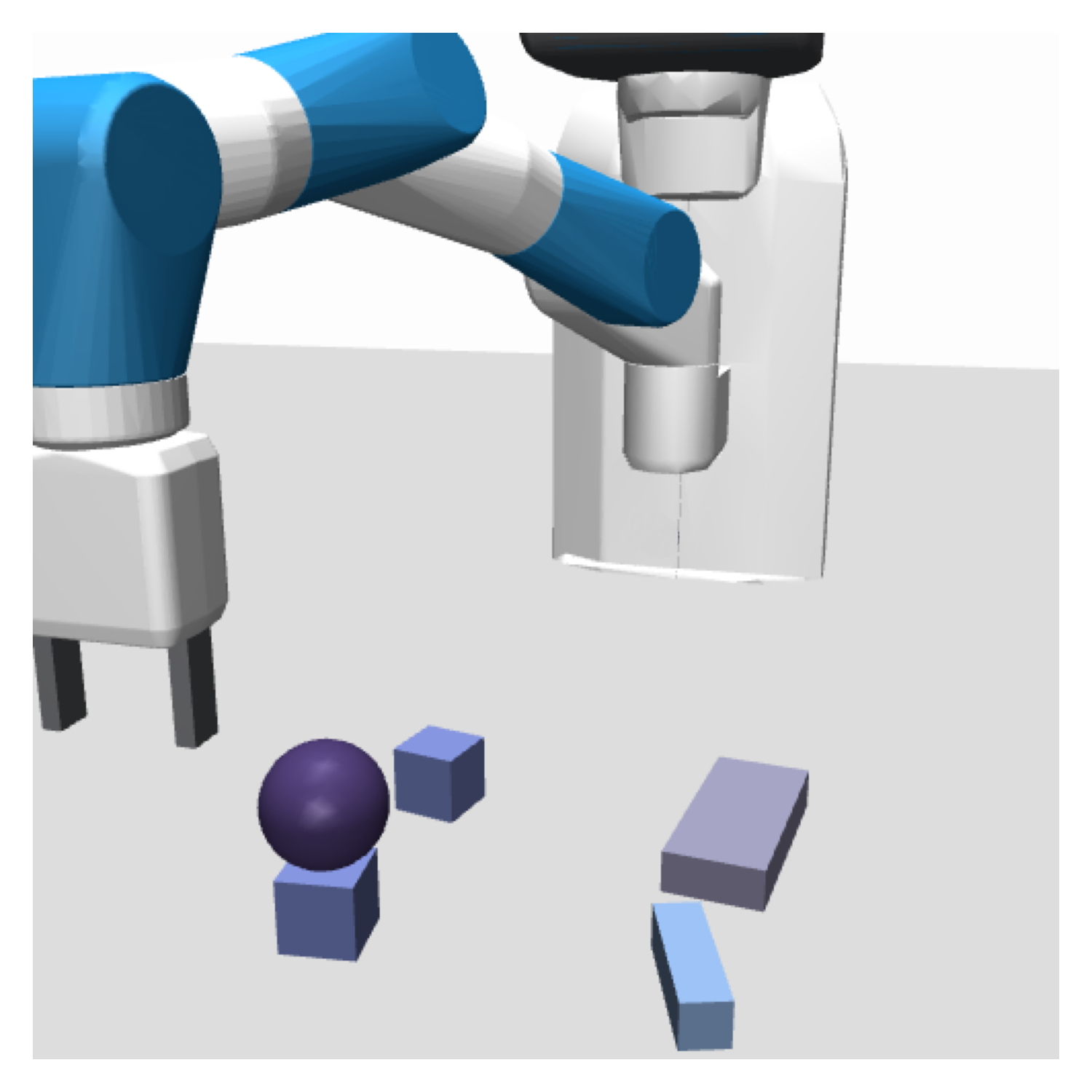}
     \end{subfigure}
     \begin{subfigure}[b]{0.16\textwidth}
        \centering
         \includegraphics[width=\textwidth]{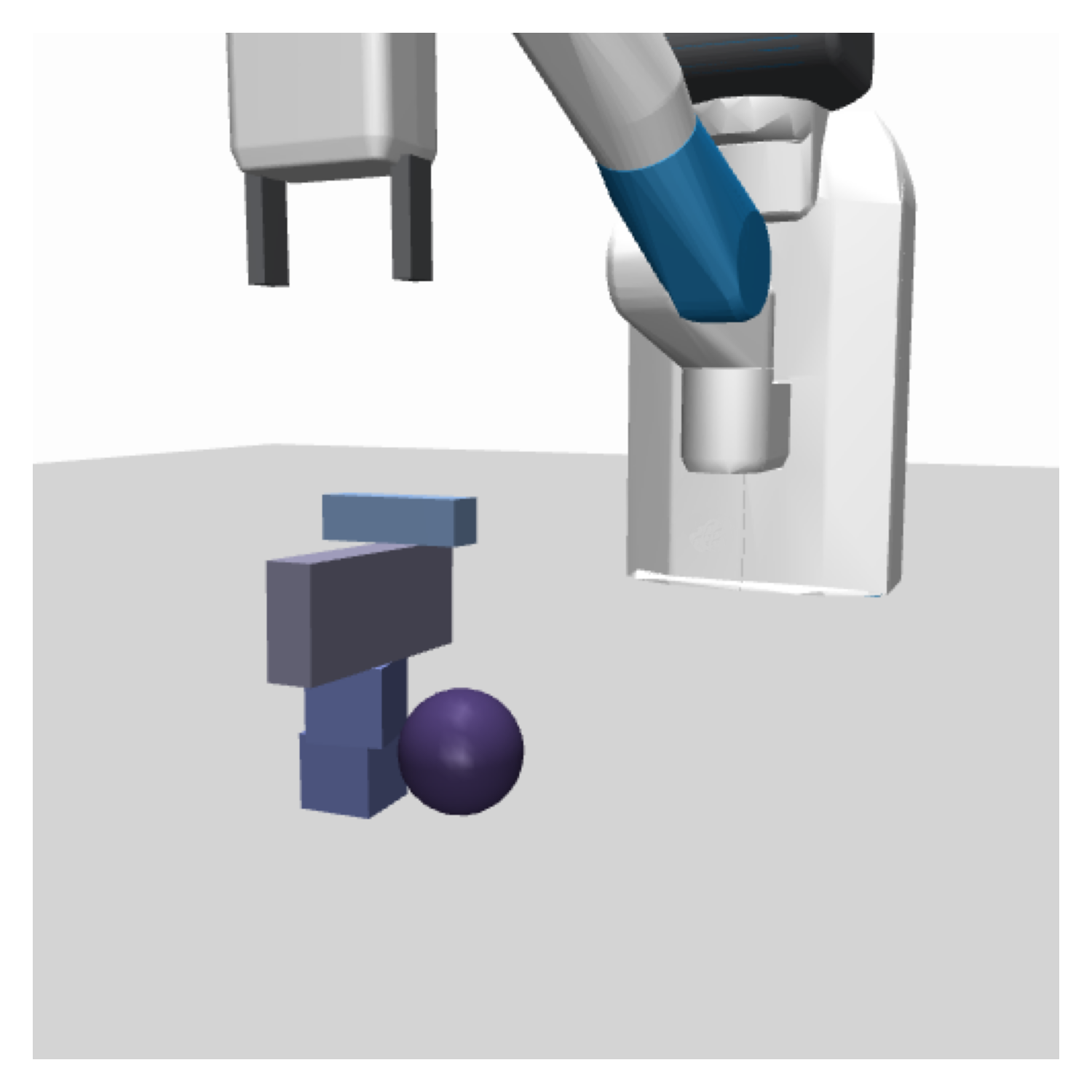}
     \end{subfigure}
     \begin{subfigure}[b]{0.16\textwidth}
         \centering
         \includegraphics[width=\textwidth]{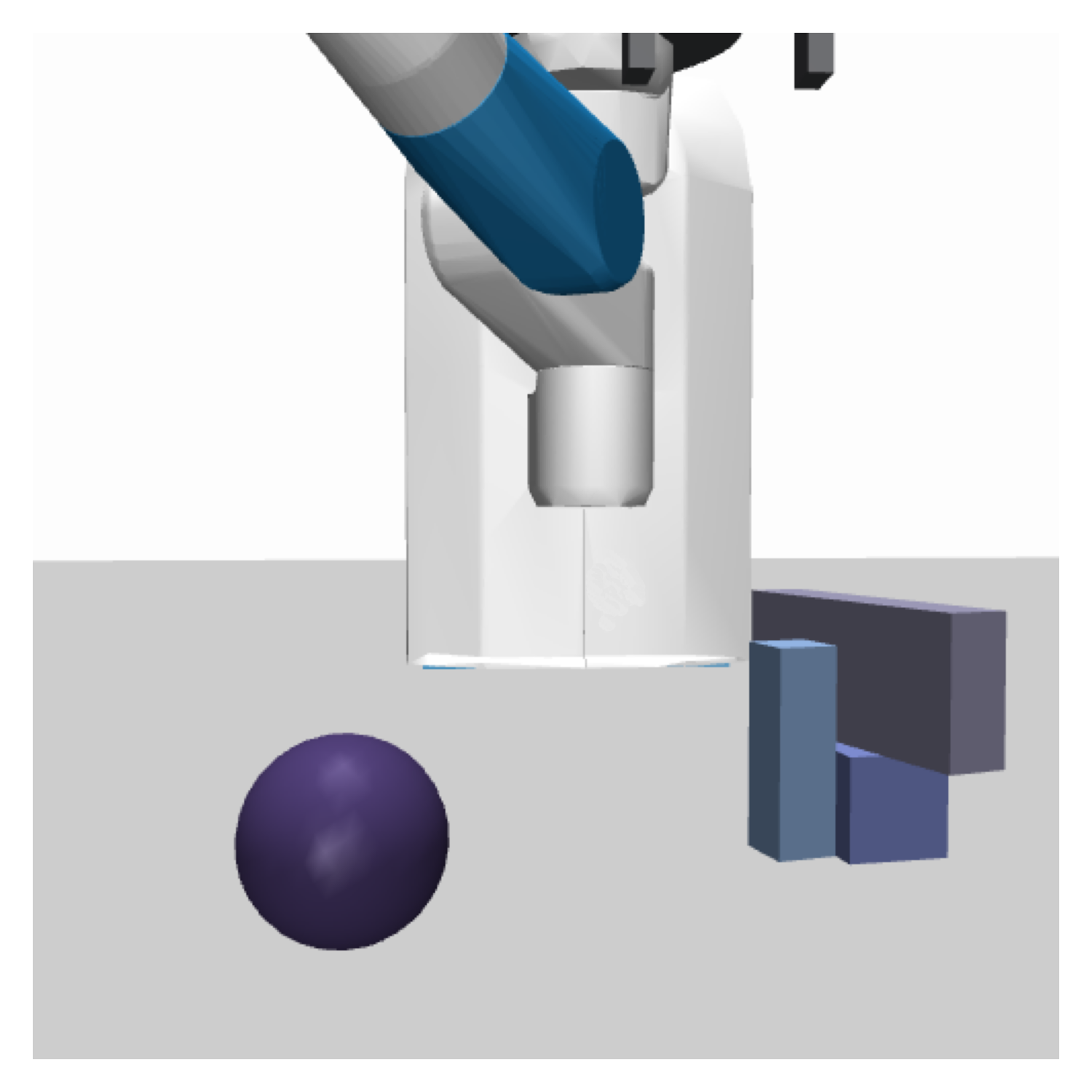}
     \end{subfigure}
  \caption{{\bf{Applying RaIR to a custom \FPP environment with different shapes and masses with GT models.}} We show generated arrangements when maximizing for \ourMethod with x-y positions of objects (center of mass) in 2 instances of the environment: 1) with 2 cubes, 2 short columns and 1 flat block (the left 2 images) and 2) with 2 cubes, 1 ball, 1 column and 1 flat block (the right 4 images).}
       \label{fig:custom_shapes}
\end{figure}

\subsection{\ourMethod in Free Play with Learned Models} \label{app:custom_fpp_perf}

We run free play in the custom \FPP. During free play, we initialize one instance from each object type such that we have 1 cube, 1 ball, 1 column and 1 flat block. We then evaluate the zero-shot downstream task performance of \ourMethodA and \oldMethod for 3 different stacking tasks (\fig{fig:custom_fpp_goals}). The results are shown in \tab{tab:custom_fpp_success}, where we obtain higher success rates with \ourMethodA.

The free play parameters (model and controller settings) are identical to the standard \FPP environment given in \tab{tab:gnn_model_training_settings}. The only difference is, we include a categorical variable to encode the different object types and append this to the observation vector for all objects. This is done in the same fashion as the static observations proposed in \citet{Sancaktaretal22}. For \ourMethodA, we again use $\lambda=0.1$.

\begin{figure}[ht]
    \begin{minipage}[c]{.43\linewidth}
  \centering
     \begin{subfigure}[t]{0.32\textwidth}
        \centering
         \includegraphics[width=\textwidth]{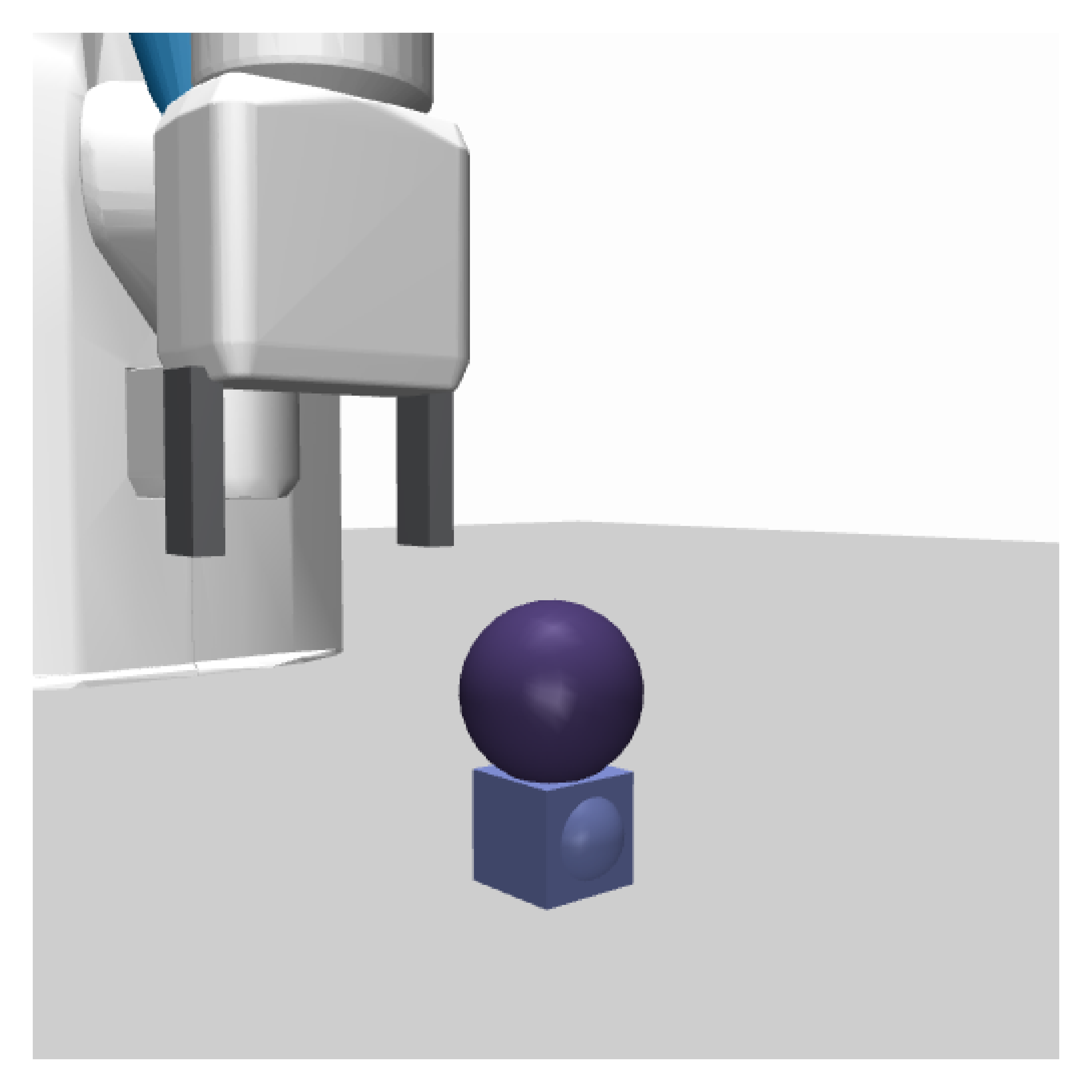}
         \centering{\scriptsize Stack Cube + Ball}
     \end{subfigure}
     \begin{subfigure}[t]{0.32\textwidth}
         \centering
         \includegraphics[width=\textwidth]{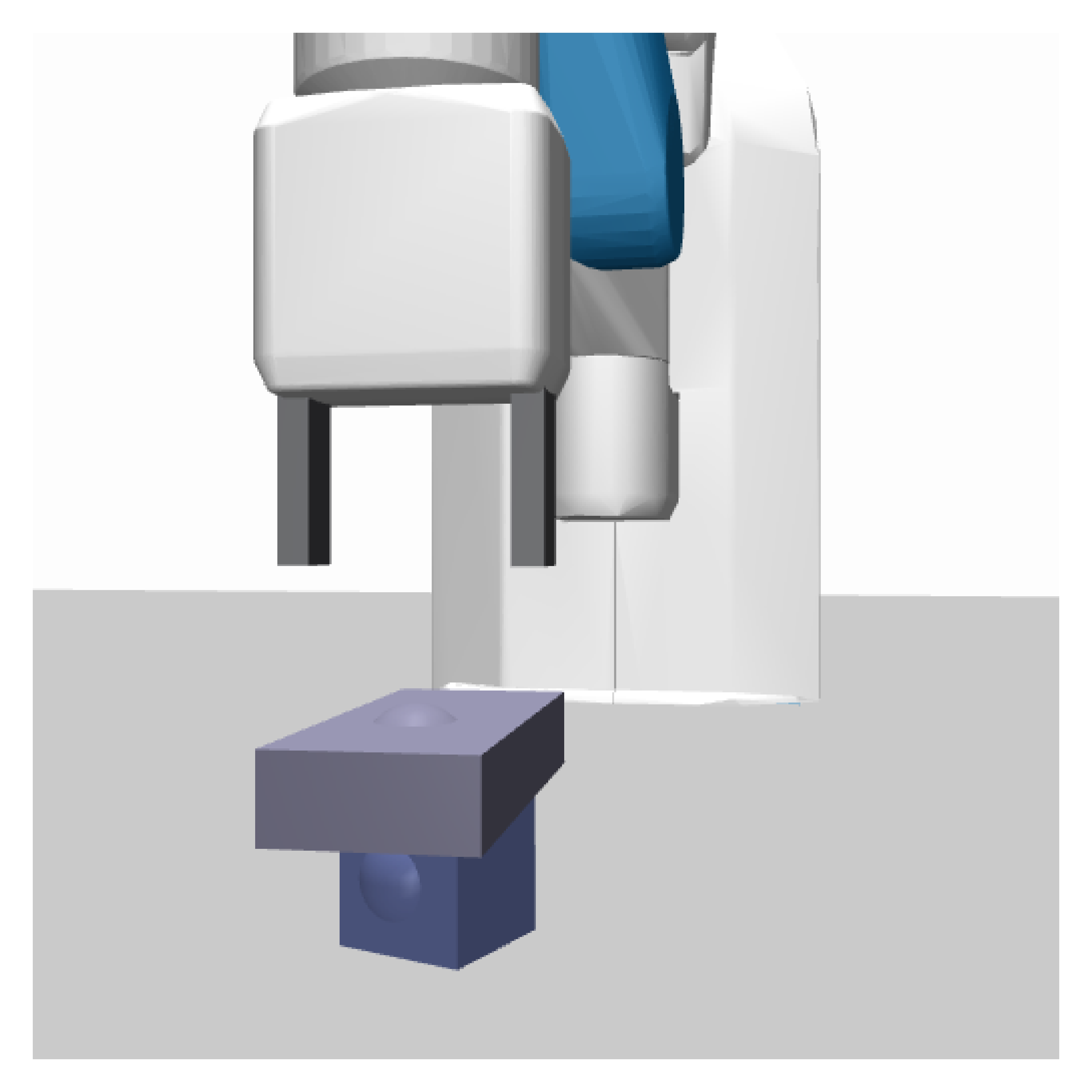}
         \centering{\scriptsize Stack Cube + Flat Block}
     \end{subfigure}
      \begin{subfigure}[t]{0.32\textwidth}
         \centering
         \includegraphics[width=\textwidth]{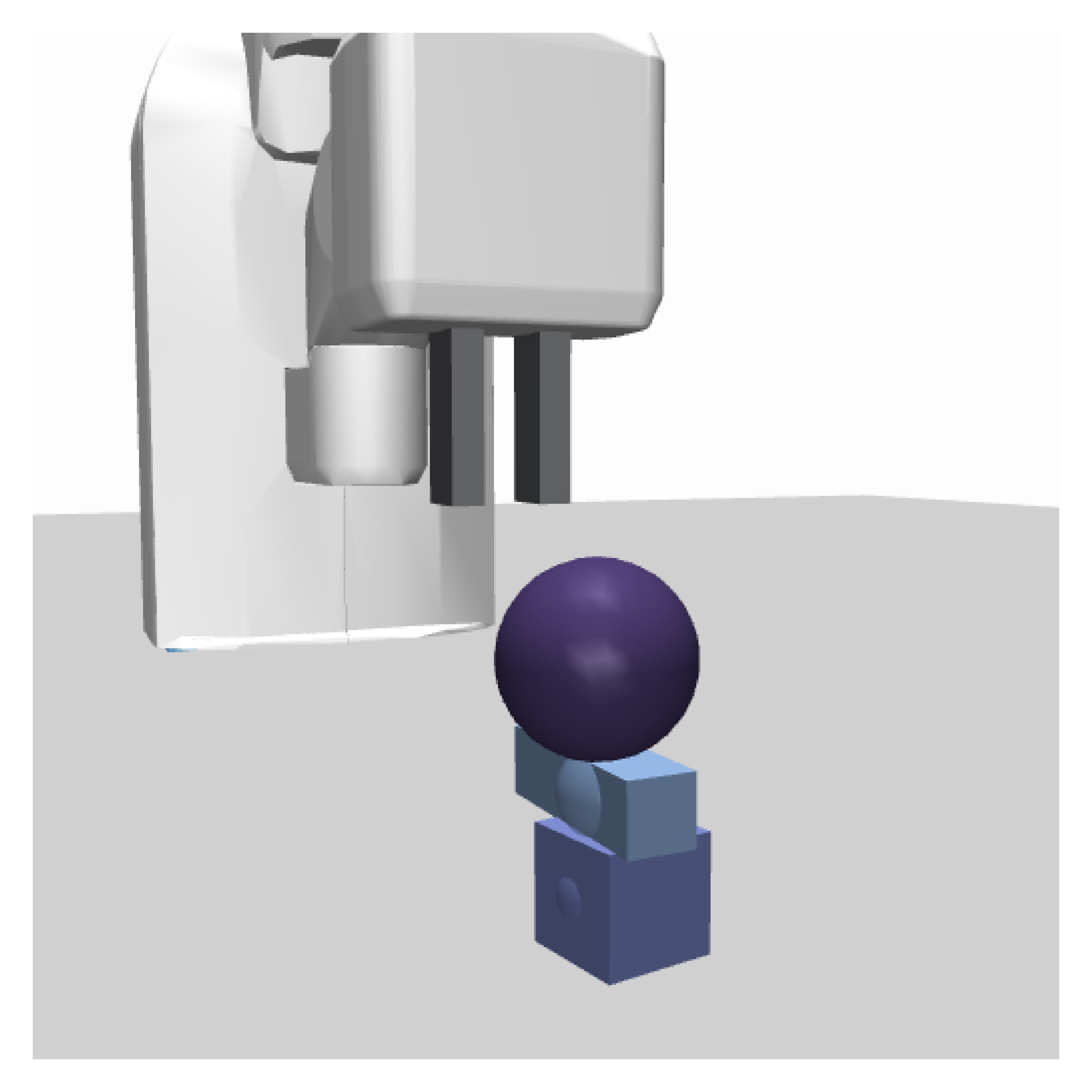}
         \centering{\scriptsize Stack Cube + Column + Ball}
     \end{subfigure}
          \\
     \vspace{.2em}
    \addtocounter{figure}{-1}
  \captionof{figure}{{\bf{Downstream tasks for custom \FPP.}} %
  }
       \label{fig:custom_fpp_goals}
    \end{minipage}
    \hfill
    \begin{minipage}[c]{.535\linewidth}
    \centering
  \captionof{table}{{\bf Zero-shot downstream task generalization performance of \ourMethodA for custom \FPP.} Results are shown for five independent seeds, for models evaluated after 300 free play iterations (equivalent to 2K $\times 300 = 600$K transitions).}
  \centering
    \renewcommand{\arraystretch}{.99}
    \resizebox{\linewidth}{!}{ %
    \begin{tabular}{@{}lccc@{}}
    \toprule
    & Stack Cube + & Stack Cube +& Stack Cube +  \\
    &  Ball & Flat Block  &  Column + Ball \\
    \midrule
    \ourMethodA & $\mathbf{0.66 \pm 0.10}$  & $\mathbf{0.67 \pm 0.09}$ & $\mathbf{0.15 \pm 0.05}$  \\
     \oldMethod & $0.38 \pm 0.11$ & $0.45 \pm 0.04$  & $0.09 \pm 0.03$ \\
  \bottomrule
    \end{tabular}
  }
    \label{tab:custom_fpp_success}
    \end{minipage}
\end{figure}

\section{Experiment Results for \ourMethod in the Quadruped and Walker Environments} \label{app:roboyoga}
We showcase that regularity also finds application in locomotion environments, where we compute \ourMethod on the world coordinates of the robot joints. Environment details for both Quadruped and Walker can be found in \app{app:envs}.

\begin{wrapfigure}[12]{r}{0.25\textwidth}
    \centering
    \includegraphics[width=\linewidth]{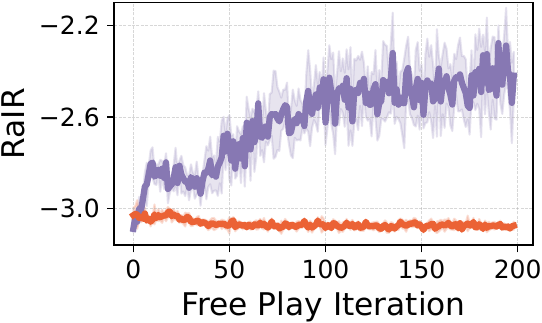}\vspace{-.6em}
    \caption{{\bf Highest \ourMethod in Quadruped free play} with {\color{ourviolet}\ourMethodA} and {\color{ourred}\oldMethod}\looseness-1.}
    \label{fig:rair_best_quadruped}
\end{wrapfigure}
\begin{figure}[ht]
  \centering
     \begin{subfigure}[b]{0.137\textwidth}
        \centering
         \includegraphics[width=\textwidth]{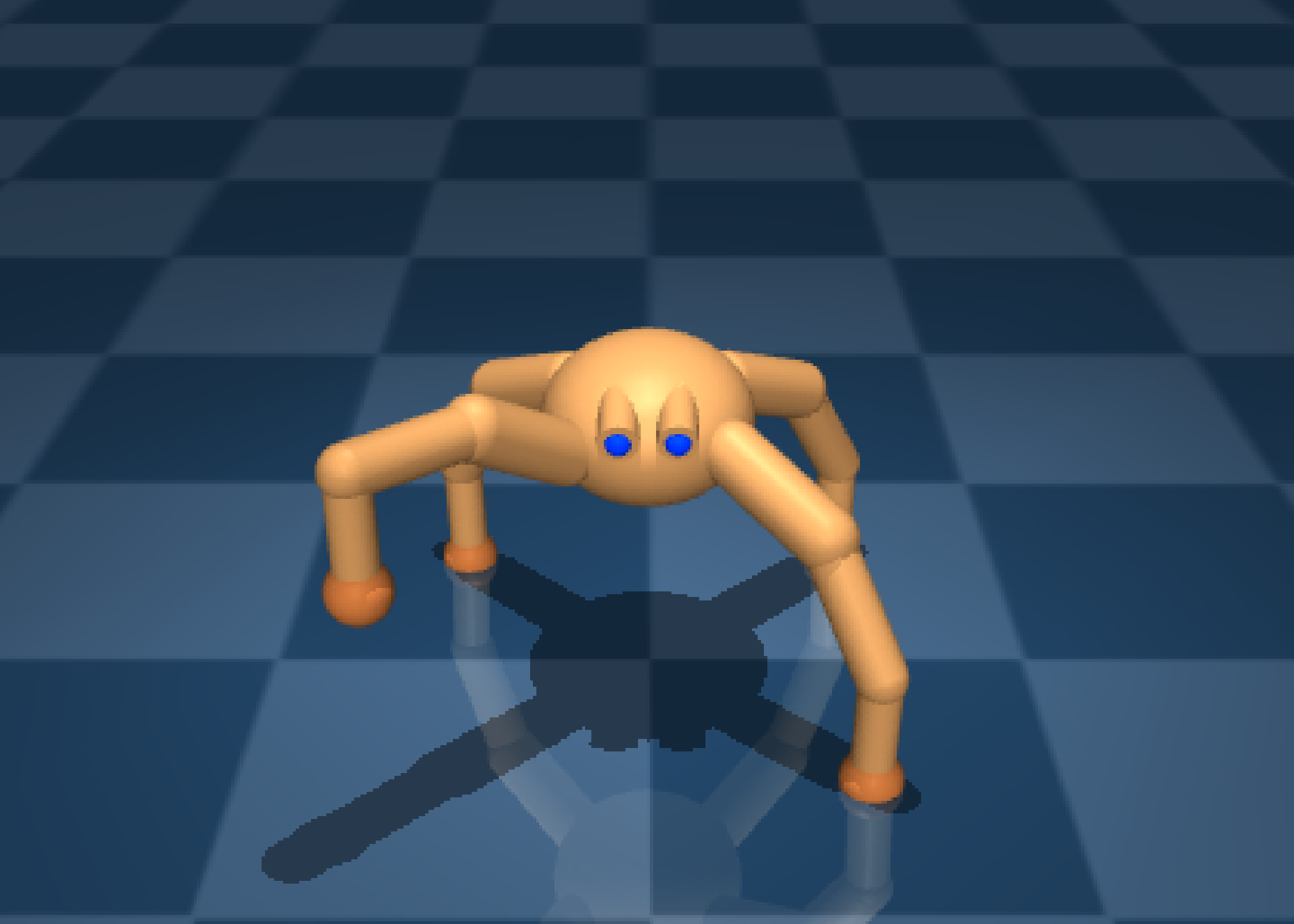}\vspace{-.3em}
         \centering{\scriptsize Stand Leg Up}
     \end{subfigure}
     \begin{subfigure}[b]{0.137\textwidth}
         \centering
         \includegraphics[width=\textwidth]{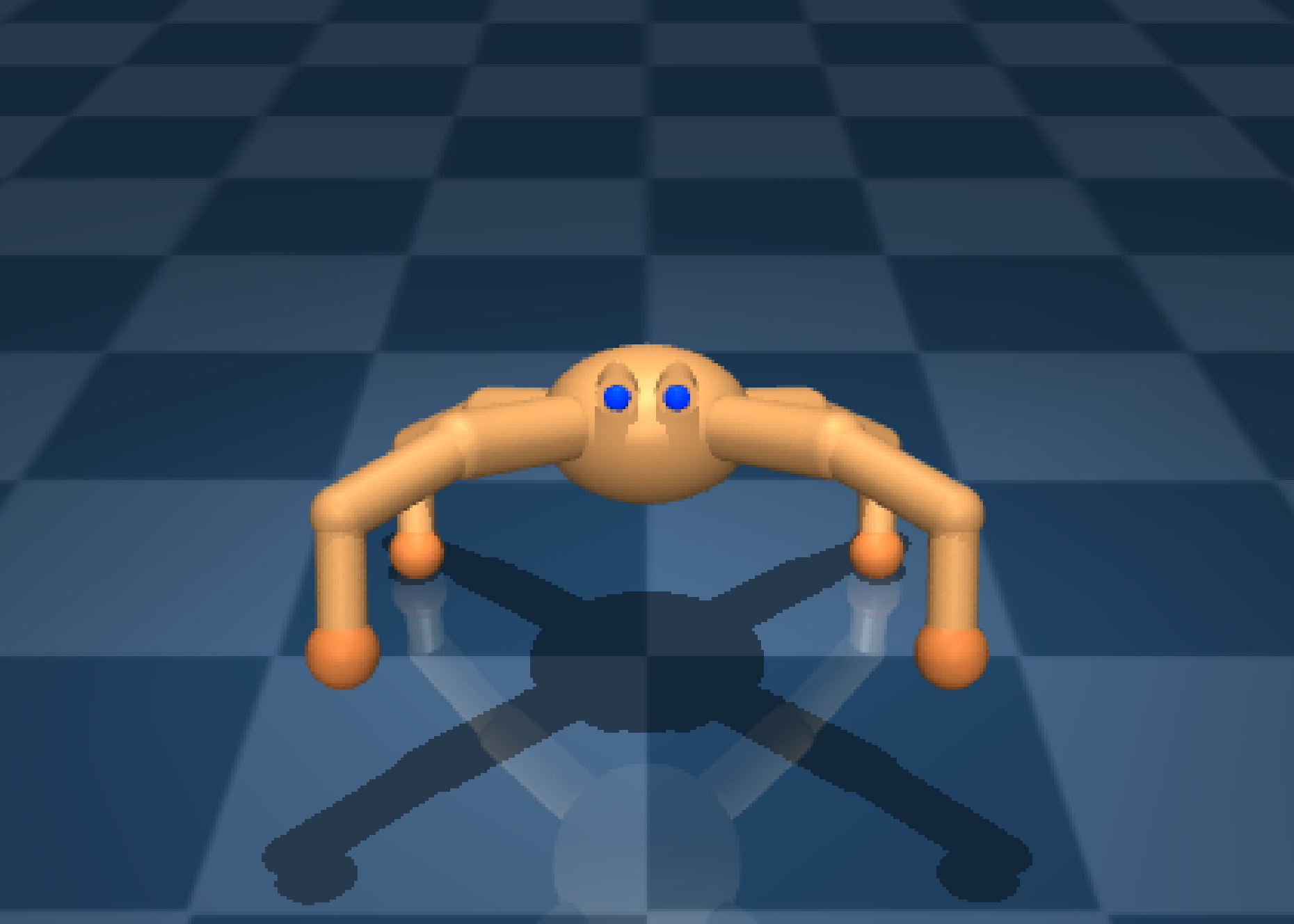}\vspace{-.3em}
         \centering{\scriptsize Attack}
     \end{subfigure}
      \begin{subfigure}[b]{0.137\textwidth}
         \centering
         \includegraphics[width=\textwidth]{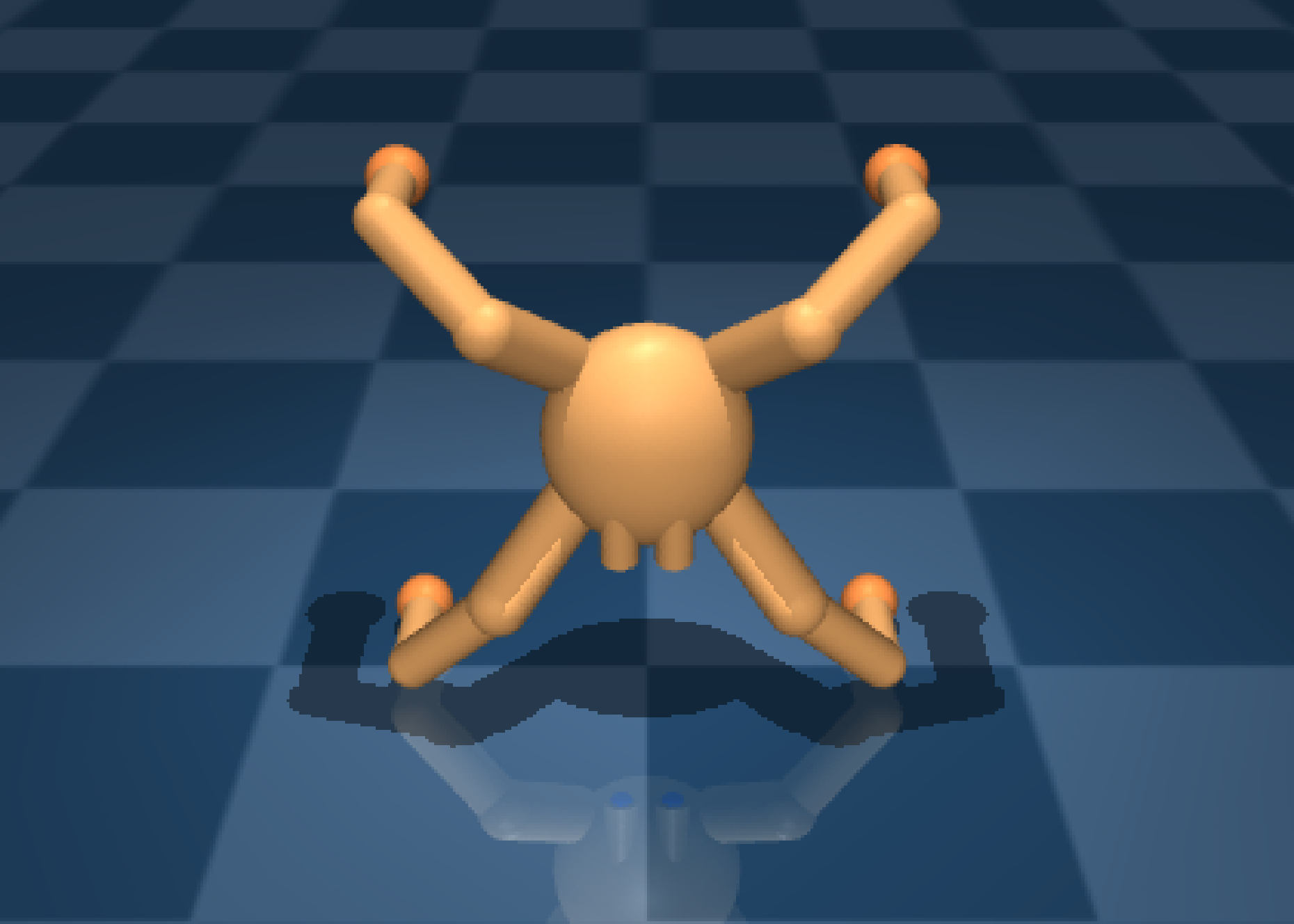}\vspace{-.3em}
         \centering{\scriptsize Balance Front}
     \end{subfigure}
     \begin{subfigure}[b]{0.137\textwidth}
         \centering
         \includegraphics[width=\textwidth]{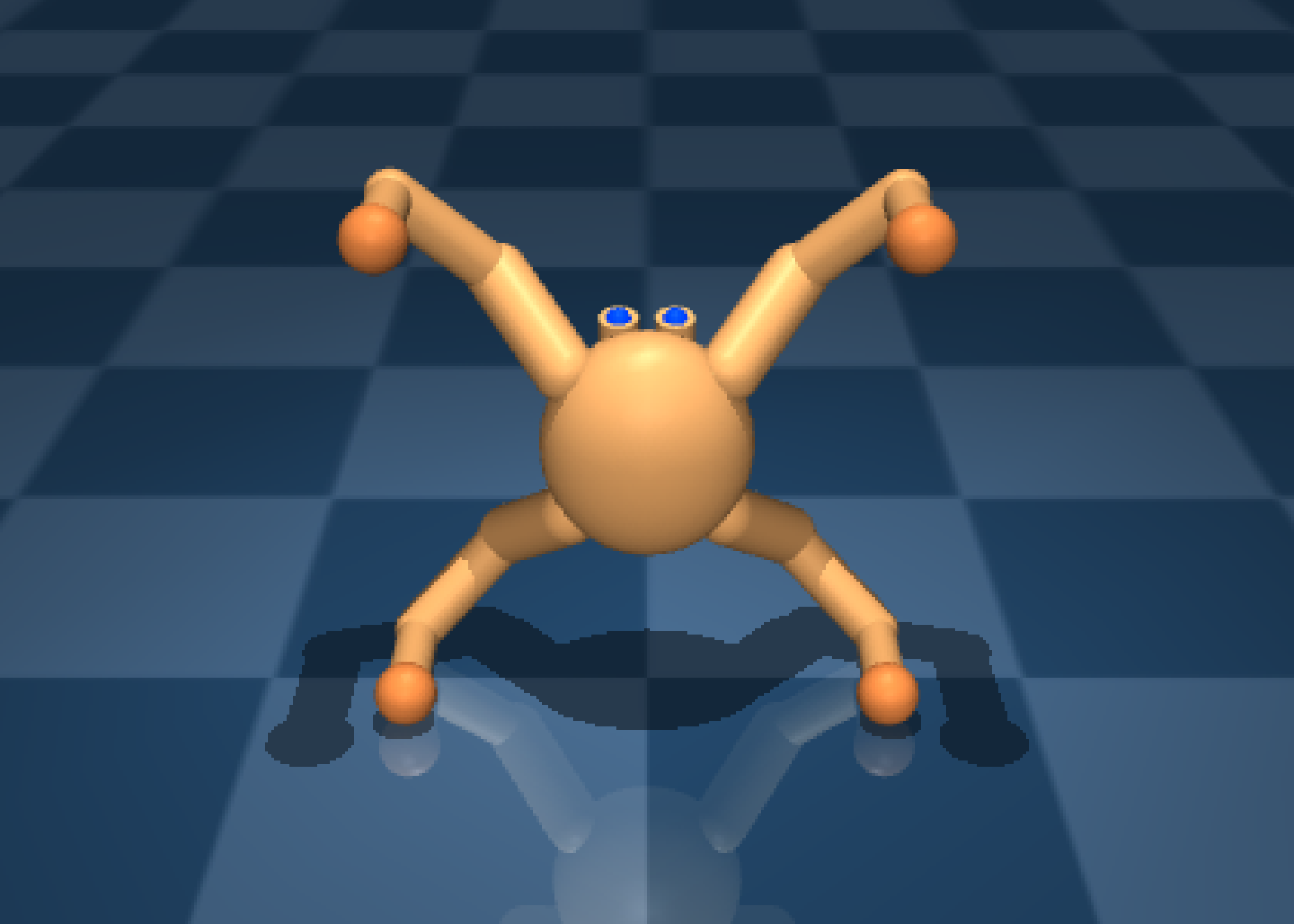}\vspace{-.3em}
         \centering{\scriptsize Balance Back}
     \end{subfigure}
 \begin{subfigure}[b]{0.137\textwidth}
         \centering
         \includegraphics[width=\textwidth]{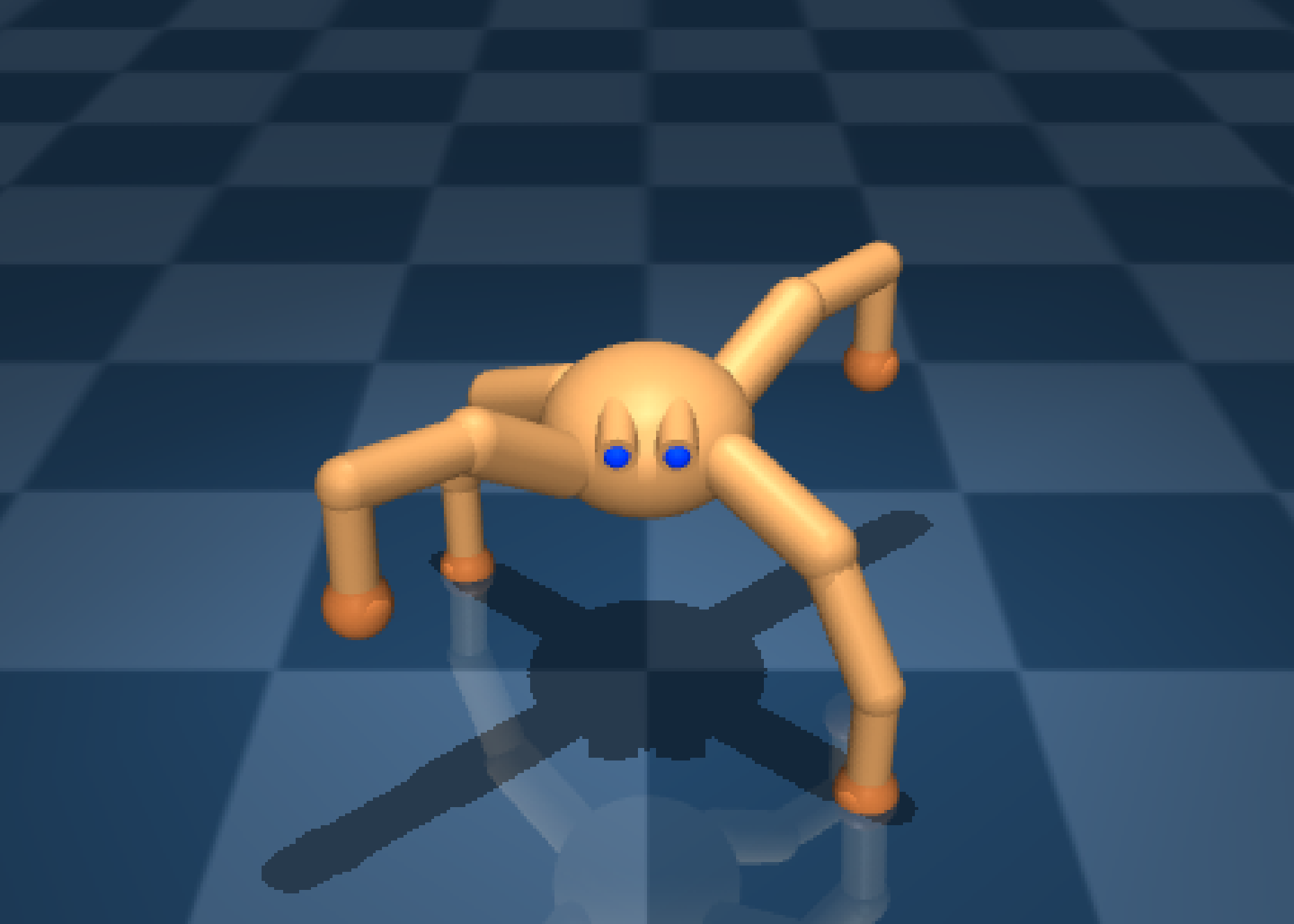}\vspace{-.3em}
         \centering{\scriptsize Balance Diagonal}
     \end{subfigure}
    \begin{subfigure}[b]{0.137\textwidth}
         \centering
         \includegraphics[width=\textwidth]{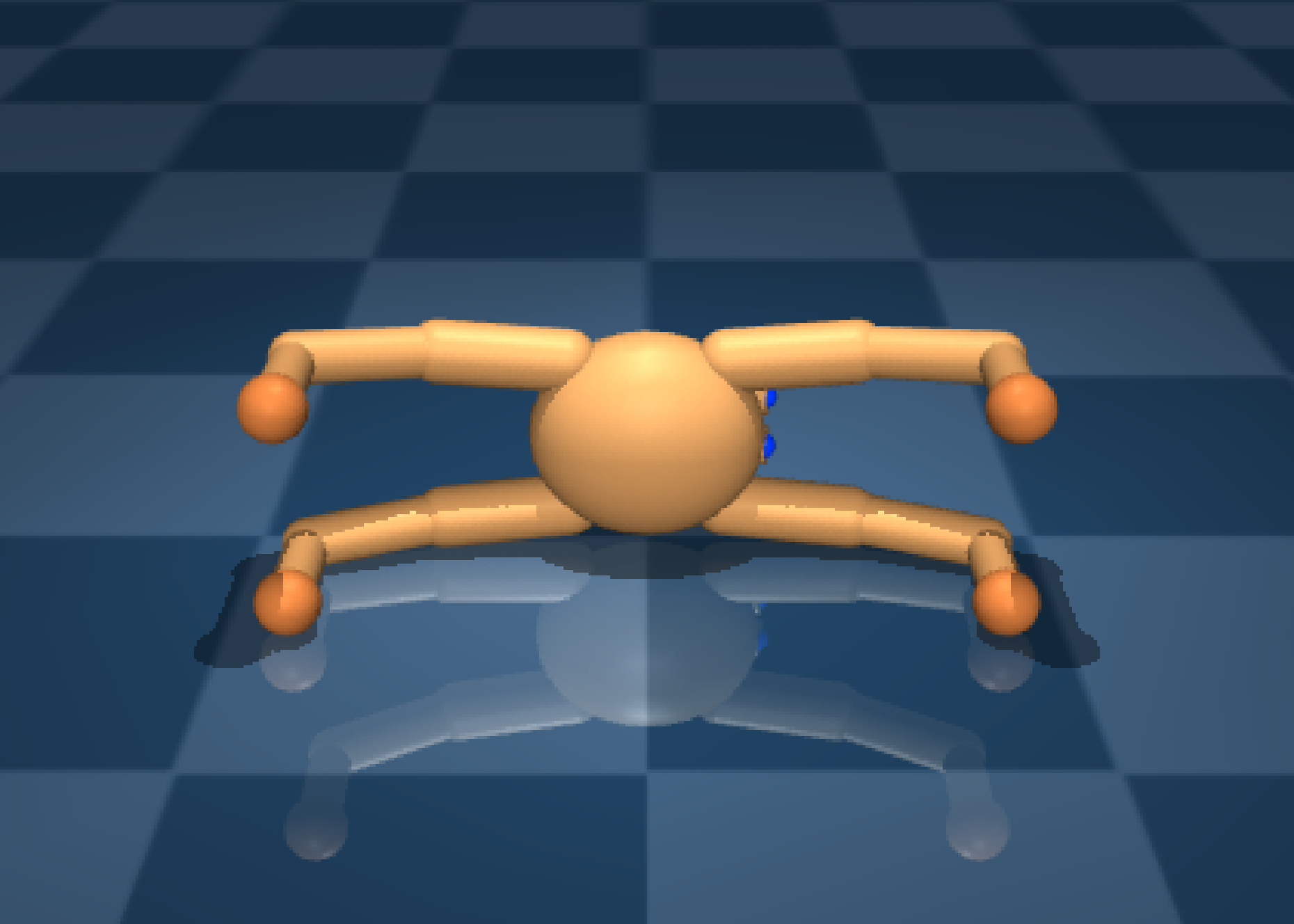}\vspace{-.3em}
                 \centering{\scriptsize Lie Side}
     \end{subfigure}
    \\
    \vspace{.3em}
    \begin{subfigure}[b]{0.137\textwidth}
         \centering
         \includegraphics[width=\textwidth]{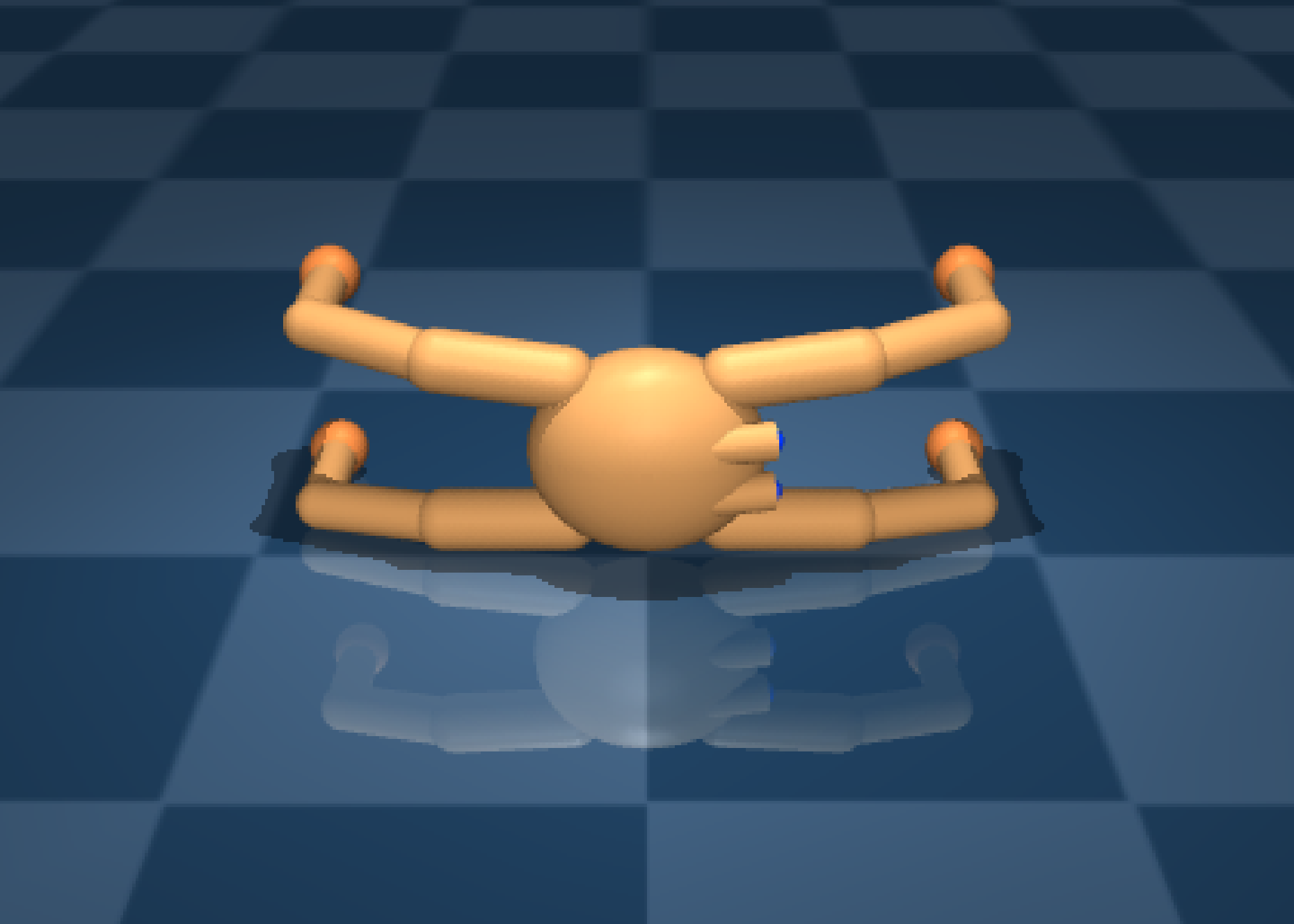}\vspace{-.3em}
         \centering{\scriptsize Lie Side Back}
     \end{subfigure}
      \begin{subfigure}[b]{0.137\textwidth}
         \centering
         \includegraphics[width=\textwidth]{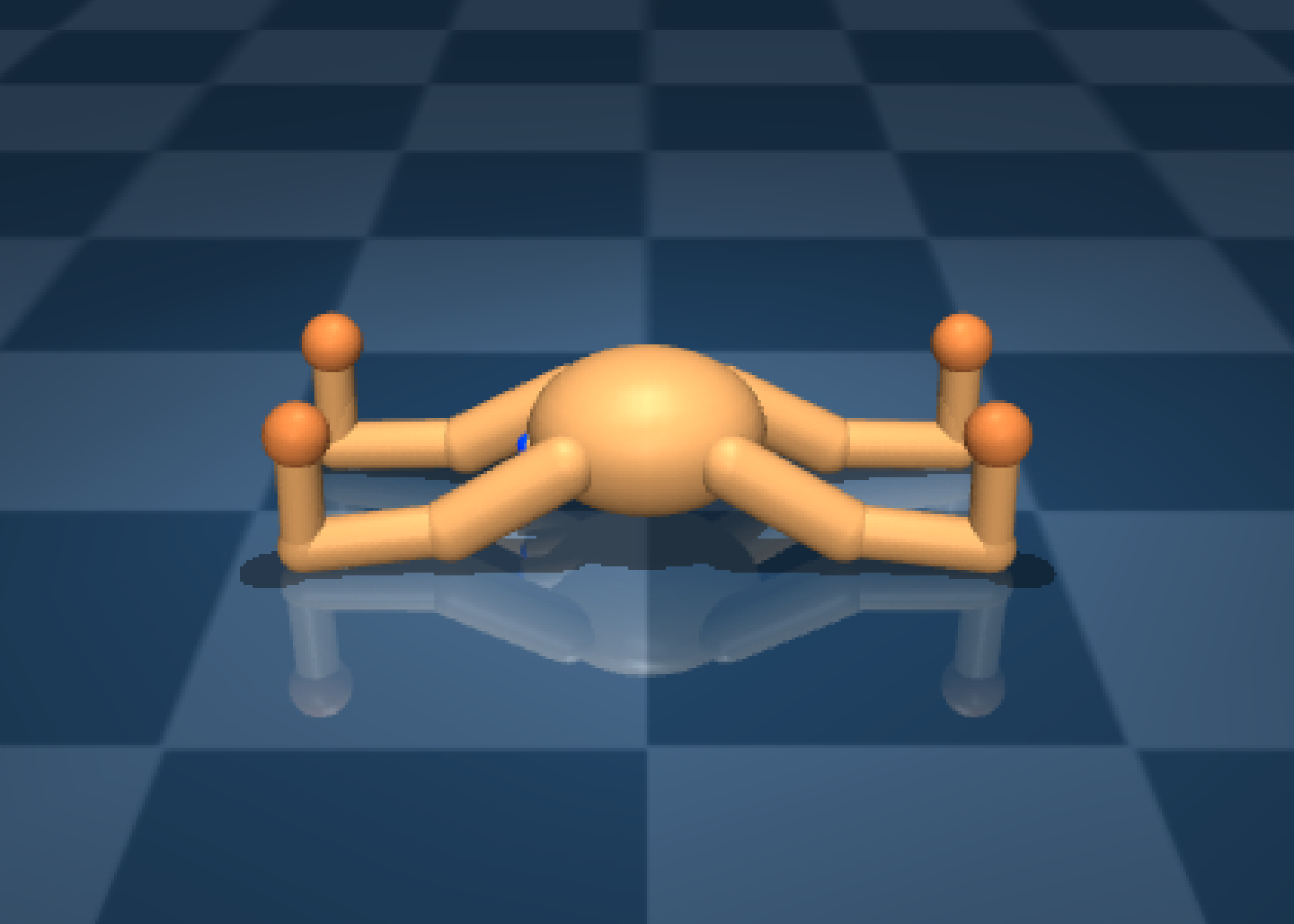}\vspace{-.3em}
         \centering{\scriptsize Lie Legs Together}
     \end{subfigure}
     \begin{subfigure}[b]{0.137\textwidth}
         \centering
         \includegraphics[width=\textwidth]{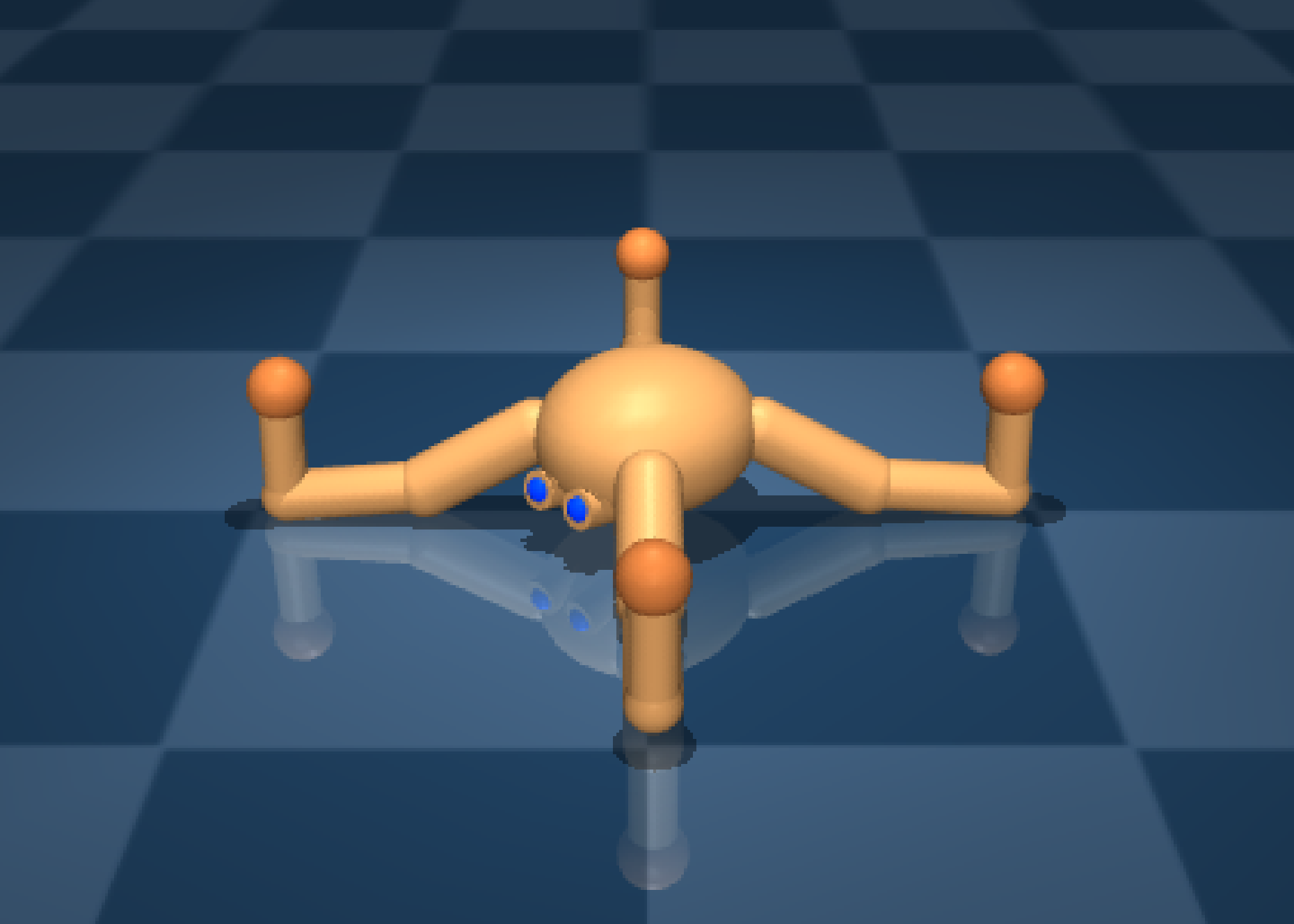}\vspace{-.3em}
         \centering{\scriptsize Lie Rotated}
     \end{subfigure}
 \begin{subfigure}[b]{0.137\textwidth}
         \centering
         \includegraphics[width=\textwidth]{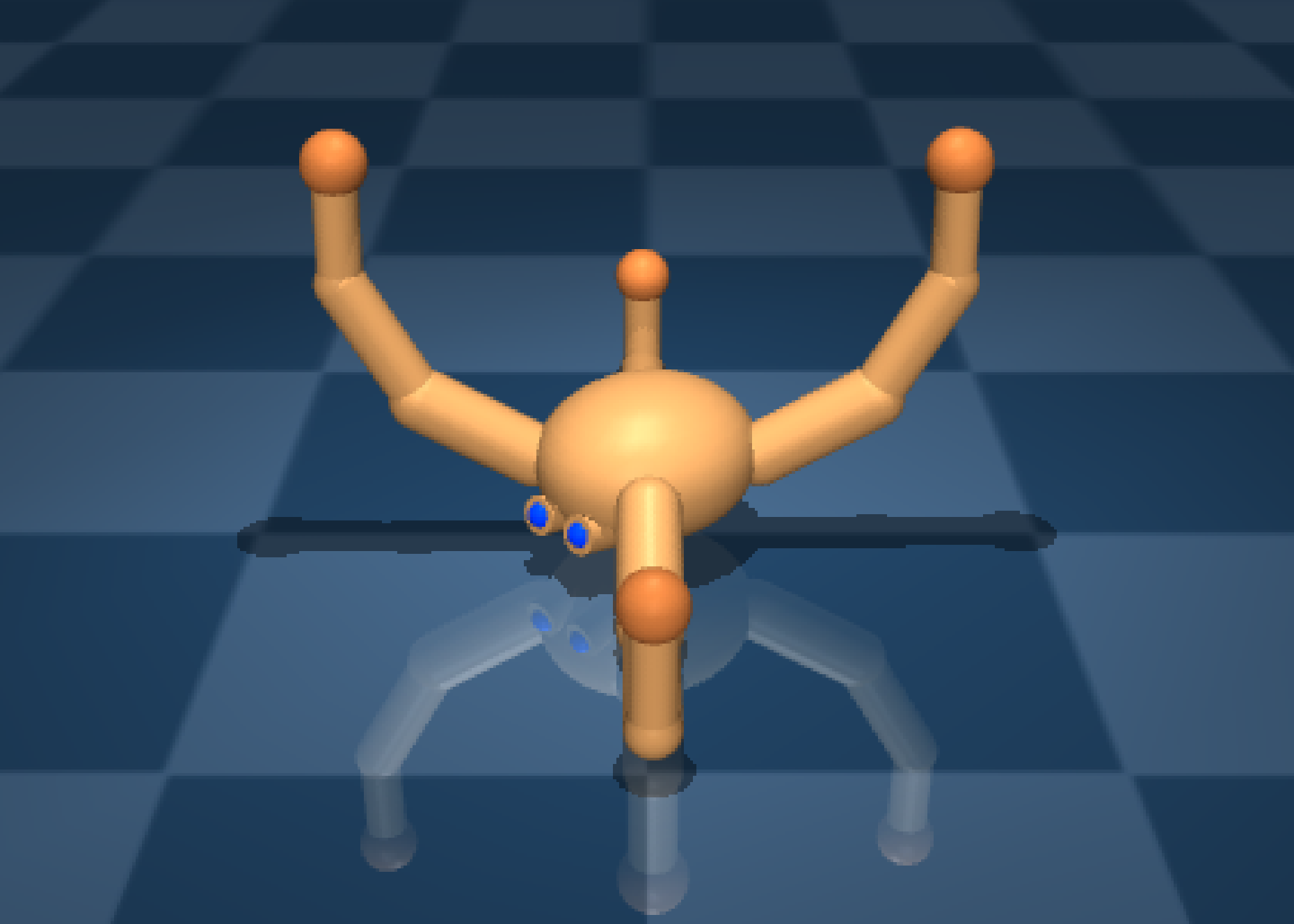}\vspace{-.3em}
         \centering{\scriptsize Lie Two Legs Up}
     \end{subfigure}
    \begin{subfigure}[b]{0.137\textwidth}
         \centering
         \includegraphics[width=\textwidth]{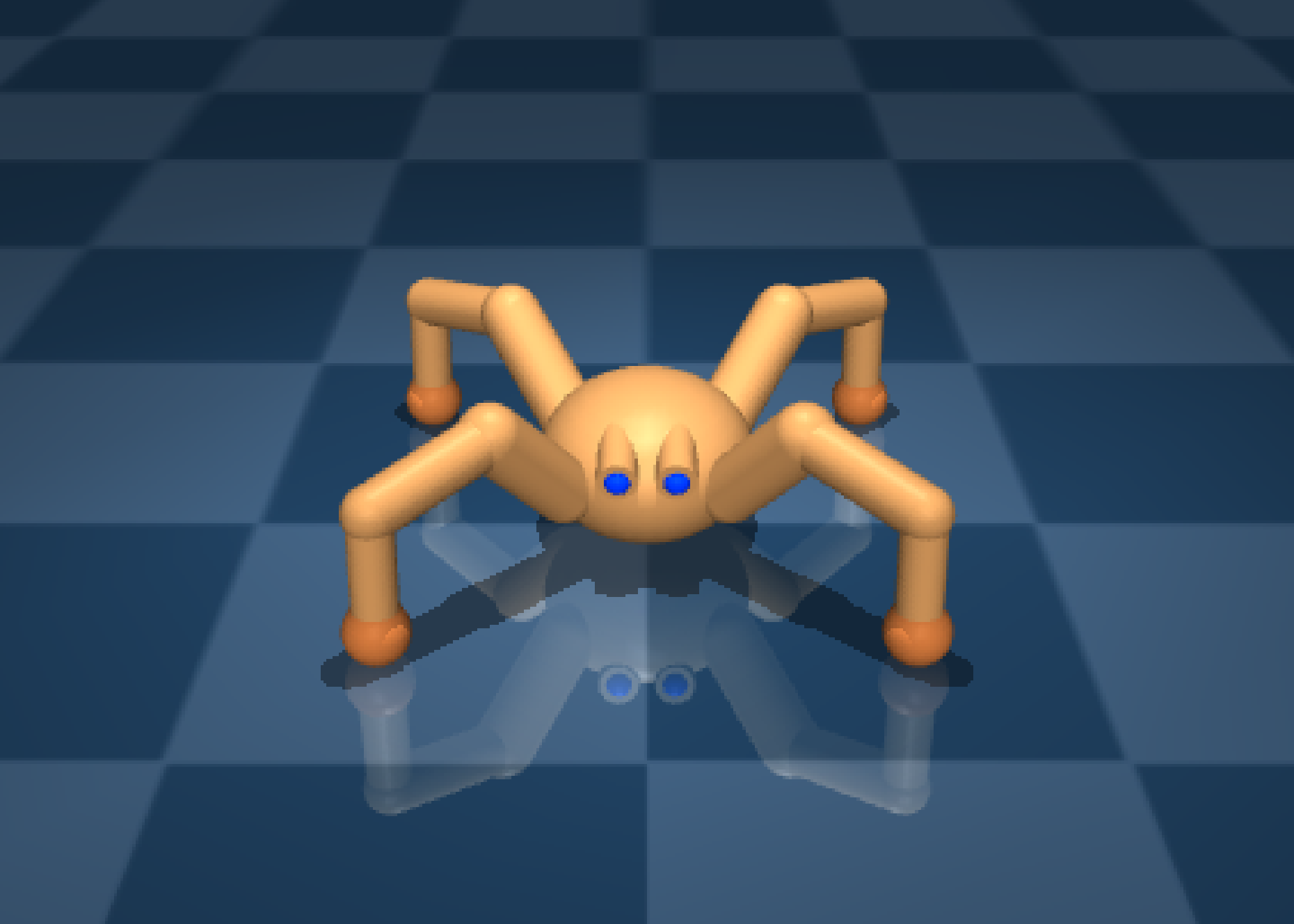}\vspace{-.3em}
                 \centering{\scriptsize Stand}
     \end{subfigure}
    \begin{subfigure}[b]{0.137\textwidth}
         \centering
         \includegraphics[width=\textwidth]{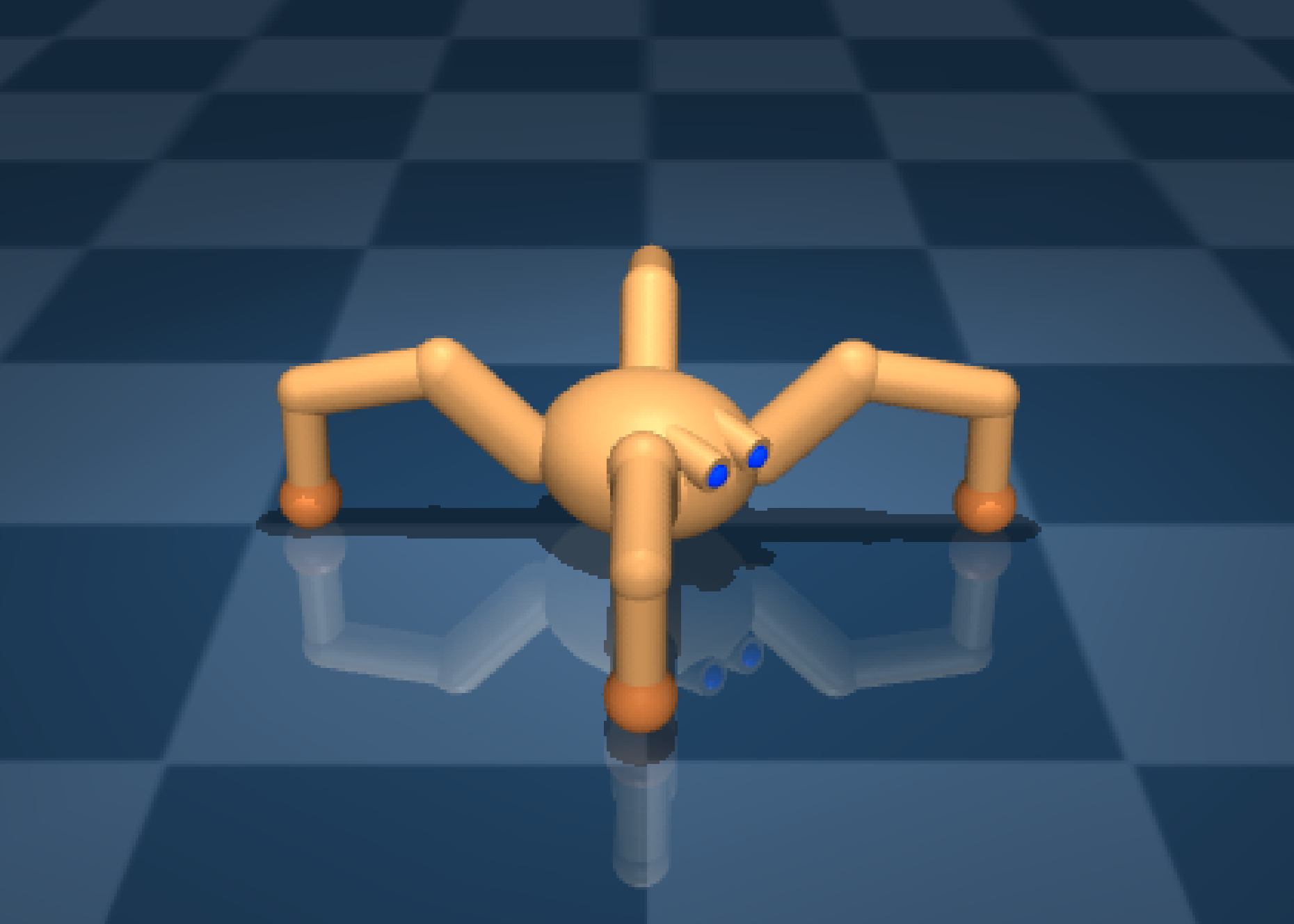}\vspace{-.3em}
         \centering{\scriptsize Stand Rotated}
     \end{subfigure}
    \\
    \vspace{1em}
     \begin{subfigure}[b]{0.137\textwidth}
        \centering
         \includegraphics[width=\textwidth]{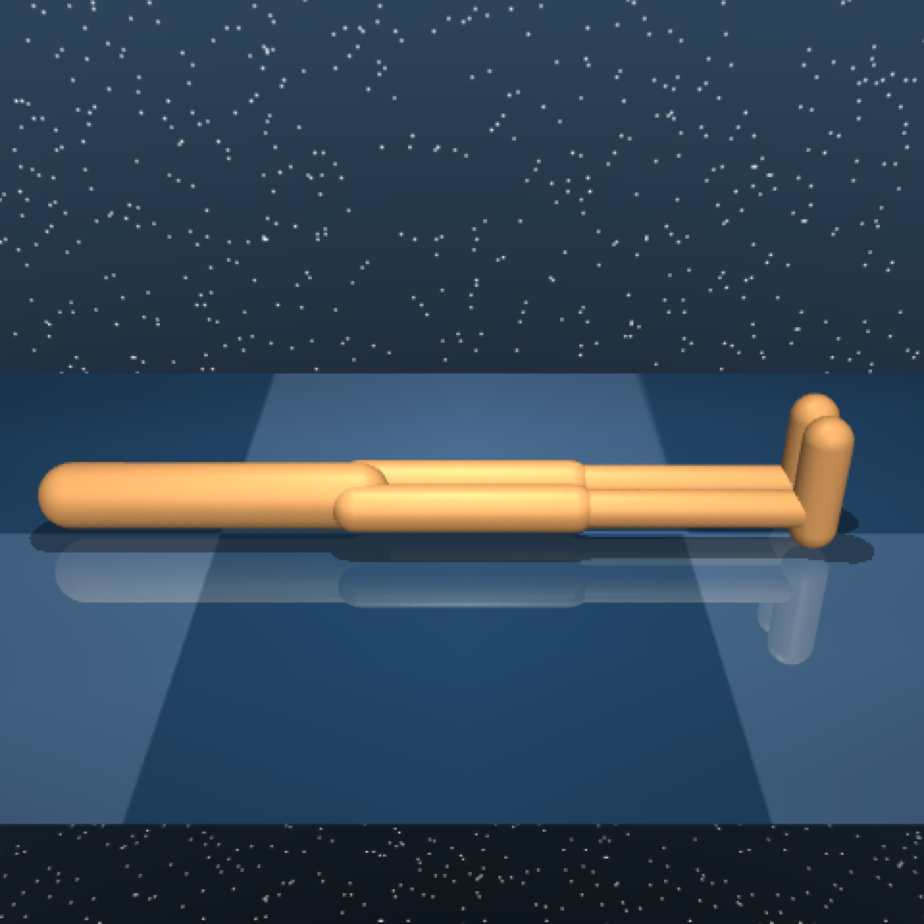}\vspace{-.3em}
         \centering{\scriptsize Lie Back}
     \end{subfigure}
     \begin{subfigure}[b]{0.137\textwidth}
         \centering
         \includegraphics[width=\textwidth]{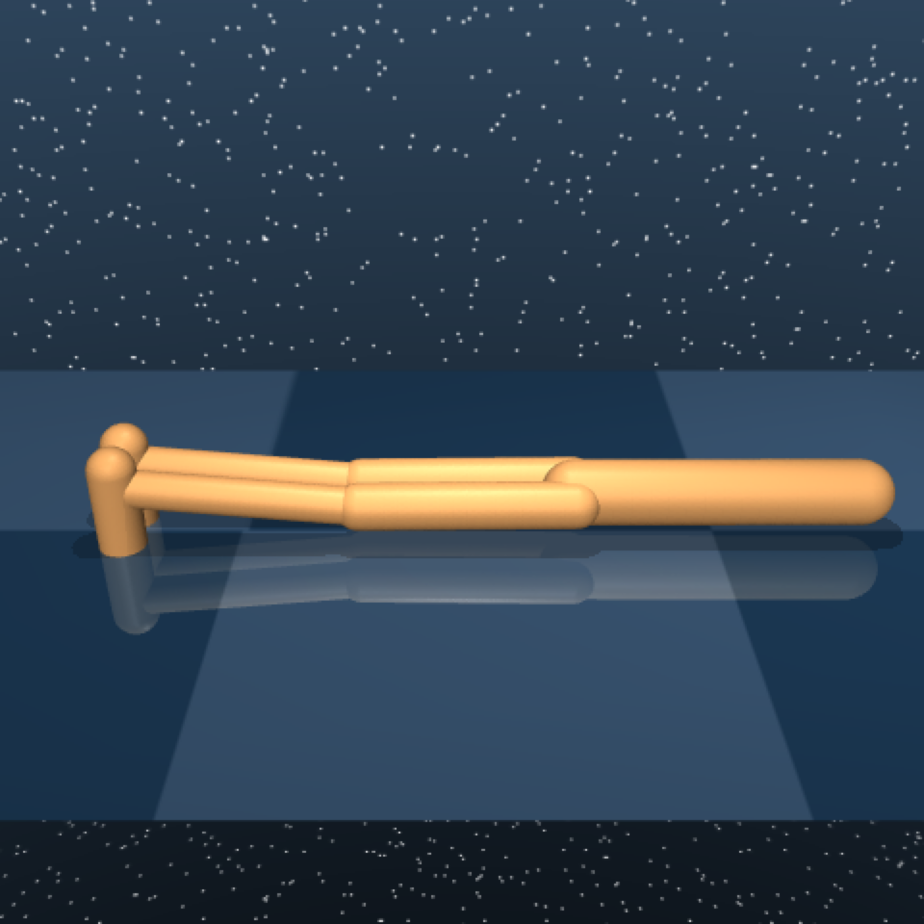}\vspace{-.3em}
         \centering{\scriptsize Lie Front}
     \end{subfigure}
      \begin{subfigure}[b]{0.137\textwidth}
         \centering
         \includegraphics[width=\textwidth]{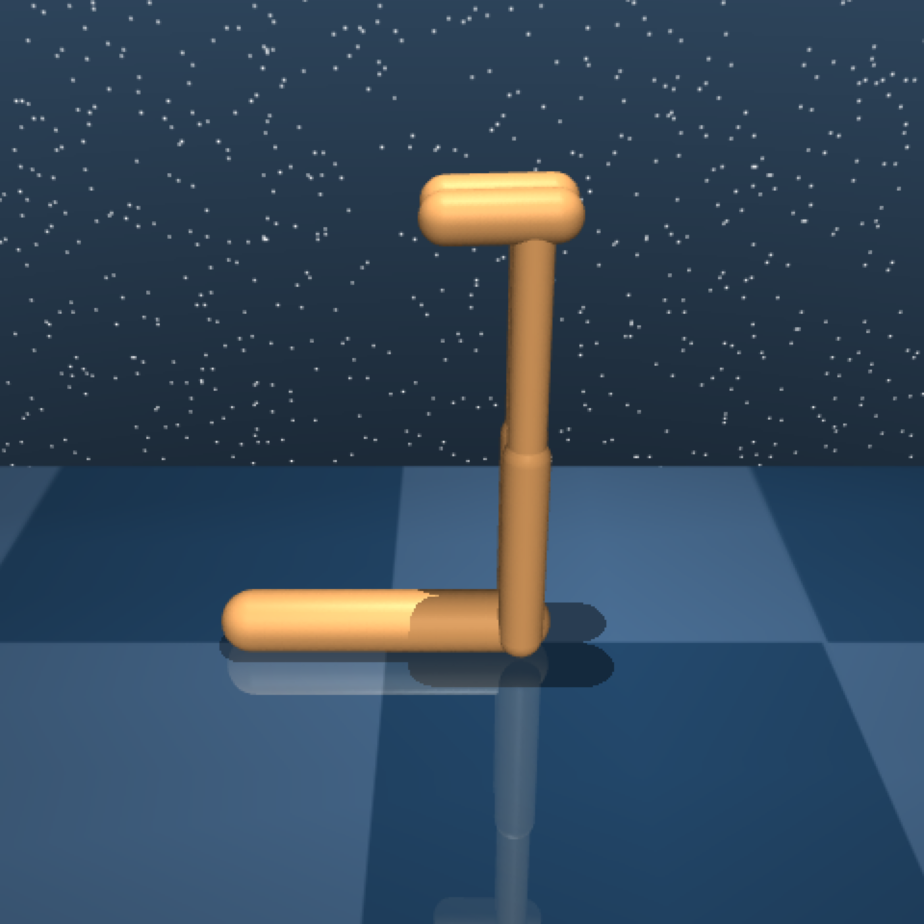}\vspace{-.3em}
         \centering{\scriptsize Legs Up}
     \end{subfigure}
     \begin{subfigure}[b]{0.137\textwidth}
         \centering
         \includegraphics[width=\textwidth]{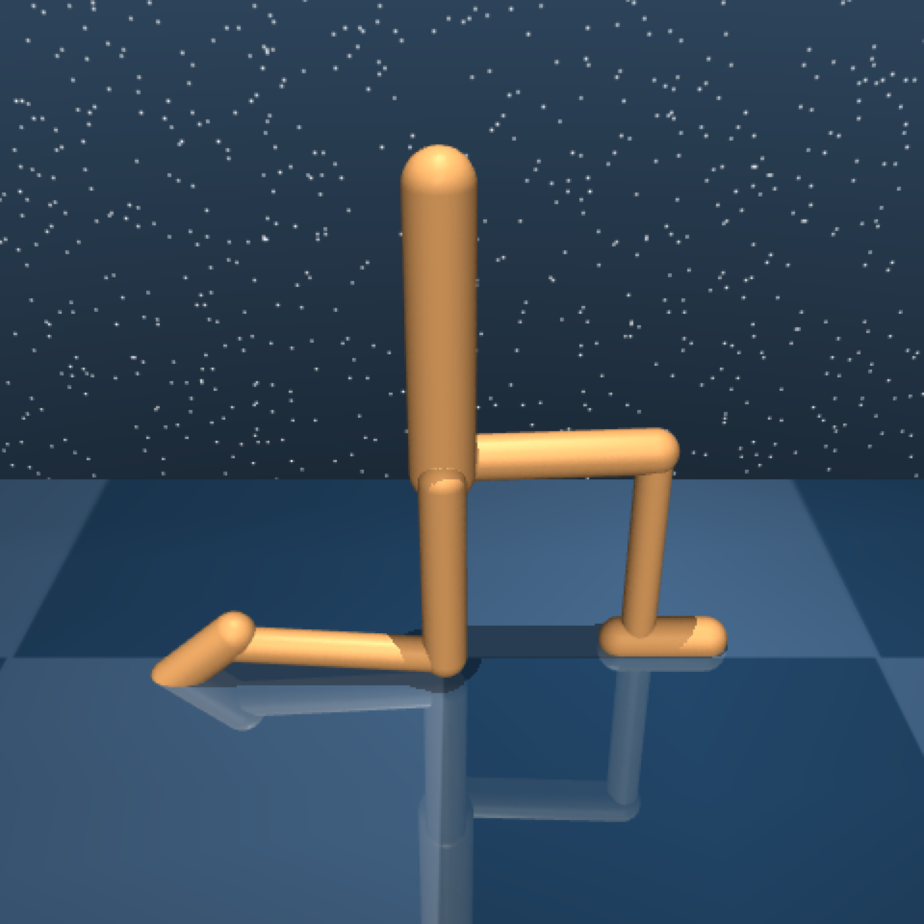}\vspace{-.3em}
         \centering{\scriptsize Kneel}
     \end{subfigure}
 \begin{subfigure}[b]{0.137\textwidth}
         \centering
         \includegraphics[width=\textwidth]{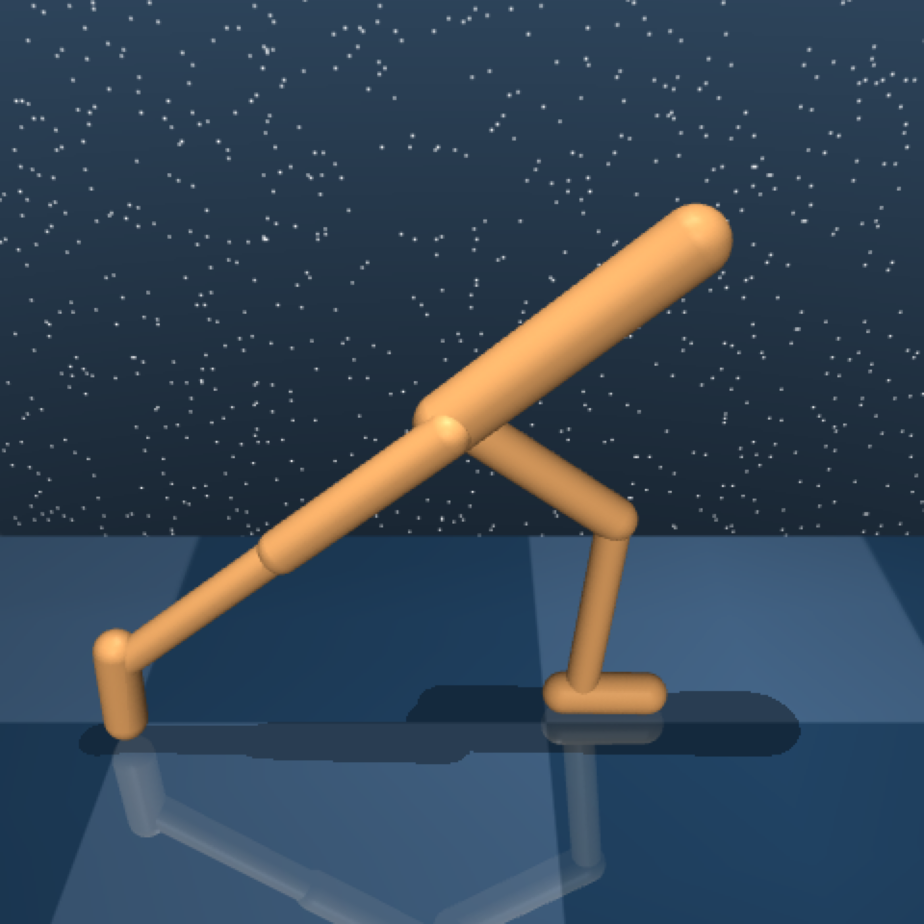}\vspace{-.3em}
         \centering{\scriptsize Side Angle}
     \end{subfigure}
    \begin{subfigure}[b]{0.137\textwidth}
         \centering
         \includegraphics[width=\textwidth]{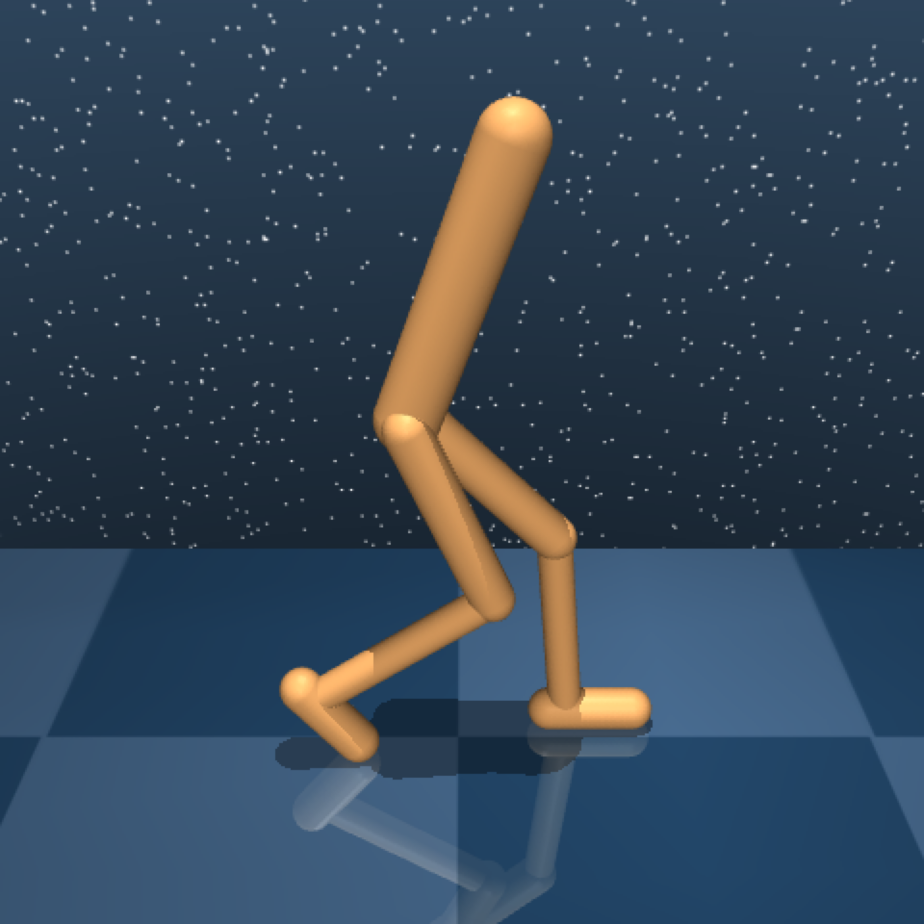}\vspace{-.3em}
                 \centering{\scriptsize Stand Up}
     \end{subfigure}
    \\
    \vspace{.3em}
    \begin{subfigure}[b]{0.137\textwidth}
         \centering
         \includegraphics[width=\textwidth]{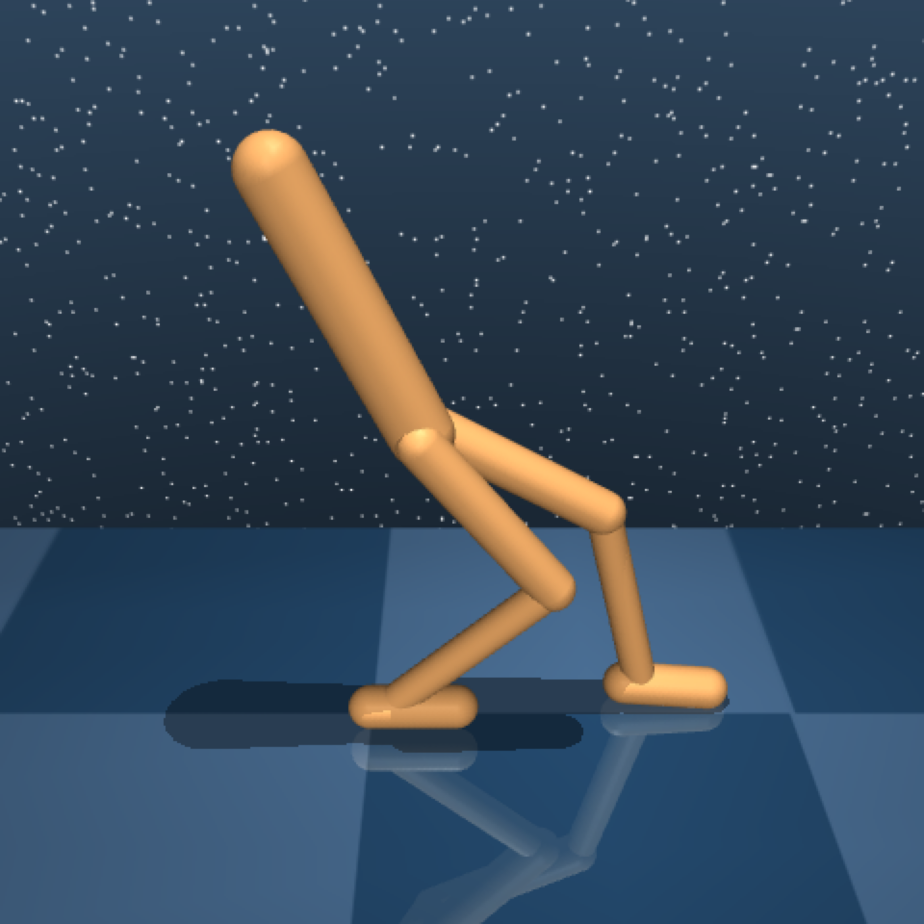}\vspace{-.3em}
         \centering{\scriptsize Lean Back}
     \end{subfigure}
    \begin{subfigure}[b]{0.137\textwidth}
         \centering
         \includegraphics[width=\textwidth]{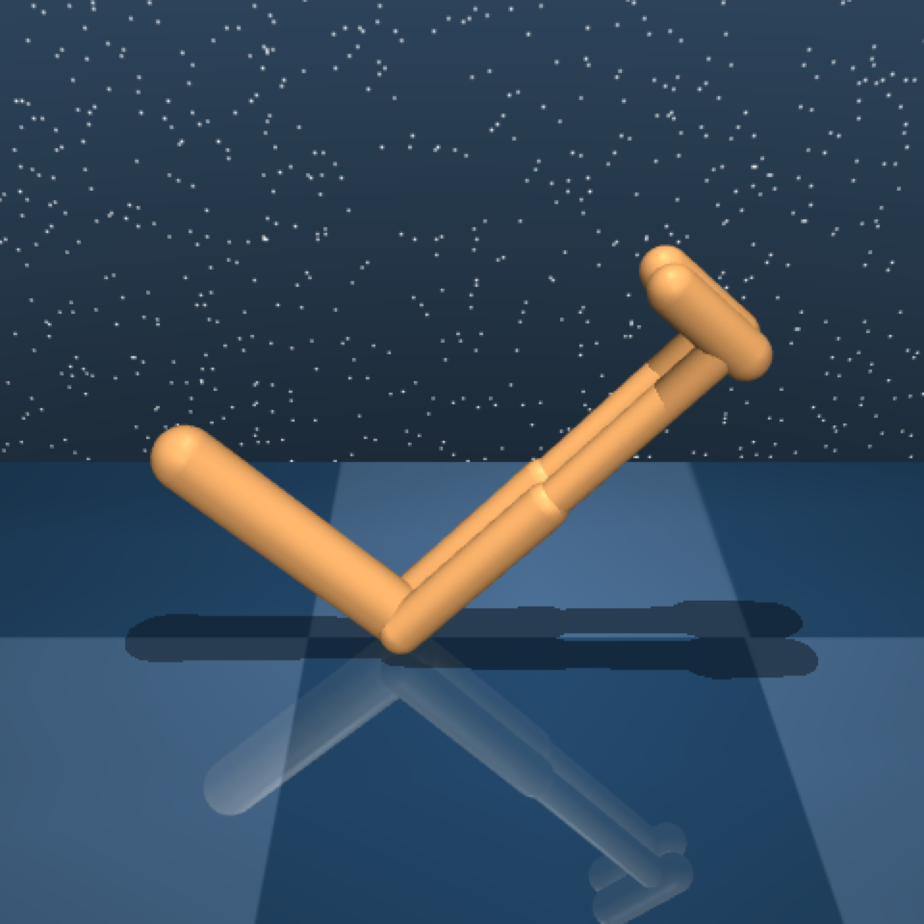}\vspace{-.3em}
         \centering{\scriptsize Boat}
     \end{subfigure}
    \begin{subfigure}[b]{0.137\textwidth}
         \centering
         \includegraphics[width=\textwidth]{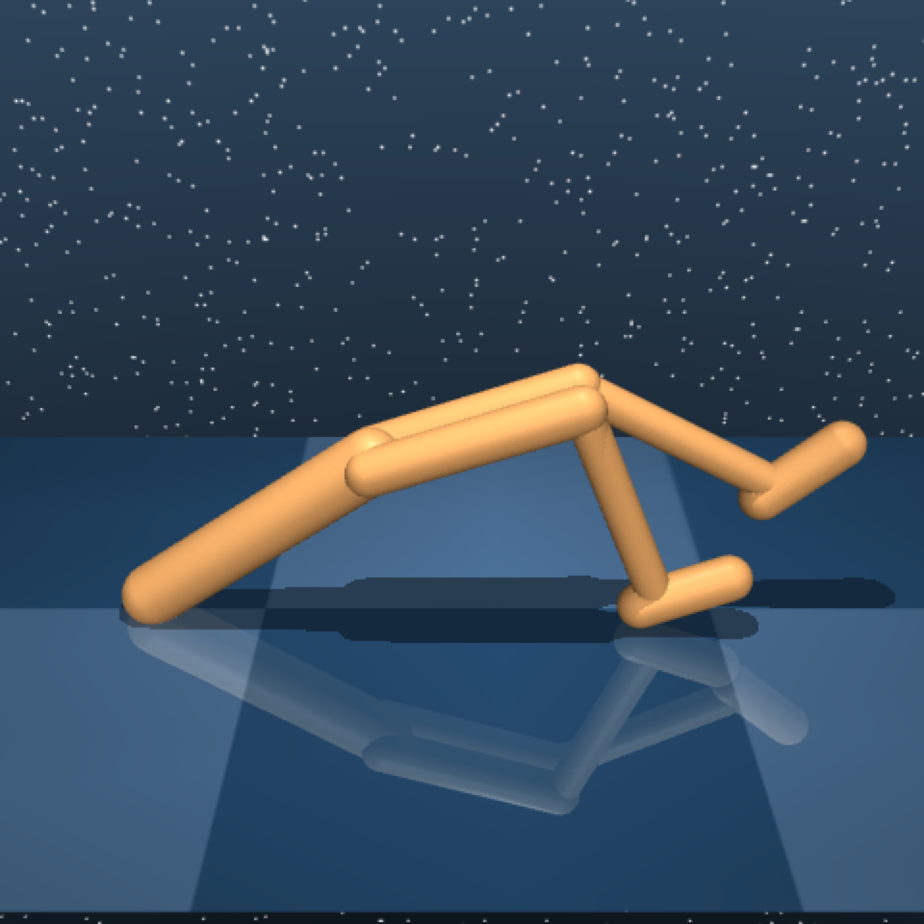}\vspace{-.3em}
         \centering{\scriptsize Bridge}
     \end{subfigure}
    \begin{subfigure}[b]{0.137\textwidth}
         \centering
         \includegraphics[width=\textwidth]{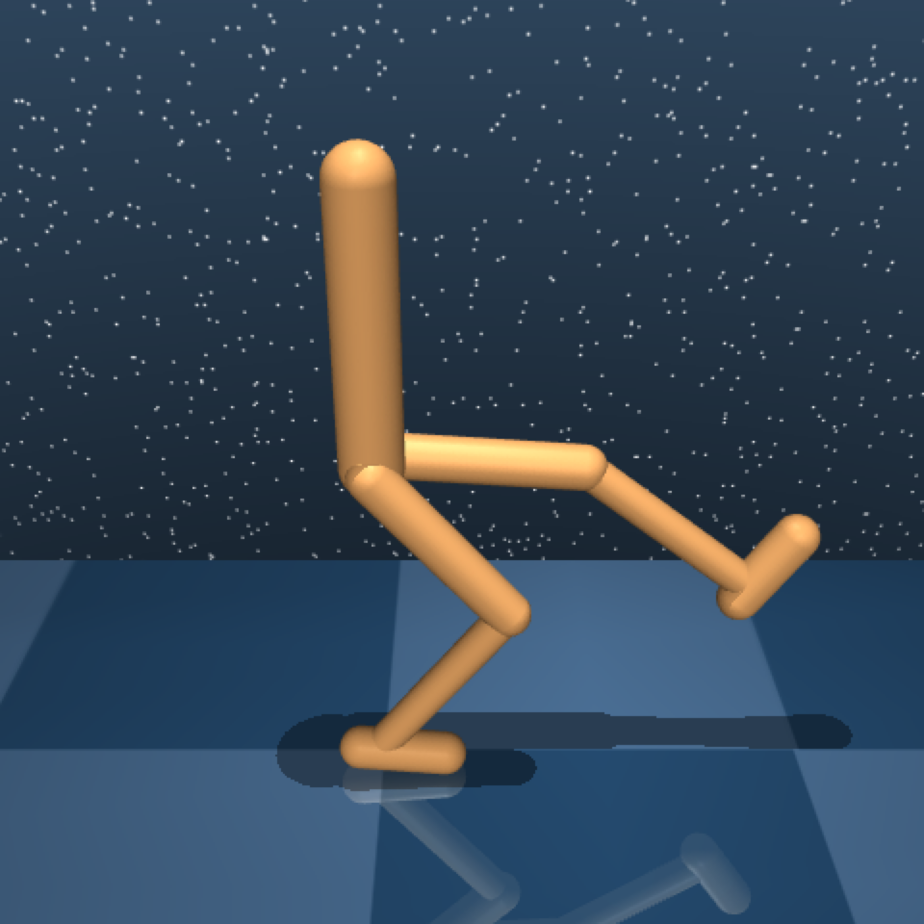}\vspace{-.3em}
         \centering{\scriptsize One Foot Up}
     \end{subfigure}
    \begin{subfigure}[b]{0.137\textwidth}
         \centering
         \includegraphics[width=\textwidth]{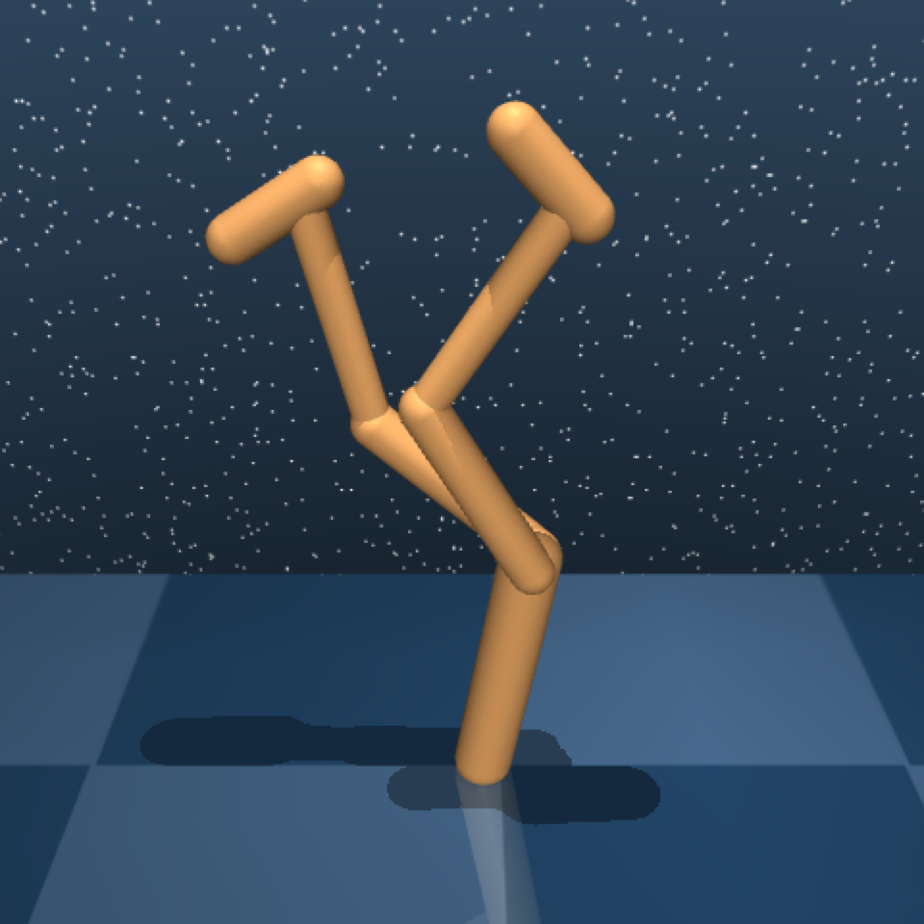}\vspace{-.3em}
         \centering{\scriptsize Head Stand}
     \end{subfigure}
    \begin{subfigure}[b]{0.137\textwidth}
         \centering
         \includegraphics[width=\textwidth]{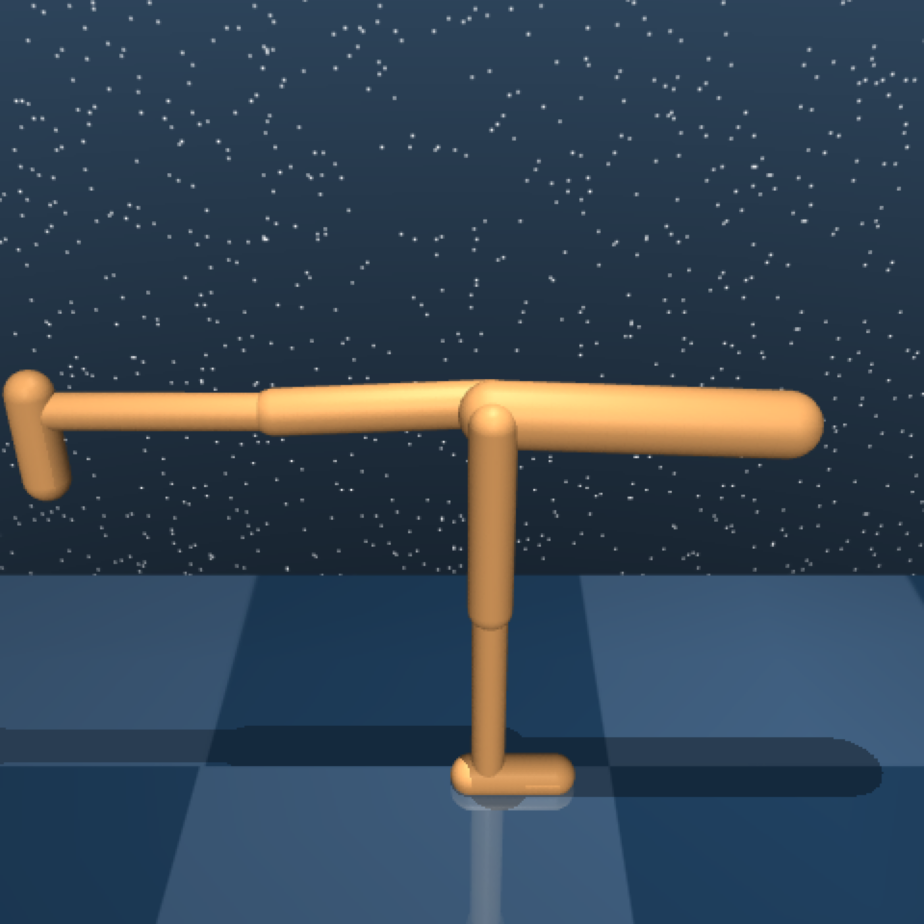}\vspace{-.3em}
         \centering{\scriptsize Arabesque}
     \end{subfigure}
     \vspace{.2em}
  \caption{{\bf{Goals in the RoboYoga benchmark}} as proposed in \citet{Mendonca2021:LEXA} for Quadruped (top 2 rows) and Walker environments (bottom 2 rows).}
       \label{fig:quadruped_goals}
\end{figure}
\begin{figure}
    \centering
    {\scriptsize
      \quad \textcolor{ours}{\rule[2pt]{20pt}{1.2pt}} \ourMethodA{}
       \quad \textcolor{cee_us_test}{\rule[2pt]{20pt}{1.2pt}} \oldMethod
    }\\ \vspace{.3em}
    \begin{subfigure}[b]{0.01\textwidth}
        \notsotiny{\rotatebox{90}{\hspace{.5cm}\textsf{success rate}}}
    \end{subfigure} 
    \begin{subfigure}[t]{.24\linewidth}
    \includegraphics[width=\linewidth]{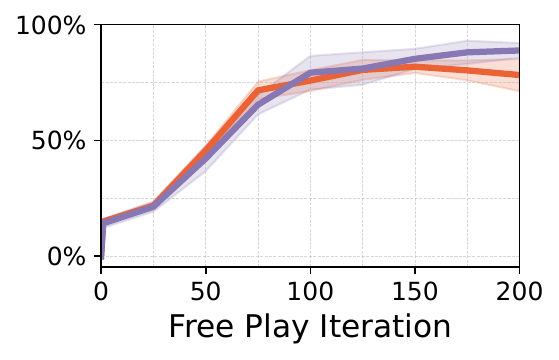}\vspace{-.3em}
    \caption{Stand Leg Up}
    \end{subfigure}
    \begin{subfigure}[t]{.24\linewidth}
    \includegraphics[width=\linewidth]{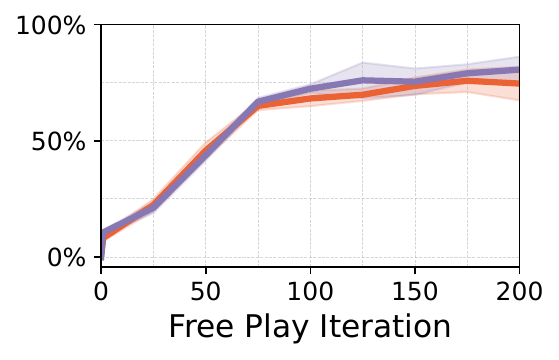}\vspace{-.3em}
    \caption{Attack}
    \end{subfigure}
    \begin{subfigure}[t]{.24\linewidth}
    \includegraphics[width=\linewidth]{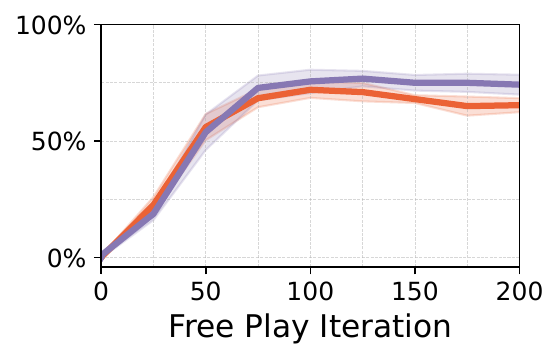}\vspace{-.3em}
    \caption{Balance Front}
    \end{subfigure}
    \begin{subfigure}[t]{.24\linewidth}
    \includegraphics[width=\linewidth]{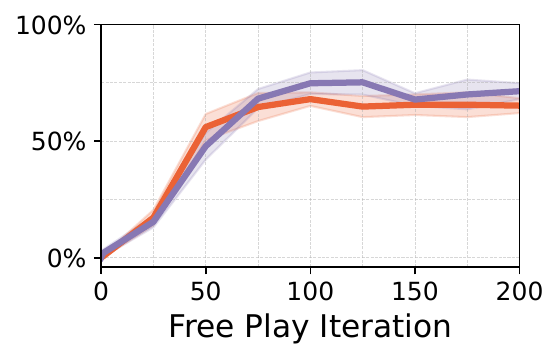}\vspace{-.3em}
    \caption{Balance Back}
    \end{subfigure}
     \\
     \vspace{.2em}
    \begin{subfigure}[b]{0.01\textwidth}
    \notsotiny{\rotatebox{90}{\hspace{.5cm}\textsf{success rate}}}
    \end{subfigure} 
    \begin{subfigure}[t]{.24\linewidth}
    \includegraphics[width=\linewidth]{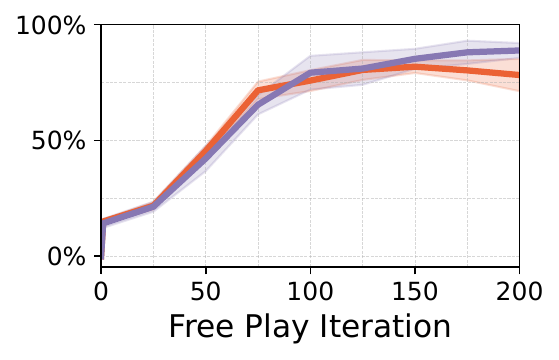}\vspace{-.3em}
    \caption{Balance Diagonal}
    \end{subfigure}
    \begin{subfigure}[t]{.24\linewidth}
    \includegraphics[width=\linewidth]{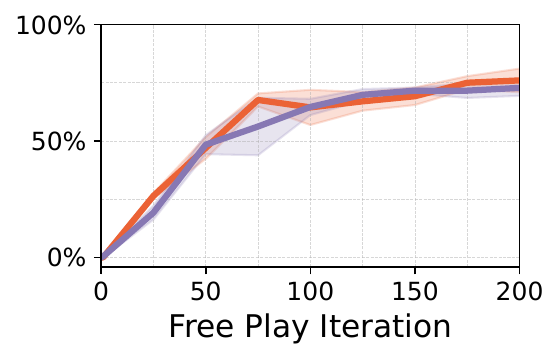}\vspace{-.3em}
    \caption{Lie Side}
    \end{subfigure}
    \begin{subfigure}[t]{.24\linewidth}
    \includegraphics[width=\linewidth]{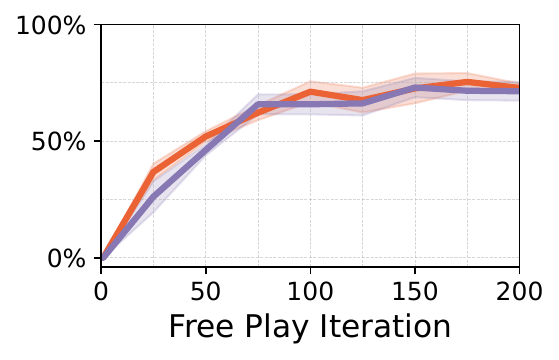}\vspace{-.3em}
    \caption{Lie Side Back}
    \end{subfigure}
    \caption{{\bf Downstream task performance for a subset of Quadruped RoboYoga tasks} with \ourMethodA ($\lambda=0.02$) and \oldMethod, evaluated for models checkpointed throughout free play. We use absolute relational $\phi$ for \ourMethod computations. (Five independent seeds)
    }
    \label{fig:roboyoga_quadruped_tasks}
\end{figure}

We run free play with \ourMethodA ($\lambda=0.02$) and \oldMethod in the Quadruped environment. With \ourMethodA, we indeed observe more regular poses during free play, as reflected in the \ourMethod values shown in \fig{fig:rair_best_quadruped}. We evaluate the zero-shot downstream performance of the learned models for the Roboyoga poses shown in \fig{fig:quadruped_goals}. The success rates for the models at the end of free play (\tab{tab:quadruped_tasks}), as well as their temporal evolution (\fig{fig:roboyoga_quadruped_tasks}) are reported. We evaluate success for a pose as having a distance smaller than a threshold of 0.4 to the goal pose for more than 10 timesteps. We only see marginal improvements with \ourMethodA compared to \oldMethod. We hypothesize that this is because model-based approaches can generalize very well in locomotion environments and \oldMethod also provides sufficient exploration. It only comes down to whether the pose can be held longer stably, where in the balancing poses \ourMethodA has a slight edge over the course of free play (especially for \emph{Balance Front}). For any goal that is shown in \fig{fig:quadruped_goals} and not included in \tab{tab:quadruped_tasks} (\ie the poses with Quadruped lying on its back), we get perfect success rates for both \ourMethodA and \oldMethod. We didn't run free play in the Walker environment, but we expect model-based approaches to be very strong there altogether, as the environment dynamics are very simple.

\begin{table}
    \centering
  \caption{{\bf Zero-shot downstream task generalization performance of \ourMethodA and \oldMethod for the Quadruped RoboYoga benchmark.} Results are shown for five independent seeds, for models evaluated after 200 free play iterations (equivalent to 2K $\times \, 200 = 400$K transitions).}
  \centering
    \renewcommand{\arraystretch}{.99}
    \resizebox{1.\linewidth}{!}{ %
    \begin{tabular}{@{}lccccccccc@{}}
    \toprule
    & Stand  & Attack & Balance   & Balance   & Balance  & Lie Side & Lie Side & Stand & Stand \\
    &  Leg Up &  &  Front  &  Back  &  Diagonal &  &  Back & &Rotated\\
    \midrule
    \ourMethodA & $0.89 \pm 0.03$  & $0.81 \pm 0.06$ & $0.74 \pm 0.04$ & $0.71 \pm 0.03$ & $0.89 \pm 0.03$ &$0.73 \pm 0.03$ & $0.71 \pm 0.04$ & $0.89 \pm 0.04$ & $0.85 \pm 0.06$\\
     \oldMethod & $0.78 \pm 0.07$ & $0.75 \pm 0.07$  & $0.65 \pm 0.03$  & $0.65 \pm 0.03$ & $0.78 \pm 0.07$ & $0.76 \pm 0.05$ & $0.73 \pm 0.02$ & $0.78 \pm 0.07$ & $0.80 \pm 0.06$\\
  \bottomrule
    \end{tabular}
  }
    \label{tab:quadruped_tasks}
\end{table}

\section{\oldMethod} \label{app:cee_us}
In this section, we present the details of \oldMethod \cite{Sancaktaretal22}, which we build upon in this work. \oldMethod uses structured world models together with model-based planning during exploration, achieving increased sample-efficiency and superior downstream task performance compared to other intrinsically-motivated RL baselines. The free-play pseudocode is presented in \alg{alg:intrinsic}. This free-play structure is used for all methods presented in our paper by swapping out the intrinsic reward term (line 5) with only ensemble disagreement (\oldMethod), combination of our regularity objective with ensemble disagreement (\ourMethodA) or pure regularity (\ourMethod), as well as the other baselines RND and Dis (see \sec{app:baselines_detail}).

\algrenewcommand\algorithmicindent{1em}
\algrenewcommand{\algorithmiccomment}[1]{\bgroup\hskip1em\textcolor{ourorange}{\hfill$\rhd$~\textsl{#1}}\egroup}
\algrenewcommand{\alglinenumber}[1]{\color{gray}\scriptsize #1:}

\begin{algorithm}[H]
  \caption{\bf Free Play in Intrinsic Phase (taken from \cite{Sancaktaretal22})} \label{alg:intrinsic}
  \begin{algorithmic}[1]
    \State \textbf{Input:} $\{(\tilde{f}_{{\theta}_m})_{m=1}^M\}$: Randomly initialized ensemble of GNNs with $M$ members, $D$: empty dataset, \texttt{Planner}: iCEM planner with horizon $H$
      \While{explore}\Comment{Explore with MPC and intrinsic reward}
        \For{$e = 1$ \textbf{to} $\texttt{num\_episodes}$}
            \For{$t = 1$ \textbf{to} $T$} \Comment{Plan to maximize intrinsic reward}
                \State $a_t \gets \texttt{Planner}(s_t, \{(\tilde{f}_{{\theta}_m})_{m=1}^M\}, r_\mathrm{intrinsic})$ \Comment{\eg \ourMethod with disagreement \eqn{eqn:combination}}
                \State $s_{t+1} \gets \texttt{env.step}(s_t, a_t)$
            \EndFor
            \State $\mathcal{D} \leftarrow \mathcal{D} \cup \{(s_t, a_t, s_{t+1})_{t=1}^T\}$
        \EndFor
      \For{$l = 1$ \textbf{to} $L$}\Comment{Train models on dataset for $L$ epochs}
         \State $\theta_m \gets$ optimize $\theta_m$ using $\mathcal L_m$ on $\mathcal D$ for $m=1,\dots,M$%
      \EndFor
      \EndWhile
      \State \textbf{return} $\{(\tilde{f}_{{\theta}_m})_{m=1}^M\}, \mathcal{D}$
  \end{algorithmic}
\end{algorithm}

\subsection{GNN Architectural Details} \label{app:gnn_architecture}
Message-passing Graph Neural Networks (GNN) are deployed as world models. The same GNN architecture is used as in \oldMethod \cite{Sancaktaretal22}. For these structured world models, we consider object-factorized state spaces with $\mathcal{S} = (\mathcal{S}_{\obj})^N \times \mathcal{S}_{\textrm{robot}}$.
Each node in the GNN corresponds to an object and the robot/actuated agent is differentiated from the object nodes as a global node.
The concatenation of the robots's state $s_t^{\text{robot}}$ and the action $a_t$ is represented as a global context $c = [s_t^{\text{robot}}, a_t]$.
We have a fully-connected GNN. 
The node update function $g_{\text{node}}$ and the edge update function $g_{\text{edge}}$ model the dynamics of the entities/objects, and their pairwise interactions respectively. These functions are both Multilayer Perceptrons (MLP). In the following, we denote the state of the $i$-th object $s_{t, \text{obj}_i}$ at timestep $t$ as $s_{t}^{i}$ for simplicity.
The object node attributes in the GNN are updated as:
\begin{align}
    e_t^{(i,j)} &= g_{\text{edge}}\big(\big[s_t^i, s_t^j, c\big]\big) \\
    \tilde{s}_{t+1}^{i} &= g_{\text{node}}\big(\big[s_t^i, c, \operatorname{aggr}_{i \neq j}\big(e_t^{(i,j)}\big)\big]\big).
\end{align}
where $[\cdot,\ldots]$ denotes concatenation, $e_t^{(i,j)}$ is the edge attribute between two neighboring nodes $(i, j)$. For the permutation-invariant aggregation function given by $\operatorname{aggr}$, we use the mean.

The robot state, which is treated as a global node, is computed using the global aggregation of all edges with a separate global node MLP $g_{\text{global}}$:
\begin{equation}
     \tilde{s}_{t+1}^{\text{robot}} = g_{\text{global}}\big(\big[c, 
     \operatorname{aggr}_{i,j}\big(e_t^{(i,j)}\big)\big]\big).
     \label{}
\end{equation}

Moreover, the GNN predicts the changes in the dynamics such that $\tilde{s}_{t+1} = s_t + \mathrm{GNN}(s_t,a_t)$.

\subsection{Planning Details}\label{app:icem}
For planning, we use the improved Cross-Entropy Method (iCEM)~\cite{pinneri2020:iCEM}. The planner minimizes the cost, corresponding to negative reward $c(s_t, a_t, s_{t+1}) = -r(s_t, a_t, s_{t+1}),$
where $r$ can be intrinsic rewards $r_\text{intrinsic}$ or extrinsic task rewards $r_\text{task}$. The extrinsic task rewards are assumed to be given by the environment.

At each timestep $t$ in the environment, the planner samples $P$ action sequences, each with length $H$, \ie the planning horizon. These actions are rolled out either in the ground truth model (perfect simulations) or in the imagination of a learned model (imperfect simulations), generating corresponding $P$ state sequences with length $H$. In order to assign a cost to each of the $P$ trajectories, we need to aggregate the cost over the horizon $H$. A typical choice here is $\texttt{sum}$, where the cost over the length of the trajectory is simply summed up: $\texttt{cost}^{(p)} = \sum_{h=0}^{H-1}c(s_{t+h}^{(p)}, a_{t+h}^{(p)}, s_{t+h+1}^{(p)})$.

However, this type of aggregation is not suitable for cases where a decrease in cost can in general be preceded by an initial increase. In these cases, using the mode $\texttt{best}$, that assigns the plan $p$ the cost of the ``best'' timestep over the planning horizon with $\texttt{cost}^{(p)}=\min\left(\{c(s_{t+h}^{(p)}, a_{t+h}^{(p)}, s_{t+h+1}^{(p)})\}_{h=0}^{H-1}\right)$ is a better suited choice.
We also empirically found this controller mode to be better at picking up \emph{sparse} signals. What we mean here is that, in the example of stacking, it is hard to find a sampled trajectory that stacks the objects in a stable way with a limited sample-budget as this poses an exploration challenge. However, if we manage to find an action sequence that brings the cubes on top of each other, albeit in an unstable way, favoring this solution with \texttt{best} and keeping this solution in the elite set is beneficial. This can be explained as follows: In iCEM the $K$ plans with the lowest assigned cost are chosen to be the elite set, which is then used to fit the sampling distribution of iCEM. As a fraction $\xi$ of these elites is potentially shifted to the next internal iCEM iteration (\texttt{keep\_elites}), and possibly to the next timestep (\texttt{shift\_elites}), keeping these solutions that ``fail'' and yet bring us closer to the actual solution provides a better strategy to solve tasks which pose an exploration challenge such as stacking. Here, we are also relying on the fact that we are re-running optimization every timestep $t$ in the environment with online model predictive control, such that we have the opportunity to correct these initially ``wrong'' solutions and find their ``stable'' counterparts. Note that this mode of the controller is a more unstable mode compared to \texttt{sum}. Especially with imperfect world models, where the model can hallucinate as the model errors accumulate over the planning horizon, mode \texttt{best} can pick up these falsely imagined future states with low cost. It also doesn't account for the fact that the planned trajectory keeps the lowest cost over multiple timesteps, such that a trajectory where an object flies through the goal location for a single timestep has the same cost as a trajectory where the object lands in the goal position and stays there. To account for this, we use 
$\texttt{cost}^{(p)}=\min\left(\{c(s_{t+h}^{(p)}, a_{t+h}^{(p)}, s_{t+h+1}^{(p)})\}_{h=1}^{H-1}\right)$, where we don't take into account the first timestep of the plan with $h=0$. Although this is not a robust solution, we found it to empirically work well.
Quantitatively, stacking 3 objects when planning with ground truth models yields 99\% success rate for mode \texttt{best}, whereas only 47 \% success rate for \texttt{sum}, using the same reward function in both cases. Even in the case of perfect dynamics predictions with GT models, this showcases the importance of the controller mode to be able solve tasks with sparse reward signals. This is further showcased in \tab{tab:all_versions}, when comparing the results for \oldMethod with default \texttt{best} and \texttt{sum}.

\section{Experiment Details} \label{app:hyperparams}
In this section, we provide experimental details and hyperparameter settings.
\subsection{Environment Details}\label{app:envs}
\paragraph{\Grid} This is a discrete 2D grid, where each entity/agent is controlled individually in the $x$-$y$ directions. This means entities are controlled one at a time and actuation keeps iterating over the entities, where we use an object persistency of 10 timesteps. The action $a_t$ is 2 dimensional, controlling the agent in $x$-$y$ directions separately and is applied on the current actuated entity $i$ in the grid. As we are operating in a discrete grid, the actions are actually discrete such that the agent can move one grid cell to the left/right and up/right (if the target grid cell is not occupied) or stay at the current grid cell. 
In order to make this environment work with the default iCEM implementation with a Gaussian sampling distribution, we perform a discretization step before inputting the sampled actions to the environment. For the experiment results with GT models presented in \fig{fig:envs}, we use a grid size of $25 \times 25$ with 16 and 32 objects.
In colored \Grid with $c$ colors, each object is assigned one of the $c$ available color options, and the observation for each entity also includes binary encoding for color with length $\operatorname{ceil}(\log_2c)$.

\paragraph{\FPP} This is a multi-object manipulation environment as an extension of the Fetch Pick \& Place environment proposed in \cite{li2020towards}. We also applied the two modifications from \citet{Sancaktaretal22}. 1) The table in front of the robot is replaced with a large plane such that objects cannot fall off during free play, but can still be thrown/pushed outside of the robot's reach. 2) In \citet{li2020towards}, the object state also contained the object's position relative to the gripper which was removed, as it already introduces a relational bias in the raw state representation. Details on the dimensionalities of the object and robot state spaces can be found in \tab{tab:environment_parameters}. 

\paragraph{Custom \FPP} As an extension of the standard \FPP, there are  4 different object types with different masses: cube (same as in original \FPP) with size $5$cm and mass 2 (mass in the default unit of Mujoco), flat block with size $16 \times 7 \times 3$cm and mass 1.5, column (upright block) with size $3 \times 3 \times 10$cm and mass 1, and a ball with diameter 8cm and mass 1. We also include categorical variables for each object type as static observations. During free play, we initialize one of each object type, so we have 4 objects. Otherwise the specifications are the same as in \tab{tab:environment_parameters}.

\paragraph{Quadruped \& Walker} These environments are taken from the DeepMind Control Suite \cite{tassa2018deepmind}. To be able to perform \ourMethod computations, we modify the observation space of each environment and append the world coordinates of the robot's joints to the observations vector. In Quadruped, we take the toe and knee positions. We used $x$-$y$-$z$ coordinates with the GT models (results shown in \fig{fig:dmc_rair}), and $x$-$y$ for free play in Quadruped with learned models.
In principle, we can also include the ankle and hip joints, but we observed similar performance in this case. For Walker, we append the feet, leg and torso $x$-$y$ positions.

\subsection{Parameters for Ground Truth Model Experiments with \ourMethod}\label{app:gt_params}
The controller parameters used when optimizing \ourMethod with ground truth (GT) models are given in \tab{tab:controller_settings}. To compute \ourMethod, we perform a discretization step in the \FPP environment as it is continuous. For GT models, that produce perfect mental simulations, we can choose a small bin size of 1cm. In comparison, the size of one block in the environment is 5 cm. The bin size also gives us the upper bound of the \emph{regularity precision} that can be achieved during optimization, \eg a perfectly aligned stack vs. a zigzagged stack. Note however that the higher the precision is, the harder it typically gets for the controller to converge to global optima with a horizon of 30 timesteps and a limited sample-budget.

As discussed in \sec{app:3dim_compression}, this is also a constraint when we are computing \ourMethod in the $x$-$y$-$z$ subspace. Due to this, for the re-creation of existing patterns experiments presented in \sec{sec:recreation} and \sec{app:recreation}, we compute \ourMethod with a bin size of 2.5cm and increase the number of sampled trajectories $P$ to 512. Although we could further increase the bin size, we choose this value to not negatively impact the precision of the re-created structures. In custom \FPP, we also use a bin size of $2.5$cm, when optimizing \ourMethod with GT models.

For Quadruped, we also use a bin size of 5cm. For the GT model experiments, we compute \ourMethod over the $x$-$y$(-$z$) positions of the Quadruped's toe and knee joints. We didn't see significant qualitative differences when including the $z$-positions in the regularity computation. For Walker, we use a bin size of 1cm and compute \ourMethod over the $x$-$y$ positions of the Walker's feet, legs and torso.

\begin{table*}[]
    \centering
    \caption{Base settings for iCEM. These hyperparameters are used when using GT models to optimize \ourMethod.}
    \label{tab:controller_settings}
    \begin{subtable}[t]{.48\textwidth}
        \centering
        \caption{General settings.}
        \begin{tabular}{@{}ll@{}}
        \toprule
        \textbf{Parameter} & \textbf{Value} \\
        \midrule
        Number of samples $P$ & $128$ \\
        Horizon $H$ & $30$ \\
        Size of elite-set $K$ & $10$ \\
        Colored-noise exponent $\beta$ & $3.5$ \\
        \textit{CEM-iterations} & $3$ \\
        Noise strength $\sigma_{\text{init}}$ & $0.8$ \\
        Momentum $\alpha$ & $0.1$ \\
        \texttt{use\_mean\_actions} & Yes \\
        \texttt{shift\_elites} & Yes \\
        \texttt{keep\_elites} & Yes \\
        Fraction of elites reused $\xi$ & $0.3$ \\
        Cost along trajectory & \texttt{best} \\
        \bottomrule
        \end{tabular}
    \end{subtable}%
    \begin{subtable}[t]{.48\textwidth}
    \centering
    \caption{Environment-specific settings.}
    \begin{tabular}{@{}ll@{}}
    \toprule
    \multicolumn{2}{c}{\Grid} \\
    \textbf{Parameter} & \textbf{Value} \\
    \midrule
    Number of samples $P$ & $64$ \\
    \bottomrule
    \\
    \toprule
    \multicolumn{2}{c}{\FPP} \\
    \textbf{Parameter} & \textbf{Value} \\
    \midrule
    \multicolumn{2}{c}{Same as general settings} \\
    \bottomrule
    \\
    \toprule
    \multicolumn{2}{c}{Quadruped \& Walker} \\
    \textbf{Parameter} & \textbf{Value} \\
    \midrule
    Number of samples $P$ & $64$ \\
    Colored-noise exponent $\beta$ & $2.5$ \\
    Noise strength $\sigma_{\text{init}}$ & $0.3$ \\
    Cost along trajectory & \texttt{sum} \\
    \bottomrule
    \end{tabular}
\end{subtable}
\end{table*}

\subsection{Free Play with Learned Models}\label{app:intrinsic}
The environment properties with the episode lengths and model training frequencies are given in \tab{tab:environment_parameters}. Six objects are present in \FPP during free play. The parameters for the GNN model architecture as well as the training parameters for model learning are listed in \tab{tab:gnn_model_training_settings}.
For the \ourMethod computations in free play, we use a bin size of 5cm, which is equivalent to the size of a block.

For custom \FPP, 4 objects (one of each object type) are present during free play. The hyperparameters are the same as for standard \FPP, and a bin size of 5cm is used. 

\begin{table*}[]
    \centering
    \caption{Environment settings for \FPP. 2000 transitions (20 episodes with 100 timesteps each) are generated within one training iteration of free play.}
    \label{tab:environment_parameters}
    \begin{subtable}[t]{.5\textwidth}
        \centering
        \begin{tabular}{@{}ll@{}}
        \toprule
        \multicolumn{2}{c}{\FPP} \\
        \textbf{Parameter} & \textbf{Value} \\
        \midrule
        Episode Length & $100$ \\
        Train Model Every & $20$ Episodes \\
        Action Dim. & $4$ \\
        Robot/Agent State Dim. & $10$ \\
        Object Dynamic State Dim. & $12$ \\
        Num. of Objects During Free Play & $6$\\
        \bottomrule
        \end{tabular}
    \end{subtable}
\end{table*}

\begin{table*}[]
    \centering
    \caption{Settings for GNN model training in intrinsic phase of \ourMethodA and \oldMethod. (Same as in \cite{Sancaktaretal22}) These settings are used for both \FPP and Custom \FPP free play runs.}
    \label{tab:gnn_model_training_settings}
    \begin{subtable}[t]{.5\textwidth}
        \centering
        \begin{tabular}{@{}ll@{}}
        \toprule
        \textbf{Parameter} & \textbf{Value} \\
        \midrule
        Network Size of $g_\text{node}$ & $2 \times 128$ \\
        Network Size of $g_\text{edge}$ & $2 \times 128$ \\
        Network Size of $g_\text{global}$ & $2 \times 128$ \\
        Activation function & ReLU \\
        Layer Normalization & Yes\\
        Number of Message-Passing & 1\\
        Ensemble Size & 5\\
        Optimizer & ADAM \\
        Batch Size & $125$ \\
        Epochs & $25$ \\
        Learning Rate & $10^{-5}$ \\
        Weight Decay & $0.001$ \\
        Weight Initialization & Truncated Normal\\
        Normalize Input & Yes \\
        Normalize Output & Yes \\
        Predict Delta & Yes \\
        \bottomrule
        \end{tabular}
    \end{subtable}%
\end{table*}

\begin{table*}[]
    \centering
    \caption{Settings for the MLP model training in intrinsic phase of \ourMethodA and \oldMethod in the Quadruped free play runs.}
    \label{tab:mlp_quadruped}
    \begin{subtable}[t]{.5\textwidth}
        \centering
        \begin{tabular}{@{}ll@{}}
        \toprule
        \textbf{Parameter} & \textbf{Value} \\
        \midrule
        Episode Length & $100$ \\
        Train Model Every & $20$ Episodes \\
        Network Size & $3 \times 600$ \\
        Activation function & SiLU \\
        Ensemble Size & 5\\
        Optimizer & ADAM \\
        Batch Size & $128$ \\
        Epochs & $25$ \\
        Learning Rate & $0.0001$ \\
        Weight decay & $$0.0001$$ \\
        Weight Initialization & Truncated Normal\\
        Normalize Input & Yes \\
        Normalize Output & Yes \\
        Predict Delta & Yes \\
        \bottomrule
        \end{tabular}
    \end{subtable}%
\end{table*}

The set of the hyperparameters for the iCEM controller used in the intrinsic phase of \ourMethodA, \ourMethod and \oldMethod are the same as presented in \tab{tab:controller_settings}. The only difference to the GT model case is, we use a planning horizon of 20 timesteps for free play. We also use the exact same settings for the RND baseline. For Disagreement, we have a planning horizon of 1 during free play.

For the Quadruped environment, we use an MLP ensemble instead of a GNN ensemble. The hyperparameters are reported in \tab{tab:mlp_quadruped}. The iCEM controller parameters used in free play for Quadruped are the same as the ones for GT models in \tab{tab:controller_settings}, except for the noise strength $\sigma_{\text{init}} = 0.5$.

\subsection{Extrinsic Phase: Zero-shot Downstream Task Generalization}\label{app:extrinsic}
In this section, we provide details on the extrinsic phase following free play, where the learned GNN ensemble is used to solve downstream tasks zero-shot via model-based planning. 

\subsubsection{Details on Downstream Tasks and Reward Functions}\label{app:rewards}
The reward functions for all the \FPP downstream tasks are computed as specified in \citet{Sancaktaretal22}, where for all the assembly tasks we use the same structure as in the stacking reward. However, we do one modification to the original reward computation in the assembly tasks. The assembly task reward is sparse incremental with reward shaping, where the reward also contains the distance between the gripper and the position of the next block to be stacked. We modify how the next block ID is computed in the original implementation from \citet{Sancaktaretal22}. Instead of naively checking the number of unsolved objects to obtain the next block ID irrespective of order, we determine the next block to be the next unsolved block in the order. We found this modification to be important especially for the Pyramid tasks, where the sub-optimal next block computation might lead to the agent receiving a reward to be close to the wrong block, in the case the robot places blocks with $i > \texttt{next\_block\_id}$ to their goal locations with just the sparse reward.

For the RoboYoga experiments, the reward function is the same as proposed in \cite{Mendonca2021:LEXA} (taken from their code in \url{https://github.com/orybkin/lexa-benchmark}). It is a dense reward as the shortest angle distance to the goal pose.

\subsubsection{Planning Details for Downstream Tasks}\label{app:planning_extrinsic}
\begin{table}
    \centering
  \caption{Settings for the iCEM controller used for zero-shot generalization in the extrinsic phase of \ourMethodA and \oldMethod. The settings not specified here are the same as the general settings given in \tab{tab:controller_settings}. The settings are exactly the same as in \cite{Sancaktaretal22}.}
    \label{tab:extrinsic_controller}
    \renewcommand{\arraystretch}{1.03}
    \resizebox{1.\linewidth}{!}{ %
    \begin{tabular}{@{}l|ccccc@{}}
    \toprule
    \textbf{Task}     & \multicolumn{5}{c}{\textbf{Controller Parameters}} \\
                      & Horizon & Colored-noise exponent  & \texttt{use\_mean\_actions} & Noise strength & Cost Along\\
                      & $h$ & $\beta$ &  & $\sigma_\text{init}$ & Trajectory \\
    \midrule
    \FPP-Stacking      &    30  &  3.5  &   No  &  0.5 &    \texttt{best}     \\
    \FPP-Pick \& Place &    30  &  3.5  &  Yes  &  0.5 &    \texttt{best}     \\
    \FPP-Throwing      &    35  &  2.0  &  Yes  &  0.5 &    \texttt{sum}     \\
    \FPP-Flipping      &    30  &  3.5  &   No  &  0.5 &    \texttt{sum}     \\
  \bottomrule
    \end{tabular}
   }

\end{table}

The controller settings for the different downstream tasks are shown in \tab{tab:extrinsic_controller}, which are the same settings used in \cite{Sancaktaretal22}. 

\section{Details on Baselines} \label{app:baselines_detail}
\subsection{Random Network Distillation} \label{app:rnd}
Random network distillation (RND) \cite{burda2018exploration} has two networks: a predictor network and a randomly-initialized and fixed target network. The predictor network $f_\theta: S \xrightarrow{} \mathbb{R}^k$ is trained on the data collected by the agent to match the outputs of the target network $\hat{f}: S \xrightarrow{} \mathbb{R}^k$, where $k$ is the embedding dimension. The error between the predictor and target networks for the state $s_t$ is used as the intrinsic reward signal: 
\begin{equation}
    r_{\text{intrinsic}} = \|f_\theta(s_t) - \hat{f}(s_t)\|^2,
\end{equation}
where the idea is that this error will be higher for novel states. RND can be seen as an extension of count-based methods to continuous domains. The parameters used for the RND module is given in \tab{tab:rnd_settings}. 

\begin{table*}[]
    \centering
    \caption{Settings for the RND predictor $ f_\theta$ and target $\hat{f}$ networks.}
    \label{tab:rnd_settings}
    \begin{subtable}[t]{.5\textwidth}
        \centering
        \begin{tabular}{@{}ll@{}}
        \toprule
        \textbf{Parameter} & \textbf{Value} \\
        \midrule
        Train Predictor Every & $2$K Transitions \\
        Network Size & $1 \times 256$ \\
        Embedding Dim $k$ & $128$ \\
        Activation function & ReLU \\
        Optimizer & ADAM \\
        Batch Size & $256$ \\
        Epochs & $10$ \\
        Learning Rate & $10^{-4}$ \\
        Weight Initialization & Orthogonal\\
        Normalize Input & Yes \\
        \bottomrule
        \end{tabular}
    \end{subtable}%
\end{table*}

Unlike prior work \cite{burda2018exploration}, we don't use RND with an exploration policy. Instead, we have a custom implementation using the same model-based planning backbone for RND as \oldMethod, \ourMethodA and \ourMethod.  By using RND with iCEM instead of an exploration policy, we gain sample-efficiency, as also showcased in \cite{Sancaktaretal22}. Note that although RND intrinsic reward itself is detached from the model and does not require an ensemble, we still deploy an ensemble of GNNs in this case. In previous work \cite{Sancaktaretal22}, a single GNN model was used. In order to compute the RND intrinsic reward, we take the mean predictions of the ensembles. We do this because ensembles have been shown to be more robust models \cite{chua2018}, such that there is no inherent disadvantage to the RND baseline.

\subsection{Disagreement}
The Disagreement baseline \cite{pathak2019self} uses the same intrinsic reward metric as \oldMethod, given in equation \eqnp{eqn:epistemic}. We implement it with the same model-based planning backbone as in \oldMethod, replacing the exploration policy in \cite{pathak2019self}. The only difference to \oldMethod is that the planning horizon during free play is only 1-step. The model and iCEM controller settings are otherwise the same as for \oldMethod, \ourMethodA and \ourMethod. Essentially, the comparison between this baseline and \oldMethod showcases the importance of multi-step planning for future novelty. Note that the planning horizon is only 1 for the free play phase. For fair comparison, we use the same horizon when solving the downstream tasks as for all other methods.

\section{Discussion on Goal-conditioned RL for Unsupervised Exploration} \label{app:gcrl}

In our setup with model-based planning for unsupervised RL, the focus is on distilling knowledge into a world model that is robust enough such that we can plan for diverse downstream tasks afterwards. In the case of the goal-conditioned RL literature, the focus is on distilling this knowledge into a goal-conditioned policy, as in \cite{Mendonca2021:LEXA, hu2023planning, openai2021:assymmetric-selfplay}.

Although the exploration itself is unsupervised and guided by intrinsic rewards, the moment we solve a downstream task we need a goal specification from the environment. Whether you use a goal-conditioned policy to solve it or as in our case, use this goal to compute a reward function, is a separate distinction.

In goal-conditioned RL, you also need access to rewards at training time: the goals are used to compute this reward signal that is used to train the policy and value function. In a way, GCRL is supervised in terms of how you specify your goals (which goal space you use) and the rewards with respect to these goals. We consider the reward function in GCRL to be actually known to the agent. The nice part in GCRL is, you can put in very general supervision with minimal assumptions, that don't require much knowledge of environment dynamics. However, doing so, GCRL is restricting the class of tasks you can solve, namely to tasks that can be described as a goal. But in a wide range of environments, it is hard to specify tasks as a goal: for example running as fast as possible in Halfcheetah.

Our goal is to learn a robust world model that we can use for planning in a later extrinsic phase. In our setup, during the intrinsic phase we make similar assumptions as Plan2Explore \cite{sekar2020planning}. For the extrinsic phase, we need some specification of the task. For CEE-US, a reward function is provided at test time, whereas for GCRL it is a goal (which in fact also implies a reward function).

In model-based planning, since we have complete freedom of choice for the reward function, we can specify any task that can be solved in an MDP setting. Note that we can use our setup to also optimize a goal-conditioned sparse reward. The challenge for us is the limitedness of the zero-order trajectory optimizer: solving long horizon tasks with just sparse rewards becomes challenging, especially with imperfect world models. That’s why in practice, we fall back to dense rewards for planning.

\section{Connections between Compression and \ourMethod} \label{app:compression}
Our regularity objective, that seeks out low-entropy states with high redundancy, shares close ties with compression, and specifically with lossless compression using entropy coding.

We implemented a version of our regularity idea using the lossless compression algorithm \texttt{bzip2} corresponding to the direct version of \ourMethod with order $k=1$ (\sec{sec:rair_intro}). In this case, we describe the state $s$ directly by the properties of each of the entities and the function $\phi: \mathcal S_\obj \to \{\mathcal X\}^+$, that maps each entity to a set of symbols and obtain $\Phi(s) = \Cup_{i=1}^N \phi(s_{\obj,i})$ as a union of all symbols for $N$ objects. 
Instead of computing the entropy for the frequencies of occurrence in the resulting multiset of symbols like in \ourMethod, we instead transform these symbols into bytes and compress them with \texttt{bzip2}. We then define the intrinsic reward for compression as the negative length of the compressed ByteString such that:
\begin{equation}
    r_{\mathrm{compression}} = -\operatorname{len}(\texttt{bzip2.compress}(\{\Cup_{i=1}^N \phi(s_{\obj,i})\}^+\texttt{.tobytes()})).
\end{equation}

We also managed to create regular shapes and patterns when optimizing for $r_{\mathrm{compression}}$ via planning with ground truth models and also for free play with learned models. The reason we chose not to pursue this direction was because 1) lossless compression algorithms like \texttt{bzip2} don't perform as well on short ByteStrings, which is the case for us, as \eg in \FPP, we compress only 6 objects with their corresponding $x$-$y$ positions, 2) artifacts are introducted to the regularity/compression metric by the transformation into bytes, where certain symbols become more compressible than others in this representation without any added regularity.
As a result, we preferred our formulation with \ourMethod as it provides better control over the generated patterns and structures.

\section{Discussion on Object-centric Representations} \label{app:object_centric}
We want to further discuss our reliance on object-centric or more broadly entity-centric representations of the world for our regularity computations.
We view the world as a collection of entities and their interactions. This is a very general assumption, and one that is commonly used for its potential to improve sample efficiency and generalization of learning algorithms in a wide range of domains \cite{locatello2020object, tsividis2021human, battaglia2016interaction}. The principles underlying entity segmentation and their importance for our perception are also studied in cognitive science \cite{spelke1990principles, spelke1993gestalt, peters2021capturing}.

Entity itself is a very abstract and fluid concept. For us as humans, it dynamically changes depending on the granularity of the control problem we want to solve. For example, if we want to move a plate of cookies from the counter to the table, each cookie on the plate is not treated as a separate entity, instead we view the whole plate as one. If we later want to eat the cookies, then the entity-view changes: now each cookie is its own entity. (Example adapted from a private discussion with Maximilian Seitzer.)

Deciding on which level of abstraction to use in which control setting is non-trivial and a research direction which we find very exciting. In object-manipulation environments that are currently encountered in RL (with hard rigid bodies and non-composite structures), we don’t find much ambiguity on entity identifications, hence our statement in \sec{sec:discussion} that observations per object in these cases are naturally disentangled. However, in open-ended real world scenarios it is indeed needed to dynamically choose the right level of abstraction in the perception of a scene and identify the necessary set of entities for the control problem at hand.
Using \ourMethod for a real-world e.g. cluttered scene would be very challenging without closing this action-perception loop. With orthogonal research in visual perception, we are hopeful that this will be possible and we see these synergies as exciting future work. 

\section{Code and Compute}
Code is available on the project webpage \url{https://sites.google.com/view/rair-project}.
We run the ground truth model experiments on CPUs. As we are using the true environment simulator as a model, each imagination step in the planning horizon takes as long as an environment step. We parallelize the ground truth models on 16 virtual cores
The controller frequency in this case is ca. $0.25 \operatorname{Hz}$, for the settings given in \tab{tab:controller_settings}.
For the free-play phase, we have a fixed number of transitions collected at each training iteration, which get added to the replay buffer. After the data collection, the model is trained on the whole replay buffer for 25 epochs. Since the buffer size increases at each training iteration with newly collected data, for this fixed number of epochs, the corresponding number of model training updates and thus also the runtime of the iteration, increase throughout free play. For \ourMethodA, the full free-play (300 training iterations) in \FPP with 6 objects, where overall 600K data points are collected, takes roughly 87 hours using a single GPU (NVIDIA GeForce RTX 3060) and 6 cores on an AMD Ryzen 9 5900X Processor. The majority of this time is spent on the model training after data collection. The controller frequency for the collected rollouts with \ourMethod and the epistemic uncertainty calculations using a GNN ensemble is ca. $5 \operatorname{Hz}$.

\end{document}